\documentclass{article}

\pdfoutput=1
\usepackage[letterpaper,margin=1in,headheight=14pt]{geometry}
\usepackage{times}
\usepackage[round,authoryear]{natbib}
\setcitestyle{citesep={;},aysep={,},yysep={;}}
\usepackage{fancyhdr}
\renewcommand{\headrulewidth}{0pt}
\renewcommand{\footrulewidth}{0pt}
\fancypagestyle{plain}{%
  \fancyhf{}%
  \fancyhead[L]{\small\itshape Under review}%
  \fancyfoot[C]{\thepage}%
  \renewcommand{\headrulewidth}{0pt}%
}

\usepackage[T1]{fontenc}

\title{FlatLands: Indoor floormap completion\\ 
from a single egocentric view}

\author{Subhransu S. Bhattacharjee \qquad Dylan Campbell \qquad Rahul Shome\\[0.5em]
\small School of Computing, The Australian National University\\
\small Canberra, Australia\\
\small\texttt{Subhransu.Bhattacharjee@anu.edu.au}}
\date{}

\usepackage{amsmath,amssymb,bbm}
\usepackage{booktabs,tabularx,longtable,array,ragged2e}
\usepackage[dvipsnames]{xcolor}
\definecolor{mapborder}{gray}{0.5}
\usepackage{graphicx}
\usepackage{float,enumitem}
\usepackage{tikz,pgf-pie,wrapfig}
\usetikzlibrary{arrows.meta,positioning,fit,calc,backgrounds}
\usepackage{caption,subcaption}
\usepackage[hyphens]{url}
\usepackage{xurl, microtype}
\usepackage[breaklinks,colorlinks,citecolor=blue,linkcolor=blue,urlcolor=blue]{hyperref}
\pdftrailerid{}
\providecommand{\doi}[1]{doi: \href{https://doi.org/#1}{\begingroup\urlstyle{rm}\nolinkurl{#1}\endgroup}}
\usepackage[capitalize]{cleveref}
\usepackage{xspace}
\newsavebox{\mapfigurebox}
\newlength{\mapgridheight}
\newcommand{\mapoverlay}[2]{%
  \begingroup
  \sbox{\mapfigurebox}{#2}%
  \setlength{\mapgridheight}{\dimexpr\ht\mapfigurebox+\dp\mapfigurebox\relax}%
  \leavevmode\hbox{%
    \usebox{\mapfigurebox}%
    \kern-\wd\mapfigurebox
    \rlap{\smash{%
      \begin{tikzpicture}[baseline=0pt]
        \path[use as bounding box] (0,0) rectangle (0,0);
        \begin{scope}[x=\wd\mapfigurebox,y=\mapgridheight,yshift=-\dp\mapfigurebox]
          #1%
        \end{scope}
      \end{tikzpicture}%
    }}%
    \kern\wd\mapfigurebox
  }%
  \endgroup
}
\newcommand{\mapgrid}[3]{%
  \mapoverlay{%
    \foreach \mapcolumn in {0,...,#1} {%
      \draw[overlay,mapborder,line width=0.3pt]
        ({\mapcolumn/#1},0) -- ({\mapcolumn/#1},1);
    }
    \foreach \maprow in {0,...,#2} {%
      \draw[overlay,mapborder,line width=0.3pt]
        (0,{\maprow/#2}) -- (1,{\maprow/#2});
    }
  }{#3}%
}
\newcommand{\mapframe}[1]{\mapgrid{1}{1}{#1}}
\newcommand{\constructionframes}[1]{%
  \mapoverlay{%
    \foreach \mapleft/\mapright in {8/208,211/411,415/615,620/820} {%
      \draw[overlay,mapborder,line width=0.3pt]
        ({(3*\mapleft-20)/2461},{332/573})
        rectangle ({(3*\mapright-20)/2461},{532/573});
    }
  }{#1}%
}

\newcommand{\eg}{\emph{e.g.}\xspace}

\newcommand{\etc}{\emph{etc.}\xspace}

\newcommand{\RVRegion}{R}
\newcommand{\RVRegionUobs}{\RVRegion_{U}}
\newcommand{\RVRegionObs}{\RVRegion_{O}}

\newcommand{\dataset}{FlatLands\xspace}

\newcommand{\FloorGT}{F^{\star}}
\newcommand{\FloorObs}{F_{\text{obs}}}
\newcommand{\FloorPredDef}{{F}_{\text{uno}}}
\newcommand{\FloorPred}{\hat{F}_{\text{uno}}}

\newcommand{\FloorComp}{F_{\text{comp}}}

\newcommand{\MaskUnknown}{U}
\newcommand{\MaskValid}{V}
\newcommand{\Ones}{\mathbf{1}}
\newcommand{\Had}{\odot}
\newcommand{\CondDist}{q_{\theta}}

\newcommand{\CellSize}{\Delta}
\newcommand{\ObsImage}{I}
\newcommand{\DepthImage}{D}
\newcommand{\FloorSegImage}{S}
\newcommand{\CamIntrinsics}{\mathbf{K}}
\newcommand{\PreprocessOp}{\Phi}
\newcommand{\ProjectOp}{\Psi}
\newcommand{\MaskObsFootprint}{O}
\newcommand{\MaskEval}{\RVRegion_{\mathrm{eval}}}
\newcommand{\CondSignalRatio}{r_{\mathrm{cond}}}
\newcommand{\NumSamples}{K}

\newcommand{\prob}{P}

\newcommand{\EnergyScoreHat}{\widehat{\mathrm{MES}}}

\begin{document}
\maketitle
\begin{abstract}
A single egocentric image typically captures only a small portion of the floor, yet a complete metric traversability map of the surroundings would better serve applications such as indoor navigation. 
We introduce 
FlatLands, a dataset and benchmark for single-view indoor floor-map completion in metric bird’s-eye view (BEV), evaluating eleven models: both deterministic and generative. FlatLands contain 270,575 observations from 17,656 scene records across six metric 3D datasets, with aligned observation, visibility, validity, and ground-truth maps and both in- and out-of-distribution protocols. 
We compare training-free, deterministic, ensemble, and generative methods on shared geometry-derived inputs and, without fine-tuning, on maps estimated from real egocentric RGB images. 
Our results show that predicting floor almost everywhere can achieve high IoU when most evaluated cells are floor, while generative sample sets outperform the best single-output model under a proper scoring rule and support safer risk-weighted planning. FlatLands establishes a rigorous testbed for evaluating both floor-map accuracy and how effectively alternative completions represent the ambiguity of unseen space.
\end{abstract}

\section{Introduction}
\label{sec:introduction}
Indoor autonomy requires decisions from incomplete, noisy, and
viewpoint-limited evidence~\citep{KAELBLING1998,Curtis-RSS-24}.
In one egocentric image, occlusion and a limited field of view hide much
of the floor. Scene completion infers unseen
geometry~\citep{Song2017SSCNet,Cao2022MonoScene}, yet several layouts
fit one image: a doorway behind furniture may or may not exist, and each
case implies a different route.
\begin{figure}[!hb]
\centering
\includegraphics[width=\textwidth]{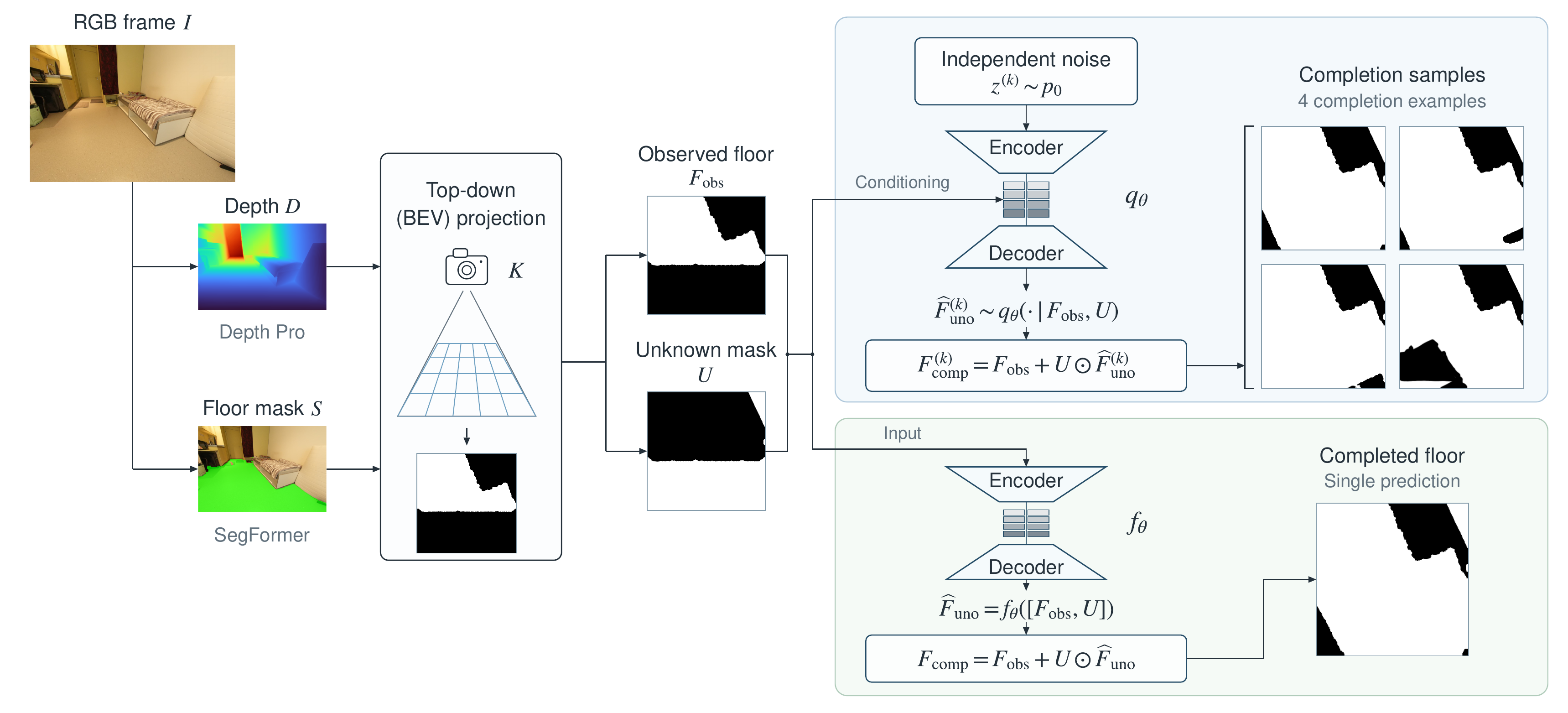}
\caption{\textbf{Single-view floormap completion.}
Depth and floor segmentation from RGB are projected using camera intrinsics
to obtain $(\FloorObs,\MaskUnknown)$.
\textbf{Upper branch:} these maps condition the generative model, while
independent noise draws produce alternative completions.
\textbf{Lower branch:} the deterministic model receives the concatenated
maps directly and predicts a single completion.}
\label{fig:task_pipeline}
\end{figure}

For planar indoor navigation, a metric 2D grid that retains floor layout
and connectivity suffices~\citep{ho2025mapex,Wang2025CogniPlan}: it
supports reachability queries, collision checking, and cost-based route
selection~\citep{lavalle}. Such grids are also economical. Widely used
navigation stacks plan on 2D
costmaps~\citep{MarderEppstein2010OfficeMarathon,Macenski2020Marathon2},
whereas volumetric maps need dedicated data structures to remain
tractable~\citep{Hornung2013OctoMap}. Our deterministic models complete a
$10{\times}10$\,m floormap in under $0.1$\,s (\cref{sec:baselines}).
Anticipating unobserved floor aids
exploration~\citep{ramakrishnan2020occant}. We therefore study
\emph{single-view indoor bird's-eye-view (BEV) floor completion}:
predicting floor labels in unobserved cells of a metric ground-plane grid
while preserving the supplied observed labels.

Generative models~\citep{ho2020denoising,Lipman2023FlowMatching} can draw
many completions of one observation and increasingly support robotic
perception and planning~\citep{reed2024scenesense,Carvalho2025MPD}.
Aggregate IoU cannot show whether these samples help navigation: it
rewards predicting floor wherever floor predominates and scores one map
at a time.
\dataset therefore judges completions individually, as plausible sets,
and by the routes they support, with every method given the same evidence.

\phantomsection
\noindent\textbf{Contributions.}\label{sec:contributions}
\textbf{(i)} We formalize single-view metric
indoor BEV floor completion and introduce \dataset, with
$270{,}575$ observations from $17{,}656$
metric scene records across six
datasets~\citep{mp3d,dai2017scannet,yeshwanth2023scannetpp,baruch2021arkitscenes,wald2019rio,zind};
each observation has four aligned binary maps and provenance metadata. Five sources
are in-distribution (ID); ScanNet++ is held out for out-of-distribution
(OOD) evaluation.
\textbf{(ii)} A source-geometry track gives all methods identical
geometric evidence; an RGB-to-BEV track evaluates the same trained weights, without
fine-tuning, on maps estimated from real images. Both tracks score single outputs, oracle
best-of-$K$ candidates, and completion sets, the last with the strictly
proper masked energy score (MES;~\citealp{gneiting2007proper};
\cref{app:mes}); a challenging subset isolates cases with scarce observed
floor and a non-floor majority.
\textbf{(iii)} Across eleven methods, overlap and utility rank methods
differently. Predicting floor everywhere scores well on overlap yet
collapses when visible floor is scarce. U-Net produces the most accurate
single maps, yet every generator scores better as a set, and planning
over generated samples yields safer routes ($83\%$ Safe Success) than
planning on U-Net's maps or on the same FM+XAttn samples collapsed into
one map ($78\%$ and $80\%$, respectively; \cref{tab:route_safety}).

\section{Background and Related Work}
\label{sec:background}

\noindent\textbf{Mapping and prediction beyond observations.}
Classical occupancy mapping integrates sensor evidence
\citep{elfes1989occupancy,thrun2005probabilistic,Hornung2013OctoMap}.
Occupancy anticipation and local 3D synthesis infer unseen geometry
\citep{ramakrishnan2020occant,reed2024scenesense}, while unobserved
object detection predicts distributions over the categories and locations
of occluded or out-of-view objects \citep{uod}.
Footprints predicts visible and occluded traversable geometry within
the camera frustum from one RGB image; it is evaluated indoors and
demonstrated in path planning \citep{Watson2020Footprints}.
\dataset evaluates metric BEV floor support in occluded and
out-of-frustum regions under shared observation and validity masks.

\noindent\textbf{Layouts, occupancy, and BEV representations.}
Related representations include room layouts
\citep{Lee2017RoomNet,zou2018layoutnet,sun2019horizonnet},
scan-derived floorplans and structured scene descriptions
\citep{Liu2018FloorNet,Chen2019FloorSPSemanticSegmentation,avetisyan2024scenescript},
and volumetric occupancy
\citep{Song2017SSCNet,Cao2022MonoScene,Zhu2026GEMOcc}.
360BEV maps panoramic RGB-D observations to indoor semantic BEV
maps \citep{Teng2024_360BEV}.
Autonomous driving methods construct BEV maps from multiple cameras, sometimes
also using previous frames \citep{PhilionFidler2020LiftSplatShoot,Li2022BEVFormer}.
\dataset isolates floor completion by giving all methods the same
geometric maps, then tests the same models on maps estimated from RGB images
(\cref{sec:interface_rationale}).

\noindent\textbf{Generating and scoring multiple completions.}
Inpainting networks \citep{Liu2018PartialConv,Suvorov2022LAMA,Lugmayr2022RePaint},
diffusion models, and flow matching \citep{ho2020denoising,Lipman2023FlowMatching,cfm}
can all complete partial maps. Conditional generators produce alternative
outcomes \citep{kohl2018probunet}; ensembles combine independently trained
predictors \citep{Lakshminarayanan2017DeepEnsembles}.
A set of completions contains multiple floormaps predicted from the
same observation. Energy scores~\citep{gneiting2007proper,rizzo2016energy}
assess their accuracy and diversity jointly, without selecting a map
using ground truth.

\noindent\textbf{Navigating with predicted maps.}
Predicted maps support exploration and navigation through ensemble
uncertainty~\citep{Georgakis2022UPEN}, estimated information
gain~\citep{ho2025mapex,Baek2025PIPE}, and multiple plausible
layouts~\citep{Wang2025CogniPlan}.
An occupancy-grid study also examines the value of retaining map
uncertainty for route selection~\citep{banfiworth}.
SceneSense applies generative priors to 3D occupancy prediction using
RGB-D observations and an accumulated map~\citep{reed2024scenesense}.
\dataset provides a common single-view floor-support representation
across six indoor datasets and includes an evaluation with
conditioning derived from a single egocentric RGB image.
For controlled comparison, we adapt and train architectures from prior
work using shared input and output representations, then assess
completion quality and the utility of the predicted maps for
point-goal navigation.

\section{Benchmark Task and Interface}
\label{sec:problem}

\phantomsection\label{subsec:prelim}
\noindent\textbf{Task: unobserved floormap completion.}
A \emph{floormap} is a binary floor-support map on an $H{\times}W$ camera-aligned ground-plane grid at resolution $\CellSize$ (m/px), with one denoting floor. The target $\FloorGT$ supplies labels on cells marked valid by $\MaskValid$. The observed and unobserved masks $\MaskObsFootprint$ and $\MaskUnknown=\Ones-\MaskObsFootprint$ have disjoint supports $\RVRegionObs$ and $\RVRegionUobs$; unobserved cells are occluded or outside the camera frustum. The map $\FloorObs$ records floor labels on $\RVRegionObs$ and is zero elsewhere, so a zero denotes observed non-floor only where $\MaskObsFootprint=1$.
The completion target $\FloorPredDef$ specifies floor labels on valid unobserved cells, selected by $\MaskUnknown\Had\MaskValid$. Multiple layouts can agree with the observed evidence, making completion a conditional inverse problem~\citep{Stuart2010BayesianInverse}. The RGB task targets $\prob(\FloorPredDef\mid\ObsImage)$; the controlled benchmark targets $\prob(\FloorPredDef\mid c)$, where $c=(\FloorObs,\MaskUnknown)$. These distributions can differ because BEV construction discards visual information.
A learned conditional distribution produces candidate completions $\FloorPred^{(k)}\sim\CondDist(\cdot\mid\FloorObs,\MaskUnknown)$ for $k=1,\ldots,\NumSamples$; for deterministic models $\CondDist$ is a point mass and $\NumSamples{=}1$.

\phantomsection\label{subsec:pipeline}
\noindent\textbf{Estimating the observed floormap from an egocentric view.}
The operator $\PreprocessOp$ estimates metric depth and floor segmentation, $(\DepthImage,\FloorSegImage):=\PreprocessOp(\ObsImage)$, using Depth Pro~\citep{bochkovskiy2025depth} for depth and an ADE20K-trained SegFormer~\citep{Xie2021SegFormer,Zhou2019ADE20K} for segmentation. The projection operator $\ProjectOp$ back-projects depth using intrinsics $\CamIntrinsics$, aligns the floor plane, and rasterizes the points by top-down orthographic projection, giving $(\FloorObs,\MaskObsFootprint):=\ProjectOp(\DepthImage,\FloorSegImage;\CamIntrinsics)$. We use calibrated intrinsics when available and estimate them otherwise. Alignment requires no global pose; the source-geometry track instead derives the same pair offline from scene geometry.

\begin{figure}
\centering
\begin{subfigure}[c]{0.49\linewidth}
\centering
\constructionframes{\includegraphics[width=\linewidth,trim=5 0 0 0,clip]{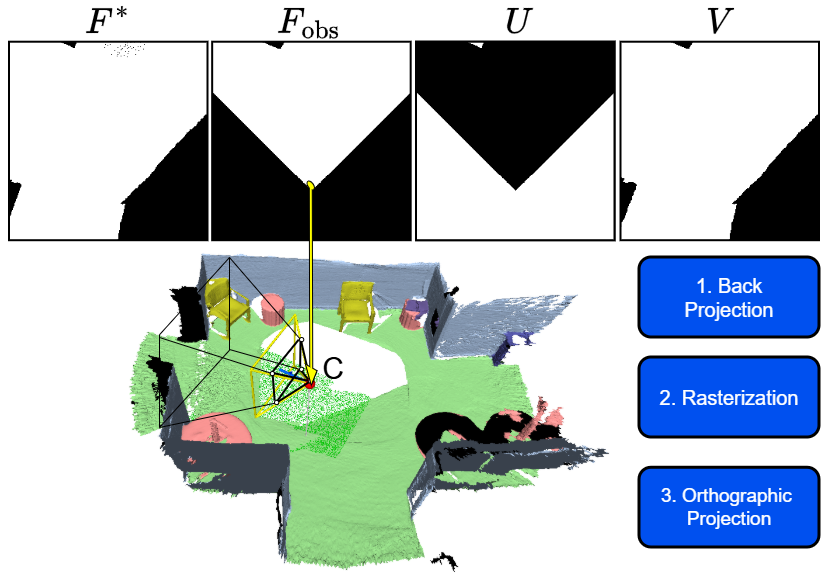}}
\caption{Constructing a camera-aligned floormap.}
\label{fig:data_observation}
\end{subfigure}\hfill
\begin{subfigure}[c]{0.48\linewidth}
\centering
\begin{minipage}[c]{0.39\linewidth}
\centering
\includegraphics[width=\linewidth]{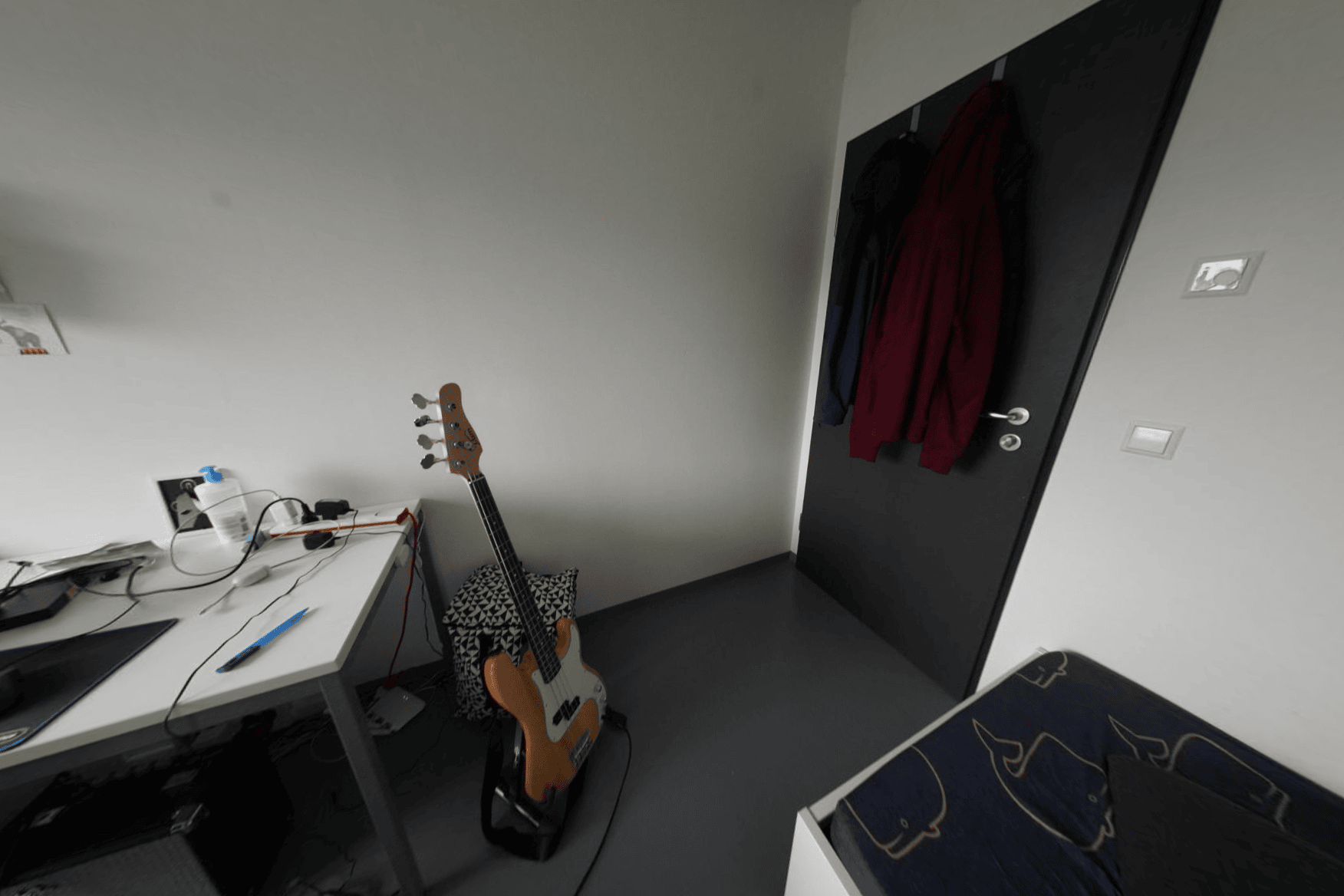}\\[2pt]{\scriptsize RGB input}
\end{minipage}\hfill
\begin{minipage}[c]{0.58\linewidth}
\centering
\setlength{\tabcolsep}{2pt}
\begin{tabular}{@{}cc@{}}
\mapframe{\includegraphics[width=0.44\linewidth]{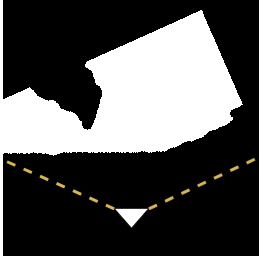}} &
\mapframe{\includegraphics[width=0.44\linewidth]{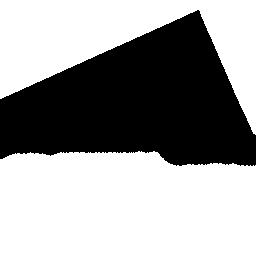}} \\
{\scriptsize $\FloorObs$} & {\scriptsize $\MaskUnknown$} \\[3pt]
\mapframe{\includegraphics[width=0.44\linewidth]{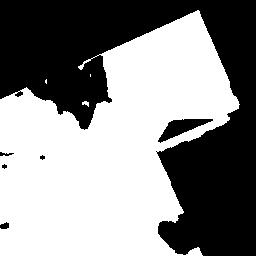}} &
\mapframe{\includegraphics[width=0.44\linewidth]{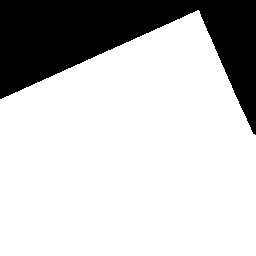}} \\
{\scriptsize $\FloorGT$} & {\scriptsize $\MaskValid$}
\end{tabular}
\end{minipage}
\caption{An RGB image and its four aligned binary maps.}
\label{fig:four_maps}
\end{subfigure}
\caption{\textbf{From a partial view to a floormap completion task.}
(a) Visible floor points are projected onto the floor plane and rasterized in BEV.
(b) Each observation provides maps of observed floor $\FloorObs$, unobserved cells $\MaskUnknown$, complete floor $\FloorGT$, and valid workspace $\MaskValid$; the white marker shows the camera anchor C. The pair $(\FloorObs,\MaskUnknown)$ is the input; $\FloorGT$ and $\MaskValid$ are used for supervision and scoring.}
\label{fig:construction_inputs}
\end{figure}

\phantomsection\label{subsec:training}
\noindent\textbf{Completion assembly and supervision.}
After binarization, each prediction $\FloorPred^{(k)}$ is assembled into a full completion as 
\begin{equation}\label{eq:completion_assembly}
\FloorComp^{(k)} = \FloorObs+\MaskUnknown\Had \FloorPred^{(k)}.
\end{equation}
This construction preserves observed labels and fills only unobserved cells, so errors in the estimated observations remain in the completed map. Supervision and completion metrics use $\MaskEval=\MaskUnknown\Had\MaskValid$ with $|\MaskEval|>0$; neither $\MaskValid$ nor $\FloorGT$ is given to the completion model (\cref{supp:observation_synthesis}). RGB completion scores therefore assess valid unobserved cells, excluding errors in the unchanged observed region.

\phantomsection\label{sec:interface_rationale}
\noindent\textbf{Why not train directly using RGB as input?}
Direct RGB-to-BEV training combines visual feature extraction and projection into metric BEV~\citep{PhilionFidler2020LiftSplatShoot,Li2022BEVFormer} with prediction of unobserved floor. Synthesizing shared BEV maps from source geometry expands viewpoint coverage beyond the captured images and isolates completion quality. Methods can then be compared under identical geometric evidence, independently of the architecture, input modality, and training data of an RGB perception front end. The RGB-to-BEV track tests the resulting models with image-derived inputs (\cref{sec:evaluation}).

\section{\dataset Dataset}
\label{sec:dataset}

\noindent\textbf{Sources and splits.}
\Cref{tab:dataset_sources} lists six sources and $17{,}656$ scene records, with up to $24$ camera poses per scene placed on valid floor; visibility comes from field-of-view and ray-occlusion tests. Curation retains $391{,}024$ of the $423{,}743$ initial observations; the filter $\CondSignalRatio=|\FloorObs|/|\FloorGT|\geq0.10$ leaves $270{,}575$ (\cref{supp:learnability}). Splitting by source scene yields $215{,}342$ training, $26{,}890$ validation, and $28{,}343$ test observations; $16{,}214$ test observations come from held-out ScanNet++ (\cref{tab:supp:sources,fig:supp:pipeline,supp:dataset_details,supp:split_consistency}). Each observation contains four aligned binary maps (\cref{fig:four_maps}) with $\FloorObs\Had\MaskUnknown=0$ and $\FloorObs\leq\MaskValid$.

\noindent\textbf{Ground-truth provenance.}
For five sources, we extract floor labels from reconstructed meshes. For ZInD, we build floormaps from DA$^2$-\emph{estimated} panoramic depth~\citep{li2025da2}, scaled using annotated room vertices. Thus $133{,}096$ of the $270{,}575$ filtered observations ($49\%$) have targets scaled from room annotations. The OOD targets come from meshes (\cref{supp:zind_calibration}).

\noindent\textbf{Representation of floormaps.} All four maps share a floor-aligned egocentric $256{\times}256$ BEV grid spanning $10{\times}10$\,m at $25.6$ pixels per meter, with the camera at pixel $(128,192)$. The median scene floor area is $16.4$\,m$^2$ (\cref{supp:delta_choice,fig:area_histogram,fig:coverage_violin}). The fraction of floor cells in $\MaskEval$ is $0.802\pm0.238$ over the pooled test split, $0.661$ when ID sources are weighted equally, and $0.854$ on OOD. All-Floor attains exactly these IoUs under the corresponding averaging conventions (\cref{tab:main_results_ioubest}). \Cref{sec:results} separately evaluates observations with lower floor prevalence (\cref{fig:label_balance}).

\begin{table}[!hb]
\centering
\normalsize
\captionsetup{font=normalsize}
\setlength{\tabcolsep}{1.5pt}
\renewcommand{\arraystretch}{1}
\caption{\textbf{Source datasets in \dataset{}.} Obs.\ (raw) counts observations before quality filtering; Obs.\ (filtered) counts only observations with $\CondSignalRatio\geq 0.10$, valid support, observed/target floor agreement, and non-degenerate crops. Curation and crop validation retain $391{,}024$ observations before applying the $\CondSignalRatio$ threshold; split-level counts for this intermediate stage appear in \cref{supp:filtering_stages}. ScanNet++ is OOD test-only.}%
\label{tab:dataset_sources}%
\begin{tabularx}{\textwidth}{@{}l r r l >{\RaggedRight\arraybackslash}X@{}}
\toprule
\textbf{Dataset} & \textbf{Scenes} & \shortstack{\textbf{Obs.}\\\textbf{(raw)}} & \shortstack{\textbf{Obs.}\\\textbf{(filtered)}} & \textbf{Asset Used} \\
\midrule

ZInD~\citep{zind} &
7{,}026 & 168{,}624 & 133{,}096 &
Panorama-to-PLY point cloud
\\

ARKitScenes~\citep{baruch2021arkitscenes} &
4{,}803 & 115{,}271 & 40{,}282 &
PLY mesh
\\

Matterport3D~\citep{mp3d} &
2{,}101 & 50{,}424 & 38{,}004 &
Room PLY mesh
\\

ScanNet~\citep{dai2017scannet} &
1{,}508 & 36{,}192 & 24{,}763 &
PLY mesh
\\

3RScan~\citep{wald2019rio} &
1{,}291 & 30{,}984 & 18{,}216 &
OBJ mesh
\\

ScanNet++~\citep{yeshwanth2023scannetpp} &
927 & 22{,}248 & 16{,}214 &
PLY mesh
\\

\midrule
\textbf{\dataset~(Ours)} &
\textbf{17{,}656} & \textbf{423{,}743} & \textbf{270{,}575} &
4$\times$  binary 2D maps\\
\bottomrule
\end{tabularx}
\end{table}

\section{Benchmark Protocol}
\label{sec:baselines}
\phantomsection
\noindent\textbf{Baselines.}
\label{sec:baseline_methods}
Eleven baselines receive the same inputs $(\FloorObs,\MaskUnknown)$ and are scored on $\MaskEval$ (\cref{tab:baseline_catalog}). Training-free methods fill unobserved cells with zero (All-Obstacle) or one (All-Floor), copy the nearest observed label (NN Propagation), or draw Bernoulli$(0.5)$ labels (Uniform Random).

\noindent\textbf{Learned predictors.}
U-Net~\citep{Ronneberger2015UNet} and PConv-UNet~\citep{Liu2018PartialConv} use masked binary cross-entropy. LaMa~\citep{Suvorov2022LAMA} uses masked $\ell_1$, adversarial, and feature-matching losses; LaMa-Ensemble combines four independently trained LaMa predictors~\citep{Lakshminarayanan2017DeepEnsembles}. Diffusion predicts noise~\citep{ho2020denoising}, Flow Matching predicts velocity~\citep{Lipman2023FlowMatching,Liu2023RectifiedFlow}, and FM+XAttn adds conditioning cross-attention~\citep{vaswani2017attention}. The three generators train with masked mean-squared error. At inference, they use guidance $s=2.0$, clamp observed labels, and produce $K=4$ samples. Diffusion uses $50$ DDIM steps; Flow Matching and FM+XAttn use $50$ and $25$ Heun steps, respectively~\citep{DDIM,heun}. \Cref{supp:model_formulations} specifies the objectives and attention resolutions.

\noindent\textbf{Training implementation and inference.}
All learned completion models train on source-geometry BEV maps from the five ID sources, with a shared schedule on $4{\times}$A100 40\,GB GPUs. Single-model runs use seed $42$; LaMa-Ensemble uses seeds $41$--$44$. The default guidance scale $s=2.0$ was selected using the training split (\cref{app:training}). Outputs $\geq0.5$ denote floor. On one A100, completion takes $63$--$81$\,ms for deterministic models, $136$\,ms for LaMa-Ensemble, and $1{,}745$--$6{,}661$\,ms to generate four samples (\cref{app:training,supp:hyperparameters,tab:supp:runtime}).

\phantomsection\label{sec:evaluation}
\noindent\textbf{Reporting protocol.}
Two tracks differ in how conditioning is constructed.
The \emph{source-geometry (canonical) track} uses synthesized BEV maps, as in training (\cref{supp:observation_synthesis}). Its $28{,}343$ test observations cover held-out scenes of the five ID sources and the held-out source ScanNet++ (OOD; \cref{supp:split_consistency}); results appear in \cref{tab:main_results_ioubest,tab:sampling_results}.
The \emph{RGB-to-BEV track} estimates metric depth with pretrained Depth Pro and floor segmentation with SegFormer for every frame, then back-projects, aligns the floor, and rasterizes (\cref{supp:camera_pipeline}). This test-only track reuses the same completion weights without fine-tuning. Its $1{,}000$ real frames comprise $200$ each from held-out test scenes of four ID sources and ScanNet++, excluding ZInD panoramas (\cref{tab:rgb_to_bev,supp:e2e_multi_source}).
In either track, ID denotes sources used for training; OOD denotes held-out ScanNet++. Both report floor-class IoU/F1, mismatch rate (UMR), and masked energy score (MES) on the valid unobserved cells $\MaskEval$, assessing individual predictions and sets of completions separately.

\noindent\textbf{Unobserved-region mismatch rate.}
The unobserved-region mismatch rate (UMR; lower is better) is the binary error rate on $\MaskEval$:
\begin{equation}
    \mathrm{UMR}=\frac{\mathrm{FP}+\mathrm{FN}}{|\MaskEval|},
\end{equation}
where FP and FN count false-positive and false-negative cells within $\MaskEval$. For RGB and challenging-subset results, F1 pools TP/FP/FN across observations, whereas IoU and UMR average scores per observation. The reported F1 and IoU therefore cannot be converted into one another.

\noindent\textbf{Masked energy score.}
For $\NumSamples>1$ completions, MES (lower is better) compares them with the target and with one another~\citep{gneiting2007proper,rizzo2016energy}:
\begin{equation}
\EnergyScoreHat =
\frac{1}{\NumSamples}\sum_{k=1}^{\NumSamples} d_{\MaskEval}\left(\FloorComp^{(k)},\FloorGT\right)
-\frac{1}{2\NumSamples(\NumSamples-1)}\sum_{k\neq\ell} d_{\MaskEval}\left(\FloorComp^{(k)},\FloorComp^{(\ell)}\right).
\label{eq:es_hat}
\end{equation}
The first term measures average error against the target; subtracting the second rewards diversity among completions. Distance is $d_{\MaskEval}(A,B)=1-\mathrm{IoU}_{\MaskEval}(A,B)$: greater floor overlap means smaller distance. This is masked Jaccard distance~\citep{Lipkus1999JaccardMetric}; we use $\NumSamples=4$ samples (\cref{supp:k_sensitivity}).
For deterministic predictions, the population score reduces to $1-\mathrm{IoU}_{\MaskEval}$ (\cref{app:mes}), enabling comparison with sets of completions.

\noindent\textbf{Sample selection.}
Single-output fidelity evaluates one completion. Best-of-$K$ selects the highest-IoU sample using ground truth, giving an oracle upper bound for selection from those candidates. 
LaMa-Ensemble's single-output score evaluates the mean of its four member probability maps, thresholded at $0.5$. IoU$_m$ averages the four IoU scores after thresholding each prediction at $0.5$. MES and variance use the same binary completions; variance divides by four (\cref{supp:metric_interpretation}).

\noindent\textbf{Pooling, precision, and table conventions.}
Equal-source ID summaries weight the five datasets equally; observation-pooled summaries weight observations equally. For equal-source summaries, SD is computed across the five source means. These SDs describe variation across sources or observations, not across training runs. Bold marks the best value and ties at the displayed precision within each comparison block; oracle-selected and single-output results are ranked separately. Per-cell variance (Var) summarizes disagreement among samples (\cref{supp:metrics,supp:extended_comparisons}).

\section{Results and Discussion}
\label{sec:results}

\noindent\textbf{Fidelity.}
\Cref{tab:main_results_ioubest} separates single-output fidelity from oracle sample selection. A single deterministic prediction is hard to beat on overlap: U-Net leads single-output IoU on ID and ties PConv-UNet on OOD, well above the generators' first samples (\cref{tab:id_source_macro}). Sampling closes the gap only with an oracle: on OOD, the best of four samples exceeds U-Net's $0.870$ for every generator, with most of the gain coming from the second sample (\cref{tab:supp:k_sensitivity_iou}). These ground-truth-selected gains measure candidate quality, not a selection rule usable at test time; MES and planning use the full sample set without oracle selection.

\begin{table}
\centering
\normalsize
\captionsetup{font=normalsize}
\setlength{\tabcolsep}{1.5pt}
\renewcommand{\arraystretch}{1}
\caption{Fidelity (mean$\pm$SD). The upper block evaluates single outputs; the lower evaluates best-of-four predictions selected using ground truth. ID single-output entries are equal-source means ($n{=}5$); ID best-of-four and all OOD entries are observation-level, so ID cross-block gains mix selection and weighting. Extended results: \cref{tab:supp:per_dataset_breakdown,tab:supp:k_sensitivity_iou}.}
\label{tab:main_results_ioubest}
\begin{tabular*}{\linewidth}{@{\extracolsep{\fill}}>{\raggedright\arraybackslash}p{72pt}ccc@{\hspace{5pt}}ccc@{}}
\toprule
Method & \multicolumn{3}{c}{ID} & \multicolumn{3}{c}{OOD} \\
\cmidrule(lr){2-4}\cmidrule(lr){5-7}
& UMR$\downarrow$ & IoU$\uparrow$ & F1$\uparrow$
& UMR$\downarrow$ & IoU$\uparrow$ & F1$\uparrow$ \\
\midrule
\multicolumn{7}{@{}l}{\emph{Single output:} one completion per observation} \\
All-Obstacle
& 0.661\textpm0.229 & 0.000\textpm0.000 & 0.000\textpm0.000
& 0.854\textpm0.155 & 0.000\textpm0.000 & 0.000\textpm0.000 \\
All-Floor
& 0.339\textpm0.229 & 0.661\textpm0.229 & 0.737\textpm0.210
& 0.146\textpm0.155 & 0.854\textpm0.155 & 0.912\textpm0.116 \\
NN Propagation
& 0.243\textpm0.078 & 0.620\textpm0.219 & 0.708\textpm0.202
& 0.198\textpm0.156 & 0.779\textpm0.186 & 0.861\textpm0.143 \\
Uniform Random
& 0.501\textpm0.002 & 0.367\textpm0.104 & 0.516\textpm0.128
& 0.502\textpm0.003 & 0.453\textpm0.057 & 0.621\textpm0.064 \\
U-Net
& \textbf{0.122\textpm0.030} & \textbf{0.763\textpm0.132} & \textbf{0.843\textpm0.098}
& \textbf{0.114\textpm0.111} & \textbf{0.870\textpm0.136} & \textbf{0.924\textpm0.094} \\
PConv-UNet
& 0.126\textpm0.033 & 0.758\textpm0.132 & 0.838\textpm0.098
& \textbf{0.114\textpm0.113} & \textbf{0.870\textpm0.139} & \textbf{0.924\textpm0.097} \\
LaMa
& 0.209\textpm0.059 & 0.623\textpm0.161 & 0.734\textpm0.131
& 0.283\textpm0.159 & 0.672\textpm0.190 & 0.787\textpm0.153 \\
\midrule
\multicolumn{7}{@{}l}{\emph{Best-of-four (oracle):} four candidates, one selected using the ground truth} \\
LaMa-Ensemble
& 0.105\textpm0.099 & 0.817\textpm0.205 & 0.882\textpm0.154
& 0.107\textpm0.094 & 0.874\textpm0.126 & 0.927\textpm0.083 \\
Diffusion
& \textbf{0.100\textpm0.097} & \textbf{0.830\textpm0.196} & \textbf{0.892\textpm0.143}
& \textbf{0.094\textpm0.092} & \textbf{0.890\textpm0.122} & \textbf{0.936\textpm0.082} \\
Flow Matching
& \textbf{0.100\textpm0.095} & \textbf{0.830\textpm0.192} & \textbf{0.892\textpm0.139}
& 0.102\textpm0.099 & 0.880\textpm0.129 & 0.930\textpm0.088 \\
FM+XAttn
& \textbf{0.100\textpm0.094} & \textbf{0.830\textpm0.192} & \textbf{0.892\textpm0.139}
& 0.109\textpm0.092 & 0.873\textpm0.124 & 0.927\textpm0.084 \\
\bottomrule
\end{tabular*}
\end{table}

\begin{table}[!htb]
\centering
\normalsize
\captionsetup{font=normalsize}
\setlength{\tabcolsep}{1.5pt}
\renewcommand{\arraystretch}{1}
\caption{Evaluating sets of completions, $K=4$. ID MES and $\mathrm{IoU}_m$ average the five source means; Var and all OOD entries are observation-level. MES combines distance to the target with pairwise sample distance; Var averages the cellwise variance of the four binary predictions (divisor four) over $\MaskEval$. Conventions: \cref{sec:evaluation,tab:supp:per_dataset_stochastic}.}
\label{tab:sampling_results}
\begin{tabular*}{\linewidth}{@{\extracolsep{\fill}}>{\raggedright\arraybackslash}p{72pt}ccc@{\hspace{5pt}}ccc@{}}
\toprule
Method & \multicolumn{3}{c}{ID} & \multicolumn{3}{c}{OOD} \\
\cmidrule(lr){2-4}\cmidrule(lr){5-7}
& MES$\downarrow$ & IoU$_m\uparrow$ & Var
& MES$\downarrow$ & IoU$_m\uparrow$ & Var \\
\midrule
LaMa-Ensemble
& 0.195\textpm0.083 & 0.637\textpm0.175 & 0.076\textpm0.062
& 0.149\textpm0.093 & 0.720\textpm0.155 & 0.084\textpm0.043 \\
Diffusion
& 0.174\textpm0.088 & 0.711\textpm0.151 & 0.044\textpm0.029
& 0.097\textpm0.102 & \textbf{0.832\textpm0.124} & 0.048\textpm0.026 \\
Flow Matching
& 0.173\textpm0.083 & \textbf{0.712\textpm0.150} & 0.037\textpm0.029
& 0.101\textpm0.103 & \textbf{0.832\textpm0.135} & 0.043\textpm0.028 \\
FM+XAttn
& \textbf{0.161\textpm0.076} & \textbf{0.712\textpm0.146} & 0.044\textpm0.032
& \textbf{0.095\textpm0.093} & 0.813\textpm0.123 & 0.061\textpm0.028 \\
\bottomrule
\end{tabular*}
\end{table}

\noindent\textbf{Challenging cases.}
Because All-Floor labels every unobserved cell as floor, its IoU equals floor prevalence, which is why it scores $0.854$ on OOD. The challenging subset removes this advantage: on $1{,}386$ cases with little visible floor and a non-floor majority, All-Floor falls to $0.191$ while U-Net keeps $0.468$, and All-Floor mislabels most evaluated cells (\cref{tab:challenging_main}). Among sampled methods, FM+XAttn leads on MES, mean IoU, and oracle IoU. LaMa-Ensemble's averaged map instead drifts toward All-Floor: members that disagree over broad unobserved regions average above the floor threshold, so its UMR and pooled F1 are worse than those of a single LaMa model.

\begin{table}[!htb]
\centering
\normalsize
\captionsetup{font=normalsize}
\setlength{\tabcolsep}{1.5pt}
\renewcommand{\arraystretch}{1}
\caption{Challenging subset: $1{,}386$ cases with $\CondSignalRatio\leq0.20$ and floor prevalence $<0.50$ (mean $0.19$). UMR and pooled F1 use the first prediction, except for LaMa-Ensemble's mean map ($\S$). IoU evaluates single outputs in the upper block and ground-truth-selected best-of-four predictions in the lower block. Reporting conventions: \cref{sec:evaluation}. Standard deviations: \cref{tab:supp:hard_fidelity}; protocol: \cref{supp:suite_hard}.}
\label{tab:challenging_main}
\begin{tabularx}{\linewidth}{@{}l*{6}{>{\centering\arraybackslash}X}@{}}
\toprule
Method & UMR$\downarrow$ & IoU$\uparrow$ & F1$_{\rm pool}\uparrow$ & MES$\downarrow$ & IoU$_m\uparrow$ & Var \\
\midrule
\multicolumn{7}{@{}l}{\emph{Single output}} \\
All-Floor & $0.809$ & $0.191$ & $0.302$ & -- & -- & -- \\
NN Propagation & $0.337$ & $0.192$ & $0.299$ & -- & -- & -- \\
U-Net & $\mathbf{0.111}$ & $\mathbf{0.468}$ & $\mathbf{0.609}$ & -- & -- & -- \\
PConv-UNet & $\mathbf{0.111}$ & $0.458$ & $0.599$ & -- & -- & -- \\
LaMa & $0.147$ & $0.317$ & $0.463$ & -- & -- & -- \\
\midrule
\multicolumn{7}{@{}l}{\emph{Best-of-four (oracle) in the IoU column}} \\
LaMa-Ensemble & $0.790^{\S}$ & $0.397$ & $0.296^{\S}$ & $0.362$ & $0.295$ & $0.174$ \\
Diffusion & $0.151$ & $0.462$ & $0.558$ & $0.352$ & $0.396$ & $0.055$ \\
Flow Matching & $0.170$ & $0.464$ & $0.515$ & $0.346$ & $0.394$ & $0.040$ \\
FM+XAttn & $0.155$ & $\mathbf{0.479}$ & $0.550$ & $\mathbf{0.316}$ & $\mathbf{0.417}$ & $0.046$ \\
\bottomrule
\end{tabularx}
\end{table}

\noindent\textbf{Evaluating multiple completions.}
Scored as sets, the generators overtake the deterministic models. A deterministic prediction's MES equals $1-\mathrm{IoU}$, so U-Net scores $0.237$ on ID and $0.130$ on OOD; every generator does better on both splits, with FM+XAttn scoring best, whereas LaMa-Ensemble beats U-Net only on ID (\cref{tab:sampling_results}). Removing FM+XAttn's cross-attention changes its scores by only $0.001$ (\cref{tab:supp:xattn_ablation}), and it is the slowest sampler (\cref{tab:supp:runtime}). All three generators perform best at the default guidance scale $s=2.0$ (\cref{tab:supp:cfg_ablation}).

\begin{figure}[!htb]
\centering
\includegraphics[width=\textwidth]{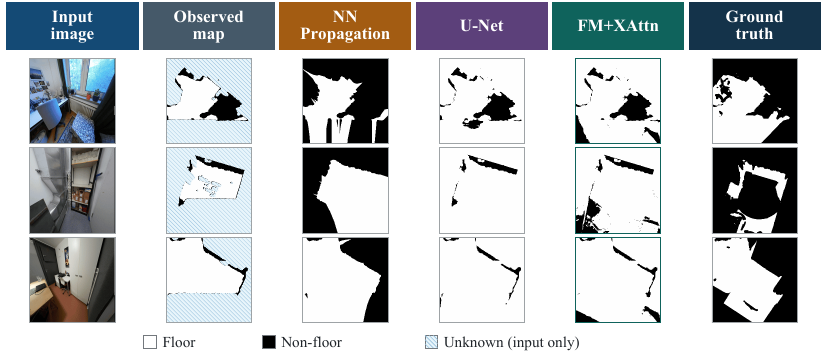}
\caption{\textbf{Three RGB-conditioned benchmark examples.} Each row compares selected baselines on the same input; ground truth at right. FM+XAttn shows its first generated sample. Hatching marks unobserved input cells for display only; the maps remain binary.}
\label{fig:qualitative_b}
\end{figure}

\noindent\textbf{Qualitative observations.}
In \cref{fig:qualitative_b}, NN Propagation extends visible patches into long bands (top row) and a broad wedge (middle row). The middle observation already marks part of the target's central non-floor region as floor; clamping preserves this error. U-Net and FM+XAttn fill extensive unseen regions, missing the bottom target's angular boundary. FM+XAttn's displayed first draw adds some corner exclusions but still overextends floor. \Cref{fig:canonical_full} compares all eleven methods on the canonical track; \cref{fig:qualitative_supp_b} extends the RGB examples.
\Cref{supp:synthetic_ambiguity} adds a constructed example in which one input has five valid candidate targets.

\begin{figure}[!htb]
\centering
\pgfmathsetlengthmacro{\completionmapwidth}{0.119\linewidth}
\setlength{\tabcolsep}{1.5pt}
\begin{tikzpicture}[
  panel/.style={draw=black!40, line width=0.5pt, rounded corners=2pt,
    inner sep=3pt, outer sep=0pt, anchor=north west},
  font=\scriptsize
]
\node[panel] (shared) {\begin{tabular}{@{}cc@{}}
  \multicolumn{2}{c}{\textbf{(a) Shared maps}} \\[2pt]
  Observed $\FloorObs$ & Unobserved $\MaskUnknown$ \\
  \mapframe{\includegraphics[width=\completionmapwidth]{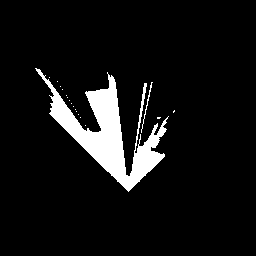}} &
  \mapframe{\includegraphics[width=\completionmapwidth]{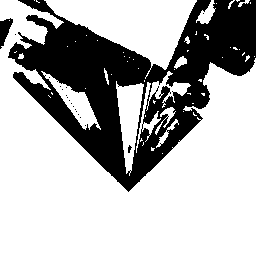}} \\[2pt]
  Target $\FloorGT$ & Validity $\MaskValid$ \\
  \mapframe{\includegraphics[width=\completionmapwidth]{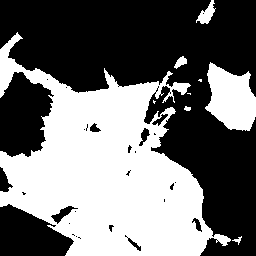}} &
  \mapframe{\includegraphics[width=\completionmapwidth]{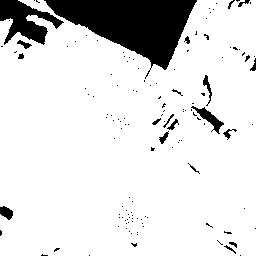}}
\end{tabular}};
\node[panel] (lama) at ([xshift=8pt]shared.north east) {\begin{tabular}{@{}cc@{}}
  \multicolumn{2}{c}{\textbf{(b) LaMa-Ensemble}} \\[2pt]
  1 & 2 \\
  \mapframe{\includegraphics[width=\completionmapwidth]{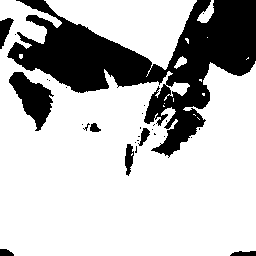}} &
  \mapframe{\includegraphics[width=\completionmapwidth]{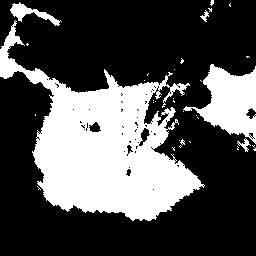}} \\[2pt]
  3 & 4 \\
  \mapframe{\includegraphics[width=\completionmapwidth]{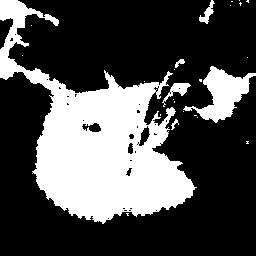}} &
  \mapframe{\includegraphics[width=\completionmapwidth]{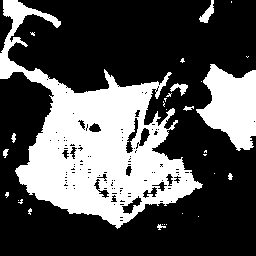}}
\end{tabular}};
\node[panel] (fmx) at ([xshift=8pt]lama.north east) {\begin{tabular}{@{}cc@{}}
  \multicolumn{2}{c}{\textbf{(c) FM+XAttn}} \\[2pt]
  1 & 2 \\
  \mapframe{\includegraphics[width=\completionmapwidth]{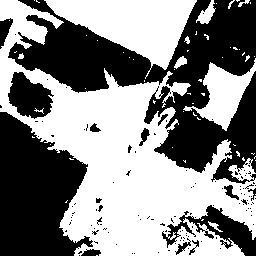}} &
  \mapframe{\includegraphics[width=\completionmapwidth]{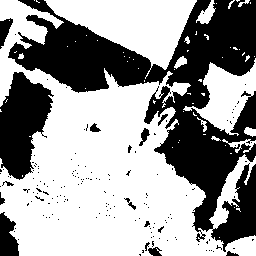}} \\[2pt]
  3 & 4 \\
  \mapframe{\includegraphics[width=\completionmapwidth]{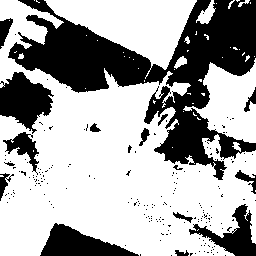}} &
  \mapframe{\includegraphics[width=\completionmapwidth]{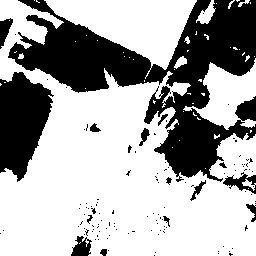}}
\end{tabular}};
\node[panel] (variance) at ([xshift=8pt]fmx.north east) {\begin{tabular}{@{}c@{}}
  \textbf{(d) Variance} \\[2pt]
  LaMa-Ensemble \\
  \mapframe{\includegraphics[width=\completionmapwidth]{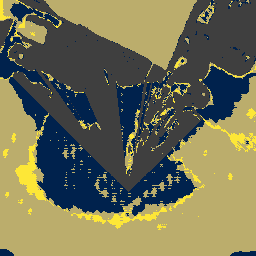}} \\[2pt]
  FM+XAttn \\
  \mapframe{\includegraphics[width=\completionmapwidth]{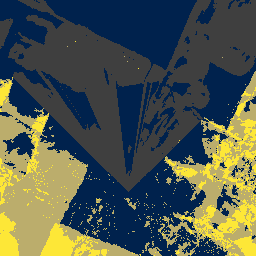}}
\end{tabular}};
\end{tikzpicture}

\begin{tikzpicture}[x=1cm,y=1cm,font=\scriptsize]
\node[anchor=east] at (-0.15,0.13) {Variance};
\node[anchor=south west,inner sep=0pt,outer sep=0pt] at (0,0)
  {\includegraphics[width=3cm,height=0.26cm]{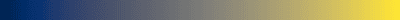}};
\draw[mapborder,line width=0.3pt] (0,0) rectangle (3,0.26);
\foreach \x/\value in {0/0,1.5/0.5,3/1} {
  \draw[line width=0.3pt] (\x,0) -- (\x,-0.06);
  \node[anchor=north] at (\x,-0.06) {$\value$};
}
\end{tikzpicture}
\caption{\textbf{Four completions and sample disagreement, multi-room ScanNet++ scene.} (a) Conditioning (top) and evaluation maps (bottom). (b,c) Four completions per method. (d) Variance of each method's four binary completions on unobserved cells, with divisor four. Both maps use the same scale: $1$ denotes the maximum variance $0.25$ (a two-two split).}
\label{fig:lama_ens_comparison}
\end{figure}

\noindent\textbf{Boundary-band IoU and interior disagreement.}
Scoring only cells near floor edges does not escape the prevalence effect: on OOD boundary-band IoU~\citep{boundary}, All-Floor ($0.820$) matches best-of-four Diffusion ($0.822$) and beats single-output U-Net ($0.809$). Where samples disagree is more telling: FM+XAttn shows almost no interior disagreement, as its samples rarely differ inside floor regions, unlike LaMa-Ensemble's members (\cref{tab:supp:boundary_iou_full,fig:lama_ens_comparison,supp:boundary_radius}).

\begin{figure}[!htb]
\centering
\resizebox{\columnwidth}{!}{%
\begin{tikzpicture}[
  img/.style={inner sep=0pt, outer sep=0pt},
  lbl/.style={font=\scriptsize, anchor=south, yshift=2pt},
  predbox/.style={draw=black!50, rounded corners=3pt, inner sep=2.5pt, line width=0.4pt},
]
\def\imgw{1.8cm}
\def\gap{0.20cm}
\def\secgap{0.35cm}

\node[img] (obs) at (0,0)
  {\fcolorbox{mapborder}{white}{\includegraphics[width=\imgw]{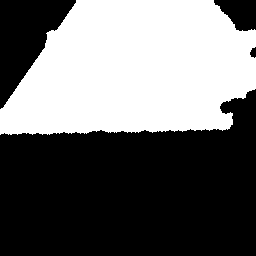}}};
\node[img, right=\gap of obs] (gt)
  {\fcolorbox{mapborder}{white}{\includegraphics[width=\imgw]{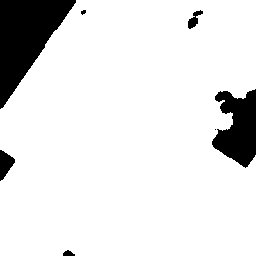}}};
\node[img, right=\gap of gt] (dpatch)
  {\mapframe{\includegraphics[width=\imgw]{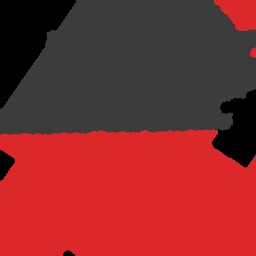}}};

\node[img, right=\secgap of dpatch] (diff)
  {\fcolorbox{mapborder}{white}{\includegraphics[width=\imgw]{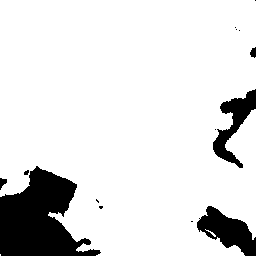}}};
\node[img, right=\gap of diff] (fm)
  {\fcolorbox{mapborder}{white}{\includegraphics[width=\imgw]{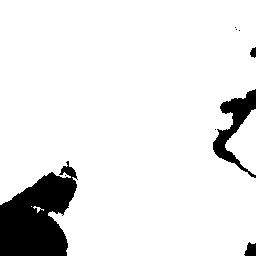}}};
\node[img, right=\gap of fm] (fmx)
  {\fcolorbox{mapborder}{white}{\includegraphics[width=\imgw]{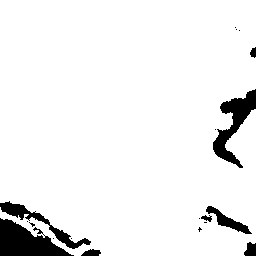}}};
\node[img, right=\gap of fmx] (lama)
  {\fcolorbox{mapborder}{white}{\includegraphics[width=\imgw]{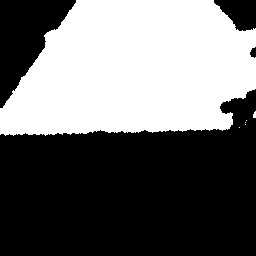}}};
\node[predbox, fit=(diff)(fm)(fmx)(lama),
  label={[font=\scriptsize\bfseries, anchor=south, yshift=-1pt]above:Predictions}] {};

\node[lbl] at (obs.north)    {$\FloorObs$};
\node[lbl] at (gt.north)     {$\FloorGT$};
\node[lbl] at (dpatch.north) {$\FloorGT \!\setminus\! \FloorObs$};
\node[font=\scriptsize, anchor=north, yshift=-2pt] at (diff.south)  {Diffusion};
\node[font=\scriptsize, anchor=north, yshift=-2pt] at (fm.south)    {Flow Matching};
\node[font=\scriptsize, anchor=north, yshift=-2pt] at (fmx.south)   {FM+XAttn};
\node[font=\scriptsize, anchor=north, yshift=-2pt] at (lama.south)  {LaMa};

\end{tikzpicture}}%
\caption{\textbf{Failure under wide occlusion.} Red: unseen target floor ($\FloorGT\setminus\FloorObs$). Sampled maps cross target boundaries; LaMa predicts almost no floor (displayed IoU $0.003$; mean over seeds 41--44: $0.006$). Each generator shows its first generated sample; LaMa shows the single model trained with seed $42$. More examples: \cref{fig:supp_failure_grid}.}
\label{fig:failure_case}
\end{figure}

\noindent\textbf{End-to-end RGB-to-BEV evaluation.}
The same pattern holds on maps estimated from real images, without fine-tuning (\cref{tab:rgb_to_bev,fig:qualitative_b}). U-Net again produces the best single map (IoU $0.564$), and the best of four generator samples at best roughly matches it, yet every generator scores better as a set, with MES of $0.293$--$0.308$ versus U-Net's $0.436$. Diffusion replaces FM+XAttn as the MES leader. Boundary leakage, missing floor, and shape drift persist (\cref{fig:failure_case}).

\begin{table}[!htb]
\centering
\normalsize
\captionsetup{font=normalsize}
\setlength{\tabcolsep}{1.5pt}
\renewcommand{\arraystretch}{1}
\caption{RGB-to-BEV over $1{,}000$ real frames ($200$ from each of five pinhole sources; ZInD excluded). UMR, $\mathrm{IoU}_1$, and pooled F1 evaluate single outputs: the first sample for generators and the thresholded mean map for LaMa-Ensemble, which can outperform every individual member. $\mathrm{IoU}_b$ reports ground-truth-selected best-of-four IoU. IoU$_m$, MES, and Var use all four outputs, with Var averaged over $\MaskEval$; All-Obstacle and Uniform Random appear in \cref{tab:supp:e2e_fidelity}. Conventions: \cref{sec:evaluation}.}
\label{tab:rgb_to_bev}
\begin{tabularx}{\linewidth}{@{}l*{7}{>{\centering\arraybackslash}X}@{}}
\toprule
Method & UMR$\downarrow$ & IoU$_1\uparrow$ & F1$_{\rm pool}\uparrow$ & IoU$_b\uparrow$ & MES$\downarrow$ & IoU$_m\uparrow$ & Var \\
\midrule
All-Floor & $0.559$ & $0.441$ & $0.508$ & -- & -- & -- & -- \\
NN Propagation & $0.467$ & $0.248$ & $0.328$ & -- & -- & -- & -- \\
\midrule
U-Net & 0.435 & \textbf{0.564} & \textbf{0.650} & -- & -- & -- & -- \\
PConv-UNet & 0.437 & 0.563 & 0.649 & -- & -- & -- & -- \\
LaMa & 0.521 & 0.148 & 0.223 & -- & -- & -- & -- \\
\midrule
LaMa-Ensemble & $0.486$ & $0.439$ & $0.539$ & $0.414$ & $0.302$ & $0.233$ & $0.141$ \\
Diffusion & \textbf{0.432} & 0.497 & 0.604 & $0.522$ & $\mathbf{0.293}$ & $0.504$ & $0.047$ \\
Flow Matching & 0.449 & 0.542 & 0.632 & $0.564$ & $0.306$ & $0.552$ & $0.038$ \\
FM+XAttn & 0.440 & 0.560 & 0.647 & $\mathbf{0.565}$ & $0.308$ & $\mathbf{0.562}$ & $0.041$ \\
\bottomrule
\end{tabularx}
\end{table}

\begin{table}[!htb]
\centering
\normalsize
\captionsetup{font=normalsize}
\setlength{\tabcolsep}{1.5pt}
\renewcommand{\arraystretch}{1}
\caption{Offline point-goal planning from RGB-derived BEV conditioning. Found: path returned; Safe Success: route entirely on valid ground-truth floor; SPL also penalizes path length. CIs: 95\% Wilson intervals~\citep{wilson}. Bold marks the best learned planner, excluding All-Floor and the oracle. $^\dagger$$^\ddagger$\,Each marked pair has identical reported scores.}
\label{tab:route_safety}
\begin{tabular*}{\linewidth}{@{\extracolsep{\fill}}lccccc@{}}
\toprule
Planner & \shortstack{Found\\(\%) $\uparrow$} & CI & \shortstack{Safe Success\\(\%) $\uparrow$} & CI & SPL $\uparrow$ \\
\midrule
All-Floor & 100.00 & [99.90, 100.00] & 62.68 & [61.16, 64.16] & 0.610 \\
U-Net / PConv-UNet~$^\dagger$ & 86.85 & [85.77, 87.86] & 78.20 & [76.89, 79.45] & 0.780 \\
FM+XAttn mean & 90.35 & [89.40, 91.23] & 79.55 & [78.27, 80.77] & 0.794 \\
Diffusion / Flow Matching~$^\ddagger$ & \textbf{96.93} & [96.34, 97.42] & 83.05 & [81.86, 84.18] & 0.800 \\
FM+XAttn risk-weighted & 95.73 & [95.05, 96.31] & \textbf{83.28} & [82.09, 84.40] & \textbf{0.829} \\
Ground-truth oracle & 100.00 & [99.90, 100.00] & 100.00 & [99.90, 100.00] & 1.000 \\
\bottomrule
\end{tabular*}
\end{table}

\phantomsection
\noindent\textbf{Offline point-goal planning.}
\label{sec:route_safety}
We evaluate $4{,}000$ start--goal tasks with feasible ground-truth paths in a separate cohort of $500$ RGB-to-BEV test scenes ($250$ ID, $250$ OOD). All planners run Dijkstra search within the valid support. Binary-map planners use unit costs on their maps; risk-weighted planners admit any cell that some sample labels floor and charge more where fewer samples agree (\cref{supp:planning_cost}).
Keeping the samples pays off: every risk-weighted generative planner reaches the goal safely more often than the planner using U-Net's map, with non-overlapping confidence intervals (\cref{tab:route_safety}). The most direct comparison is within FM+XAttn: collapsing its samples into one thresholded map lowers Safe Success from $83.28\%$ to $79.55\%$, although cell admission and costs both change. All-Floor finds every path, but only $62.68\%$ are safe. The three generators perform similarly, so the benefit is not tied to one model.

\section{Conclusion}
\label{sec:conclusion}

\dataset grounds the inverse problem of inferring multiple plausible spatial layouts from partial observations in a concrete robotics task, enabling systematic evaluation. Eleven methods, from training-free rules to generative models, receive identical inputs, masks, and scores across six indoor datasets, a held-out source, and maps estimated from real images. Its answer is that the most accurate single map is not the most useful one. Overlap rewards guessing floor wherever floor is common. Sets of completions, scored with a proper rule and used for risk-aware planning, lead to safer routes than any single prediction. And the way samples disagree is itself informative: a conditional generator such as FM+XAttn is nearly certain inside floor regions, whereas ensemble members are not. These findings hold on the held-out OOD source and on real RGB inputs.

\noindent\textbf{Applications.} Because the interface is an abstract binary grid, completion is sensor-agnostic. Completed floormaps can serve as probabilistic spatial priors~\citep{matterdoor} for belief-space planning~\citep{kurniawati2022partially,Curtis-RSS-24}, active perception~\citep{ramakrishnan2020occant}, SLAM back-ends~\citep{slam}, generative motion planning~\citep{Serifi2024RobotMDM}, and planning over multiple map hypotheses~\citep{Wang2025CogniPlan}. The natural next step is closed-loop navigation under map uncertainty~\citep{Savva2019Habitat,ho2025mapex}, where a robot acts on its samples and observes the consequences.

\noindent\textbf{Limitations \& Future Work.}
This benchmark targets \emph{binary} traversability only, without semantic cues, and assumes a simplified pinhole camera model that does not capture the diversity of real robotic cameras. It also omits \emph{graded} traversability (\eg stairs, slopes, \etc), which
would better reflect
the traversability constraints of heterogeneous embodied platforms like mobile, legged, or aerial robots. Failure analysis (\cref{fig:failure_case}) indicates that when most ground-truth floor is unobserved, boundary placement becomes underdetermined; consequently, all methods produce local defects poorly captured by IoU alone. Promising directions include expanding the source datasets to cover more
architectural styles and building types, incorporating RGB and semantic cues for conditioning, adopting traversability definitions for different robotic systems, moving beyond single-view input to multi-view settings, and extending baselines to more model architectures~\citep{backbones}.

\label{maintext:end}

\clearpage
\subsection*{Reproducibility statement}
All learned completion models train on source-geometry BEV maps from the five ID sources. The source-geometry (canonical) track evaluates held-out scenes from these sources and the entirely held-out source ScanNet++ (OOD). The RGB-to-BEV track evaluates real images from held-out test scenes of four ID sources and ScanNet++, using the same completion weights without fine-tuning. Every frame uses depth estimated by Depth Pro and floor segmentation from SegFormer to construct conditioning.
\Cref{sec:problem,sec:evaluation} define the shared interface, evaluation tracks, and metrics. Dataset construction and scene splits, model objectives, and training settings appear in \cref{supp:dataset_details,supp:model_formulations,supp:hyperparameters}, respectively. RGB preprocessing is specified in \cref{supp:camera_pipeline} and evaluation-frame selection in \cref{supp:e2e_multi_source}. \Cref{supp:applications,supp:synthetic_ambiguity} specify offline planning and the constructed multi-solution example; \cref{supp:extended_comparisons} gives extended comparisons.
Code, trained weights, the FlatLands dataset, metadata, and experiment artifacts will be released alongside a public leaderboard. Source assets remain distributed by their providers under the original licenses.

\subsection*{Ethics statement}
\label{supp:ethics}
\dataset targets assistive robotics, accessibility mapping, and indoor navigation. RGB imagery and binary floormaps can reveal sensitive spatial information and enable unauthorized mapping, so collection, release, and use require appropriate permissions and privacy safeguards. Source assets are excluded from our release and must be obtained from their providers under the original dataset licenses. Consent, institutional review, and image filtering depend on source protocols; permissions and privacy considerations also apply to derived maps. The emphasis on North American and European interiors may limit generalization across building types, accessibility needs, and floor-prevalence distributions. Physical traversability and closed-loop safety require platform- and user-specific assessment. Compute is approximately $280$ A100 GPU-hours for seven learned method families on a shared institutional cluster partly powered by renewable energy.
\subsection*{AI use statement}
Large language models assisted with editing prose, developing code, and refining figure presentation. They were not used to formulate the research questions or to develop the methods.
Separately, the pipeline itself uses pretrained vision models: DA$^2$ for ZInD panoramic depth (\cref{sec:dataset}), and Depth Pro and SegFormer for depth estimation and floor segmentation in the RGB-to-BEV track (\cref{sec:evaluation,supp:camera_pipeline}).
The authors take full responsibility for all claims, results, and artifacts.

\bibliographystyle{abbrvnat}
\bibliography{references}

\clearpage

\setcounter{equation}{0}
\renewcommand{\theequation}{A\arabic{equation}}
\renewcommand{\theHequation}{appendix.\arabic{equation}}
\crefalias{section}{appendix}
\crefalias{subsection}{appendix}
\crefalias{subsubsection}{appendix}

\appendix
\section*{Appendix}
\phantomsection\label{supp:guide}
\noindent\textbf{Reader's guide.}
The appendix supports each main-paper result with its protocol, extended tables, and qualitative evidence. \Cref{tab:reader_guide} maps each result to its supporting material. Sections are ordered as extended results (\cref{supp:expanded_metrics}), dataset construction (\cref{supp:dataset_details}), the RGB front end (\cref{supp:camera_pipeline}), baselines and implementation (\cref{supp:model_formulations,app:training}), metrics (\cref{supp:metrics}), and qualitative examples (\cref{supp:qualitative_section}).

\begin{table}[!h]
\centering
\normalsize
\captionsetup{font=normalsize}
\setlength{\tabcolsep}{4pt}
\renewcommand{\arraystretch}{1.15}
\hyphenpenalty=10000\exhyphenpenalty=10000
\caption{\textbf{Where to find supporting material for each main-paper result.}}
\label{tab:reader_guide}
\begin{tabularx}{\linewidth}{@{}>{\RaggedRight\arraybackslash}p{0.38\linewidth}>{\RaggedRight\arraybackslash}X@{}}
\toprule
\textbf{Main-paper result} & \textbf{Supporting material} \\
\midrule
Single-output and best-of-$K$ fidelity (\cref{tab:main_results_ioubest}) & Per-source fidelity, UMR by conditioning ratio, and sample-count sensitivity (\cref{supp:extended_comparisons,supp:k_sensitivity}) \\
Sets of completions and MES (\cref{tab:sampling_results}) & Per-source MES (\cref{tab:supp:per_dataset_stochastic}); MES propriety and estimators (\cref{supp:metrics}) \\
Challenging subset (\cref{tab:challenging_main}) & Selection criteria and standard deviations (\cref{supp:suite_hard}) \\
RGB-to-BEV transfer (\cref{tab:rgb_to_bev}) & Cohort design (\cref{supp:e2e_multi_source}); front end (\cref{supp:camera_pipeline}); galleries (\cref{fig:qualitative_supp_b}) \\
Offline point-goal planning (\cref{tab:route_safety}) & Task construction, map-dependent costs, and SPL (\cref{supp:applications}) \\
Interior disagreement (\cref{fig:lama_ens_comparison}) & Boundary radius, regional variance, and boundary-band IoU (\cref{supp:boundary_radius}) \\
Constructed multi-solution example & Construction and distributional metrics (\cref{supp:synthetic_ambiguity}) \\
Dataset, splits, and filtering (\cref{sec:dataset}) & Sources, observation synthesis, curation, conditioning tiers, and splits (\cref{supp:dataset_details}) \\
Baselines, training, and runtime (\cref{sec:baselines}) & Formulations, hyperparameters, runtime, and guidance and cross-attention ablations (\cref{supp:model_formulations,app:training}) \\
Qualitative results and failures (\cref{fig:qualitative_b,fig:failure_case}) & Full canonical comparison, RGB galleries, and failure grid (\cref{supp:qualitative_section}) \\
\bottomrule
\end{tabularx}
\end{table}

\section{Extended Benchmark Results}
\label{supp:expanded_metrics}

\subsection{Extended Comparisons}
\label{supp:extended_comparisons}

\begin{figure}
\centering
\includegraphics[width=0.78\linewidth]{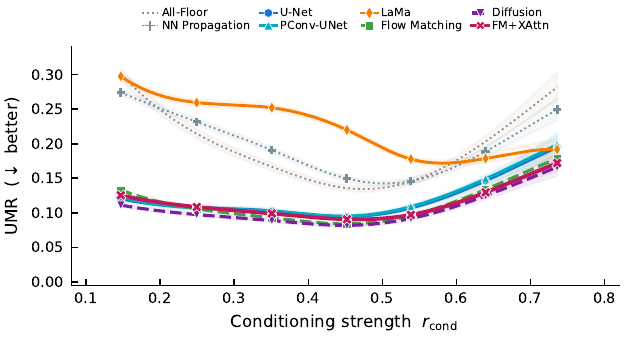}
\caption{UMR on the canonical track by conditioning ratio. All-Floor UMR equals one minus target-floor prevalence within each conditioning bin.
Generator UMR is computed from the first generated completion per observation.
}
\label{fig:umr_rcond}
\end{figure}

\Cref{tab:id_source_macro} compares single-output and best-of-$K$ IoU; \cref{tab:supp:per_dataset_breakdown,tab:supp:per_dataset_stochastic} report source-level fidelity and sample-set scores. \Cref{fig:umr_rcond} stratifies canonical UMR by conditioning ratio.

\begin{table}[!hb]
\centering
\normalsize
\captionsetup{font=normalsize}
\setlength{\tabcolsep}{1.5pt}
\renewcommand{\arraystretch}{1}
\caption{\textbf{Single-output and best-of-$K$ IoU.} $\mathrm{IoU}_1$ evaluates one output: a generator's first sample or, for LaMa-Ensemble, the mean of its four member probability maps thresholded at $0.5$. Best of 2 and $\mathrm{IoU}_b$ select the highest-IoU prediction among the first two and four candidates, respectively, using ground truth; ensemble candidates are individual members in seed order $41,42,43,44$. ID $\mathrm{IoU}_1$ ($\ddagger$) uses equal-source averaging, and all OOD entries use observation-level averaging (\cref{sec:evaluation}).}
\label{tab:id_source_macro}
\begin{tabular*}{\linewidth}{@{\extracolsep{\fill}}lcccc@{}}
\toprule
& ID$^{\ddagger}$ & \multicolumn{3}{c}{OOD} \\
\cmidrule(lr){2-2}\cmidrule(lr){3-5}
Method & $\mathrm{IoU}_1\uparrow$ & $\mathrm{IoU}_1\uparrow$ & Best of 2 $\uparrow$ & $\mathrm{IoU}_b\uparrow$ \\
\midrule
\multicolumn{5}{@{}l}{\emph{Deterministic}} \\
U-Net & $\mathbf{0.763}$ & $\mathbf{0.870}$ & -- & -- \\
PConv-UNet & $0.758$ & $\mathbf{0.870}$ & -- & -- \\
\midrule
\multicolumn{5}{@{}l}{\emph{Sampled, $K{=}4$}} \\
LaMa-Ensemble & $0.661$ & $0.853$ & $\mathbf{0.867}$ & $0.874$ \\
Diffusion & $0.656$ & $0.734$ & $0.866$ & $\mathbf{0.890}$ \\
Flow Matching & $0.667$ & $0.771$ & $0.862$ & $0.880$ \\
FM+XAttn & $0.659$ & $0.725$ & $0.842$ & $0.873$ \\
\bottomrule
\end{tabular*}
\end{table}

\begin{table}
\centering
\normalsize
\captionsetup{font=normalsize}
\setlength{\tabcolsep}{1.5pt}
\renewcommand{\arraystretch}{1}
\caption{Per-source single-output fidelity (mean$\pm$SD). ID covers the five in-distribution sources; OOD is ScanNet++. Single-model runs use seed $42$; LaMa-Ensemble uses seeds $41$--$44$. Each fidelity entry evaluates one output per observation. For LaMa-Ensemble, this is the mean of four member probability maps, thresholded at $0.5$ (\cref{sec:evaluation}). ARKit denotes ARKitScenes; MP3D denotes Matterport3D. RGB rows evaluate a separate cohort of $1{,}000$ ScanNet++ RGB frames; multi-source RGB results appear in \cref{tab:supp:e2e_fidelity}.}
\label{tab:supp:per_dataset_breakdown}
\begin{tabular*}{\linewidth}{@{\extracolsep{\fill}}lclcc@{}}
\toprule
Method & Split & Dataset & IoU$\uparrow$ & F1$\uparrow$ \\
\midrule
LaMa & ID & 3RScan & $0.519\pm0.272$ & $0.640\pm0.244$ \\
LaMa & ID & ARKit & $0.420\pm0.189$ & $0.567\pm0.187$ \\
LaMa & ID & MP3D & $0.738\pm0.197$ & $0.832\pm0.155$ \\
LaMa & ID & ScanNet & $0.620\pm0.198$ & $0.745\pm0.171$ \\
LaMa & ID & ZInD & $0.818\pm0.193$ & $0.884\pm0.148$ \\
LaMa & OOD & ScanNet++ & $0.672\pm0.190$ & $0.787\pm0.153$ \\
LaMa-Ensemble & ID & 3RScan & $0.546\pm0.372$ & $0.621\pm0.357$ \\
LaMa-Ensemble & ID & ARKit & $0.308\pm0.265$ & $0.414\pm0.285$ \\
LaMa-Ensemble & ID & MP3D & $0.829\pm0.211$ & $0.887\pm0.177$ \\
LaMa-Ensemble & ID & ScanNet & $0.765\pm0.179$ & $0.853\pm0.134$ \\
LaMa-Ensemble & ID & ZInD & $0.859\pm0.197$ & $0.907\pm0.160$ \\
LaMa-Ensemble & OOD & ScanNet++ & $0.853\pm0.153$ & $0.911\pm0.115$ \\
LaMa-Ensemble & RGB & ScanNet++ RGB & $0.690\pm0.223$ & $0.792\pm0.191$ \\
Diffusion & ID & 3RScan & $0.607\pm0.249$ & $0.722\pm0.222$ \\
Diffusion & ID & ARKit & $0.485\pm0.207$ & $0.626\pm0.197$ \\
Diffusion & ID & MP3D & $0.738\pm0.169$ & $0.837\pm0.130$ \\
Diffusion & ID & ScanNet & $0.681\pm0.157$ & $0.799\pm0.120$ \\
Diffusion & ID & ZInD & $0.769\pm0.165$ & $0.858\pm0.126$ \\
Diffusion & OOD & ScanNet++ & $0.734\pm0.148$ & $0.837\pm0.109$ \\
Diffusion & RGB & ScanNet++ RGB & $0.721\pm0.209$ & $0.818\pm0.169$ \\
Flow Matching & ID & 3RScan & $0.598\pm0.273$ & $0.707\pm0.243$ \\
Flow Matching & ID & ARKit & $0.451\pm0.209$ & $0.593\pm0.204$ \\
Flow Matching & ID & MP3D & $0.770\pm0.179$ & $0.856\pm0.136$ \\
Flow Matching & ID & ScanNet & $0.706\pm0.161$ & $0.816\pm0.122$ \\
Flow Matching & ID & ZInD & $0.810\pm0.182$ & $0.882\pm0.137$ \\
Flow Matching & OOD & ScanNet++ & $0.771\pm0.151$ & $0.862\pm0.109$ \\
Flow Matching & RGB & ScanNet++ RGB & $0.704\pm0.196$ & $0.808\pm0.162$ \\
\bottomrule
\end{tabular*}
\end{table}

\begin{table}
\centering
\normalsize
\captionsetup{font=normalsize}
\setlength{\tabcolsep}{1.5pt}
\renewcommand{\arraystretch}{1}
\ContinuedFloat
\caption[]{Per-source single-output fidelity (continued). Reporting conventions and cohort definitions follow the first part of this table.}
\begin{tabular*}{\linewidth}{@{\extracolsep{\fill}}lclcc@{}}
\toprule
Method & Split & Dataset & IoU$\uparrow$ & F1$\uparrow$ \\
\midrule
FM+XAttn & ID & 3RScan & $0.599\pm0.252$ & $0.714\pm0.223$ \\
FM+XAttn & ID & ARKit & $0.466\pm0.192$ & $0.612\pm0.185$ \\
FM+XAttn & ID & MP3D & $0.746\pm0.169$ & $0.842\pm0.128$ \\
FM+XAttn & ID & ScanNet & $0.675\pm0.152$ & $0.795\pm0.118$ \\
FM+XAttn & ID & ZInD & $0.808\pm0.179$ & $0.881\pm0.135$ \\
FM+XAttn & OOD & ScanNet++ & $0.725\pm0.144$ & $0.832\pm0.106$ \\
FM+XAttn & RGB & ScanNet++ RGB & $0.729\pm0.205$ & $0.824\pm0.166$ \\
U-Net & ID & 3RScan & $0.694\pm0.260$ & $0.787\pm0.215$ \\
U-Net & ID & ARKit & $0.567\pm0.209$ & $0.700\pm0.184$ \\
U-Net & ID & MP3D & $0.867\pm0.157$ & $0.919\pm0.113$ \\
U-Net & ID & ScanNet & $0.803\pm0.160$ & $0.880\pm0.115$ \\
U-Net & ID & ZInD & $0.885\pm0.161$ & $0.929\pm0.119$ \\
U-Net & OOD & ScanNet++ & $0.870\pm0.136$ & $0.924\pm0.094$ \\
U-Net & RGB & ScanNet++ RGB & $0.761\pm0.219$ & $0.844\pm0.170$ \\
PConv-UNet & ID & 3RScan & $0.694\pm0.261$ & $0.787\pm0.215$ \\
PConv-UNet & ID & ARKit & $0.560\pm0.213$ & $0.693\pm0.187$ \\
PConv-UNet & ID & MP3D & $0.862\pm0.164$ & $0.915\pm0.117$ \\
PConv-UNet & ID & ScanNet & $0.791\pm0.171$ & $0.872\pm0.126$ \\
PConv-UNet & ID & ZInD & $0.881\pm0.165$ & $0.926\pm0.122$ \\
PConv-UNet & OOD & ScanNet++ & $0.870\pm0.139$ & $0.924\pm0.097$ \\
PConv-UNet & RGB & ScanNet++ RGB & $0.766\pm0.218$ & $0.847\pm0.170$ \\
\bottomrule
\end{tabular*}
\end{table}

\begin{table}
\centering
\normalsize
\captionsetup{font=normalsize}
\setlength{\tabcolsep}{1.5pt}
\renewcommand{\arraystretch}{1}
\caption{Per-source and pooled sample-set scores (observation-level mean$\pm$SD), using all $K=4$ outputs. ID covers five sources; OOD is ScanNet++. Pooled ID weights observations across all five sources. ID MES and IoU$_m$ in \cref{tab:sampling_results} average the five source means equally. ARKit denotes ARKitScenes; MP3D denotes Matterport3D.}
\label{tab:supp:per_dataset_stochastic}
\begin{tabular*}{\linewidth}{@{\extracolsep{\fill}}lclcc@{}}
\toprule
Method & Split & Dataset & MES$\downarrow$ & IoU$_m$$\uparrow$ \\
\midrule
LaMa-Ensemble & ID & 3RScan & $0.243\pm0.148$ & $0.530\pm0.270$ \\
LaMa-Ensemble & ID & ARKit & $0.308\pm0.111$ & $0.398\pm0.182$ \\
LaMa-Ensemble & ID & MP3D & $0.132\pm0.111$ & $0.764\pm0.179$ \\
LaMa-Ensemble & ID & ScanNet & $0.188\pm0.102$ & $0.660\pm0.159$ \\
LaMa-Ensemble & ID & ZInD & $0.102\pm0.116$ & $0.831\pm0.180$ \\
LaMa-Ensemble & ID & Pooled & $0.158\pm0.139$ & $0.713\pm0.246$ \\
LaMa-Ensemble & OOD & ScanNet++ & $0.149\pm0.093$ & $0.720\pm0.155$ \\
Diffusion & ID & 3RScan & $0.215\pm0.179$ & $0.637\pm0.276$ \\
Diffusion & ID & ARKit & $0.306\pm0.142$ & $0.483\pm0.211$ \\
Diffusion & ID & MP3D & $0.107\pm0.124$ & $0.823\pm0.160$ \\
Diffusion & ID & ScanNet & $0.150\pm0.125$ & $0.763\pm0.145$ \\
Diffusion & ID & ZInD & $0.091\pm0.121$ & $0.849\pm0.157$ \\
Diffusion & ID & Pooled & $0.143\pm0.152$ & $0.763\pm0.222$ \\
Diffusion & OOD & ScanNet++ & $0.097\pm0.102$ & $0.832\pm0.124$ \\
Flow Matching & ID & 3RScan & $0.206\pm0.161$ & $0.640\pm0.266$ \\
Flow Matching & ID & ARKit & $0.298\pm0.134$ & $0.487\pm0.206$ \\
Flow Matching & ID & MP3D & $0.109\pm0.123$ & $0.824\pm0.169$ \\
Flow Matching & ID & ScanNet & $0.156\pm0.125$ & $0.757\pm0.157$ \\
Flow Matching & ID & ZInD & $0.094\pm0.122$ & $0.853\pm0.167$ \\
Flow Matching & ID & Pooled & $0.143\pm0.148$ & $0.766\pm0.226$ \\
Flow Matching & OOD & ScanNet++ & $0.101\pm0.103$ & $0.832\pm0.135$ \\
FM+XAttn & ID & 3RScan & $0.194\pm0.156$ & $0.642\pm0.256$ \\
FM+XAttn & ID & ARKit & $0.275\pm0.125$ & $0.495\pm0.193$ \\
FM+XAttn & ID & MP3D & $0.101\pm0.112$ & $0.819\pm0.154$ \\
FM+XAttn & ID & ScanNet & $0.146\pm0.116$ & $0.748\pm0.144$ \\
FM+XAttn & ID & ZInD & $0.089\pm0.115$ & $0.855\pm0.162$ \\
FM+XAttn & ID & Pooled & $0.133\pm0.138$ & $0.766\pm0.218$ \\
FM+XAttn & OOD & ScanNet++ & $0.095\pm0.093$ & $0.813\pm0.123$ \\
\bottomrule
\end{tabular*}
\end{table}

\subsection{End-to-End Evaluation Design}
\label{supp:e2e_multi_source}
The RGB-to-BEV track uses image-derived conditioning from \cref{supp:camera_pipeline} and reuses the source-geometry-trained completion weights without fine-tuning. Both tracks share the BEV interface and metrics (\cref{sec:evaluation,tab:main_results_ioubest,tab:sampling_results}).
The end-to-end cohort contains $N=1{,}000$ egocentric frames, with $200$ frames from each of ScanNet, ARKitScenes, Matterport3D, 3RScan, and ScanNet++; ZInD panoramas are excluded from this pinhole-camera protocol.
Frames come from real captured camera poses rather than synthesized viewpoints. The four ID sources contribute frames only from held-out test scenes, disjoint from completion-model training and validation scenes. ScanNet++ provides held-out-source evaluation within this RGB cohort.
The evaluation retains observations satisfying the conditioning-signal threshold $\CondSignalRatio \ge 0.10$, computed using ground truth. The geometric frame-validity filter in \cref{supp:camera_pipeline} applies at test time.
\Cref{tab:supp:e2e_fidelity} reports the quantitative results; \cref{fig:qualitative_supp_b} shows the associated map structure and boundary errors.

\begin{table}
\centering
\normalsize
\captionsetup{font=normalsize}
\setlength{\tabcolsep}{1.5pt}
\renewcommand{\arraystretch}{1}
\caption{Extended RGB-to-BEV completion results ($N{=}1{,}000$; \cref{tab:rgb_to_bev}). Entries with dispersion show observation-level mean$\pm$SD. $\mathrm{IoU}_1$ evaluates a single output: the first sample for generators and the thresholded mean map for LaMa-Ensemble, which can outperform every individual member. $\mathrm{IoU}_b$ is ground-truth-selected best-of-four IoU. $\mathrm{IoU}_m$ averages four separately computed IoUs; MES and Var use all four outputs, and Var averages the per-cell variance of the four binary outputs over $\MaskEval$. A dash marks a metric omitted from this table.}
\label{tab:supp:e2e_fidelity}
\begin{tabular*}{\linewidth}{@{\extracolsep{\fill}}lccccc@{}}
\toprule
Method & IoU$_1\uparrow$ & IoU$_b\uparrow$ & MES$\downarrow$ & IoU$_m\uparrow$ & Var \\
\midrule
All-Obstacle & $0.000\pm0.000$ & -- & -- & -- & -- \\
All-Floor & $0.441\pm0.228$ & -- & -- & -- & -- \\
Uniform Random & $0.329\pm0.119$ & -- & -- & -- & -- \\
NN Propagation & $0.248\pm0.175$ & -- & -- & -- & -- \\
\midrule
U-Net & $\mathbf{0.564\pm0.195}$ & -- & -- & -- & -- \\
PConv-UNet & $0.563\pm0.198$ & -- & -- & -- & -- \\
LaMa & $0.148\pm0.153$ & -- & -- & -- & -- \\
\midrule
LaMa-Ensemble & $0.439\pm0.236$ & $0.414$ & $0.302\pm0.147$ & $0.233\pm0.188$ & $0.141\pm0.063$ \\
Diffusion & $0.497\pm0.214$ & $0.522$ & $\mathbf{0.293\pm0.136}$ & $0.504\pm0.217$ & $0.047\pm0.029$ \\
Flow Matching & $0.542\pm0.190$ & $0.564$ & $0.306\pm0.141$ & $0.552\pm0.192$ & $0.038\pm0.040$ \\
FM+XAttn & $0.560\pm0.177$ & $\mathbf{0.565}$ & $0.308\pm0.132$ & $\mathbf{0.562\pm0.185}$ & $0.041\pm0.022$ \\
\bottomrule
\end{tabular*}
\end{table}

\subsection{Sample Size and Sensitivity}
\label{supp:k_sensitivity}

We fix $\NumSamples{=}4$ for all stochastic methods (generators and LaMa-Ensemble), balancing estimation quality against compute while matching the ensemble member count.
The sensitivity protocol evaluates $K\in\{1,2,3,4\}$ predictions per observation.
\Cref{tab:supp:k_sensitivity_iou} reports best-of-$K$ IoU on both ID and OOD splits. The $K=1$ column reports single-output fidelity; the larger-$K$ columns report ground-truth-selected candidate quality.

\begin{table*}
\centering
\normalsize
\captionsetup{font=normalsize}
\setlength{\tabcolsep}{1.5pt}
\renewcommand{\arraystretch}{1}
\caption{Sample-count sensitivity ($K\in\{1,2,3,4\}$) on ID and OOD: observation-level mean$\pm$SD of best-of-$K$ IoU ($\uparrow$). The $K=1$ column gives single-output fidelity, which for LaMa-Ensemble is its mean map. For $K>1$, ground truth selects the highest-IoU prediction among the first $K$ candidates; LaMa-Ensemble's candidates are individual members in seed order $41,42,43,44$. These scores give an oracle upper bound for selection from those candidates.}
\label{tab:supp:k_sensitivity_iou}
\begin{tabular*}{\linewidth}{@{\extracolsep{\fill}}ll cccc@{}}
\toprule
Method & Split & $K{=}1$ & $K{=}2$ & $K{=}3$ & $K{=}4$ \\
\midrule
LaMa-Ensemble & ID & $0.735\pm0.304$ & $0.787\pm0.236$ & $0.806\pm0.214$ & $0.817\pm0.205$ \\
LaMa-Ensemble & OOD & $0.853\pm0.153$ & $0.867\pm0.130$ & $0.871\pm0.127$ & $0.874\pm0.126$ \\
Diffusion & ID & $0.699\pm0.207$ & $0.796\pm0.220$ & $0.819\pm0.205$ & $0.830\pm0.196$ \\
Diffusion & OOD & $0.734\pm0.148$ & $0.866\pm0.122$ & $0.879\pm0.123$ & $0.890\pm0.122$ \\
Flow Matching & ID & $0.722\pm0.233$ & $0.792\pm0.223$ & $0.817\pm0.204$ & $0.830\pm0.192$ \\
Flow Matching & OOD & $0.771\pm0.151$ & $0.862\pm0.127$ & $0.874\pm0.127$ & $0.880\pm0.129$ \\
FM+XAttn & ID & $0.716\pm0.222$ & $0.790\pm0.214$ & $0.818\pm0.201$ & $0.830\pm0.192$ \\
FM+XAttn & OOD & $0.725\pm0.144$ & $0.842\pm0.120$ & $0.862\pm0.121$ & $0.873\pm0.124$ \\
\bottomrule
\end{tabular*}
\end{table*}

\subsection{Structurally Challenging Subset}
\label{supp:suite_hard}

Floor prevalence measures class balance on $\MaskEval$; $\CondSignalRatio$ measures the observed fraction of target floor.
The \emph{structurally challenging} subset requires both low conditioning
signal ($\CondSignalRatio\le0.20$; Learnable tier, \cref{supp:learnability})
and low floor prevalence (floor fraction on $\MaskEval$ below $0.50$).
Thus, at least $80\%$ of target-floor cells are unobserved, and non-floor
cells outnumber floor cells in the evaluation region.
This yields $N{=}1{,}386$ observations
(mean $\CondSignalRatio{=}0.14$, floor prevalence~$0.19$).
The subset reuses the canonical-track test predictions.

\Cref{tab:challenging_main} gives the main comparison; \cref{tab:supp:hard_fidelity} also reports standard deviations. All-Floor IoU is $0.191$, U-Net obtains $0.468$, and FM+XAttn has the largest best-of-four IoU ($0.479$) and lowest MES ($0.316$) among the sampled methods.

\begin{table}
\centering
\normalsize
\captionsetup{font=normalsize}
\setlength{\tabcolsep}{1.5pt}
\renewcommand{\arraystretch}{1}
\caption{Structurally challenging subset (mean$\pm$SD, $N{=}1{,}386$): $\CondSignalRatio\leq0.20$ and floor prevalence $<0.50$ on $\MaskEval$. IoU evaluates single outputs for $K=1$ and ground-truth-selected best-of-four predictions for $K=4$. $\mathrm{IoU}_m$ averages four IoUs; MES and Var use all four outputs, and Var averages per-cell binary variance over $\MaskEval$. UMR and pooled F1 are reported in \cref{tab:challenging_main}, where LaMa-Ensemble uses its thresholded mean map.}
\label{tab:supp:hard_fidelity}
\begin{tabular*}{\linewidth}{@{\extracolsep{\fill}}lc cccc@{}}
\toprule
Method & $K$ & IoU$\uparrow$ & MES$\downarrow$ & IoU$_m\uparrow$ & Var \\
\midrule
All-Floor & 1 & $0.191\pm0.168$ & -- & -- & -- \\
NN Propagation & 1 & $0.192\pm0.153$ & -- & -- & -- \\
\midrule
U-Net & 1 & $0.468\pm0.193$ & -- & -- & -- \\
PConv-UNet & 1 & $0.458\pm0.198$ & -- & -- & -- \\
LaMa & 1 & $0.317\pm0.187$ & -- & -- & -- \\
\midrule
LaMa-Ensemble & 4 & $0.397\pm0.231$ & $0.362\pm0.158$ & $0.295\pm0.216$ & $0.174\pm0.072$ \\
Diffusion & 4 & $0.462\pm0.198$ & $0.352\pm0.153$ & $0.396\pm0.202$ & $0.055\pm0.034$ \\
Flow Matching & 4 & $0.464\pm0.195$ & $0.346\pm0.148$ & $0.394\pm0.199$ & $0.040\pm0.028$ \\
FM+XAttn & 4 & $\mathbf{0.479\pm0.191}$ & $\mathbf{0.316\pm0.139}$ & $\mathbf{0.417\pm0.192}$ & $0.046\pm0.031$ \\
\bottomrule
\end{tabular*}
\par\smallskip
ID\,=\,1{,}221, OOD\,=\,165.
\end{table}

\subsection{Offline Planning Protocol}
\label{supp:applications}

\noindent\textbf{Task construction and common planner.}
The evaluation in \cref{sec:route_safety} uses $4{,}000$ point-goal tasks from $500$ RGB-to-BEV test scenes: $250$ ID and $250$ held-out-source scenes, with eight tasks per scene. This planning cohort is separate from the $1{,}000$-frame completion cohort in \cref{supp:e2e_multi_source}.

Starts are sampled on observed floor cells, and every planner treats the goal cell as traversable. Valid-support masks constrain all planners. All methods use Dijkstra's algorithm on an 8-connected grid, with map-dependent admissibility and traversal costs defined below; diagonal corner cutting is disallowed. The robot is represented by a point, with no footprint inflation; a route is feasible if it passes only through valid floor cells.

\phantomsection
\noindent\textbf{Map-dependent planning costs.}
\label{supp:planning_cost}
The planning comparison measures how completed maps support routes between supplied endpoints. Thresholded and probabilistic occupancy maps provide different ways to use uncertain geometry~\citep{banfiworth}. We combine path length with a clipped negative-log floor-support penalty, defined below.

For Diffusion, Flow Matching, and FM+XAttn, four binary completions define the empirical floor frequency $p_{\mathrm{floor}}(x)=\tfrac14\sum_{k=1}^{4}b_k(x)$. Besides the always-traversable goal cell, their risk-weighted planners admit cells with $p_{\mathrm{floor}}(x)>0$. Unobserved cells have cost $c(x)=1-\log(\max\{p_{\mathrm{floor}}(x),0.5\})$, and observed floor has $c(x)=1$. Each allowed move $x\to y$ has positive edge weight $w(x,y)=\lVert y-x\rVert_2 c(y)$. Dijkstra therefore minimizes the following objective over admissible paths $\pi=(x_0,\ldots,x_T)$ within valid support:
\begin{equation}
J(\pi)=\sum_{t=1}^{T}\lVert x_t-x_{t-1}\rVert_2\,c(x_t),
\label{eq:planning_path_cost}
\end{equation}
with coordinates in grid cells: axial moves have length $1$ and diagonals $\sqrt{2}$. The FM+XAttn mean-map comparator admits cells with $p_{\mathrm{floor}}\geq0.5$ and uses unit cell costs, giving ordinary shortest-path search on the thresholded map.

The binary-map planners (All-Floor, U-Net, and PConv-UNet) admit floor-labeled cells at the same threshold, clamp observed evidence, and use unit cell costs. They share the search algorithm, connectivity, corner-cutting rule, valid-support constraint, and endpoint handling specified above.

\noindent\textbf{Parameter roles.}
The budget $K=4$ matches completion evaluation (\cref{supp:k_sensitivity}), giving floor frequencies $\{0,\tfrac14,\tfrac12,\tfrac34,1\}$. The mean-map threshold requires at least two votes; positive-probability admission requires one. Clipping at $0.5$ caps the added penalty at $\log 2$: one-vote and two-vote cells cost $1+\log 2$, and unanimous floor costs $1$. The unit term charges for path length; the log term penalizes lower predicted floor support. All generators share the unit offset and clipping threshold. The resulting traversal penalty ranks routes without assigning calibrated probabilities of whole-path success. Eight-connected search permits diagonals, while the corner-cutting rule excludes diagonal moves that pass between two blocked cells. Safety is evaluated against ground truth.

\noindent\textbf{Route evaluation.}
``Found'' is the percentage of tasks with a returned path (\cref{tab:route_safety}). Safe Success requires every visited cell to be valid ground-truth floor. SPL~\citep{Anderson2018Evaluation} weights success by geometric path efficiency:
\[
\mathrm{SPL}=\frac{1}{N}\sum_{i=1}^{N}S_i\frac{\ell_i}{\max\{p_i,\ell_i\}}.
\]
Here $S_i$ indicates Safe Success, $\ell_i$ is the ground-truth shortest-path length, and $p_i$ is the returned-path length; failed tasks contribute zero. Risk-weighted FM+XAttn achieves $3.73$ percentage points higher Safe Success and $5.38$ percentage points higher Found than its mean-map comparator.

\subsection{Boundary Radius and Sample Disagreement}
\label{supp:boundary_radius}

For the boundary-variance analysis, the spatial regions depend on a radius~$r$ (in pixels).
Given~$\FloorGT$, we extract the 1\,px-wide floor edge via morphological erosion ($3{\times}3$) and subtraction, then dilate with a $(2r{+}1){\times}(2r{+}1)$ square kernel to form~$\Omega_{\mathrm{bnd}}$ (intersected with~$\MaskUnknown$).
Unobserved floor cells surviving erosion by the same kernel form~$\Omega_{\mathrm{int}}$. The boundary band spans the target floor edge and its neighborhood on both sides. Both regions exclude observed cells, so their averages describe variation in the completed part of the map.
We use $r=7$ pixels, corresponding to approximately $0.27$\,m along each grid axis at the canonical scale. This radius provides $10^3$--$10^4$ boundary cells per scene. The radius sweep gives an ensemble-to-FM+XAttn interior variance ratio of $700$--$1000$ for $r\in\{5,\ldots,9\}$ and a boundary ratio of $2$--$4$. A wider band ($r=15$) reduces the observed boundary--interior contrast.
The square dilation measures distance in the grid's maximum-coordinate norm.

\noindent\textbf{Interior and boundary disagreement.}
Binary sample variance uses divisor four (\cref{eq:binary_variance}). We call the mean per-cell variance over $\Omega_{\mathrm{int}}$ the \emph{interior disagreement}, and over $\Omega_{\mathrm{bnd}}$ the \emph{boundary disagreement}. FM+XAttn has interior disagreement $6\times10^{-5}$ and boundary disagreement $0.037$; the LaMa-Ensemble values are $0.052$ and $0.125$.
FM+XAttn therefore shows almost no interior disagreement, whereas LaMa-Ensemble's members disagree inside floor interiors and also show greater boundary disagreement (\cref{fig:lama_ens_comparison}).

\phantomsection
\noindent\textbf{Boundary-band and interior IoU.}
\label{supp:boundary_iou}
Boundary-sensitive evaluation distinguishes local transitions from floor interiors~\citep{boundary}. \Cref{tab:supp:boundary_iou_full} evaluates IoU within the ground-truth boundary band and within the eroded floor interior defined above, each restricted to $\MaskEval$. For either region $\Omega$, the score is $|\FloorComp\cap\FloorGT\cap\Omega|/|(\FloorComp\cup\FloorGT)\cap\Omega|$, using the predicted and target floor sets.

The stratified canonical test sample contains $1{,}710$ ID and $1{,}984$ OOD observations (seed $42$). Single-output methods use one prediction. For stochastic methods, full-region IoU selects the best of four; the same selected prediction is scored on both regions.

Diffusion has the highest boundary-band IoU on ID ($0.751$) and OOD ($0.822$); FM+XAttn scores $0.750/0.804$. LaMa scores $0.605/0.580$. The interior contains only ground-truth floor cells, so All-Floor attains IoU $1.000$ there by construction; lower scores indicate omitted floor within that region.

On OOD boundary-band IoU, All-Floor reaches $0.820$ against best-of-four Diffusion's $0.822$ and single-output U-Net's $0.809$. Learned-method interior IoUs range from $0.847$ to $0.982$ on ID and $0.744$ to $0.991$ on OOD.

\begin{table}
\centering
\normalsize
\captionsetup{font=normalsize}
\setlength{\tabcolsep}{1.5pt}
\renewcommand{\arraystretch}{1}
\caption{Boundary-band and interior IoU on the stratified canonical sample ($1{,}710$ ID and $1{,}984$ OOD observations). The boundary band uses radius $r=7$ pixels. Interior columns report means; boundary columns report mean$\pm$SD. The top five methods use single outputs; the bottom four use best-of-four selection by full-region IoU. OOD is ScanNet++. Bold marks the highest boundary-band IoU per split.}
\label{tab:supp:boundary_iou_full}
\begin{tabular*}{\linewidth}{@{\extracolsep{\fill}}lcccc@{}}
\toprule
& \multicolumn{2}{c}{ID} & \multicolumn{2}{c}{OOD} \\
\cmidrule(lr){2-3}\cmidrule(lr){4-5}
Method & Interior$\uparrow$ & Boundary$\uparrow$ & Interior$\uparrow$ & Boundary$\uparrow$ \\
\midrule
All-Floor & 1.000 & $0.731\pm0.171$ & 1.000 & $0.820\pm0.131$ \\
NN Propagation & 0.930 & $0.604\pm0.191$ & 0.917 & $0.639\pm0.186$ \\
U-Net & 0.974 & $0.744\pm0.154$ & 0.972 & $0.809\pm0.124$ \\
PConv-UNet & 0.971 & $0.744\pm0.156$ & 0.975 & $0.812\pm0.126$ \\
LaMa & 0.847 & $0.605\pm0.174$ & 0.744 & $0.580\pm0.185$ \\
\midrule
LaMa-Ensemble & 0.979 & $0.735\pm0.167$ & 0.991 & $0.801\pm0.140$ \\
Diffusion & 0.982 & $\mathbf{0.751\pm0.148}$ & 0.984 & $\mathbf{0.822\pm0.121}$ \\
Flow Matching & 0.972 & $0.740\pm0.158$ & 0.974 & $0.808\pm0.129$ \\
FM+XAttn & 0.975 & $0.750\pm0.151$ & 0.964 & $0.804\pm0.131$ \\
\bottomrule
\end{tabular*}
\end{table}

\subsection{Constructed Multi-Solution Example}
\label{supp:synthetic_ambiguity}

To illustrate structural ambiguity in BEV completion, we construct a single conditioning input with multiple valid ground-truth completions (\cref{fig:inverse_problem}). We select five observations from different scenes across the dataset whose floormaps overlap in a shared region of the BEV grid. The shared conditioning map $\FloorObs^{\text{syn}}$ is the intersection of the five observed floormaps, and the maps shown in \cref{fig:inverse_problem} are four of the five real ground-truth floormaps, not model outputs.

\noindent\textbf{Construction.}
Let the five selected observations be indexed by $i\in\{1,\dots,5\}$. We combine them into a single test instance:
\begin{align}
\FloorObs^{\text{syn}} &= \bigcap_i \FloorObs^{(i)}, &
\MaskValid^{\text{syn}} &= \bigcap_i \MaskValid^{(i)}, \\
\MaskUnknown^{\text{syn}} &= \Bigl(\bigcup_i \MaskUnknown^{(i)}\Bigr)\cup\Delta_d, &
\MaskEval^{\text{syn}} &= \MaskUnknown^{\text{syn}}\odot\MaskValid^{\text{syn}},
\end{align}
where $\Delta_d$ contains $473$ cells that are observed in all five inputs but have conflicting floor labels; promoting them to the unobserved set makes the shared conditioning consistent with every candidate target. The resulting regions contain $|\FloorObs^{\text{syn}}|=5{,}998$, $|\MaskValid^{\text{syn}}|=11{,}124$, and $|\MaskEval^{\text{syn}}|=5{,}126$ cells. We denote the five candidate targets by $\{G^{(j)}\}_{j=1}^{5}$; each is the corresponding floormap restricted to $\MaskValid^{\text{syn}}$. Pairwise target IoU on $\MaskEval^{\text{syn}}$ ranges from $0.719$ to $0.948$, with $268$ to $1{,}438$ disagreeing cells per pair, confirming genuine multi-modality rather than annotation noise. By construction, every candidate target matches the shared conditioning on observed valid cells.

\noindent\textbf{Distributional metrics.}
Given $K$ model samples $\{Y^{(k)}\}_{k=1}^K$ and the five candidate targets $\{G^{(j)}\}_{j=1}^{5}$, we define the IoU distance
$d(A,B)=1-\mathrm{IoU}(A\odot\MaskEval^{\text{syn}},\,B\odot\MaskEval^{\text{syn}})$
and report symmetric Chamfer distance in IoU space~\citep{Barrow1977Chamfer}:
\begin{align*}
d_{\text{p}\to\text{g}} &= \frac{1}{K}\sum_{k=1}^{K}\min_{1\leq j\leq5} d(Y^{(k)},G^{(j)}), \\
d_{\text{g}\to\text{p}} &= \frac{1}{5}\sum_{j=1}^{5}\min_{1\leq k\leq K} d(G^{(j)},Y^{(k)}), \\
d_{\text{sym}} &= \tfrac12\left(d_{\text{p}\to\text{g}}+d_{\text{g}\to\text{p}}\right).
\end{align*}
The first term measures each prediction's agreement with its nearest candidate target; the second measures how well the prediction set represents each target. All distances use the evaluation mask $\MaskEval^{\text{syn}}$.

\textbf{Coverage} is the fraction of candidate targets matched by at least one prediction at distance strictly below $0.1$:
\[
\mathrm{coverage}=\frac15\sum_{j=1}^{5}\mathbf{1}\!\left\{\min_{1\leq k\leq K}d(G^{(j)},Y^{(k)})<0.1\right\}.
\]
\textbf{Diversity} averages pairwise distances between distinct predictions:
\[
\mathrm{diversity}=\frac{2}{K(K-1)}\sum_{1\leq k<\ell\leq K}d(Y^{(k)},Y^{(\ell)}),\qquad K>1.
\]
Diversity applies to $K>1$; single-prediction entries use a dash. One prediction may match several targets, so coverage need not count distinct recovered modes.

\begin{table}
\centering
\normalsize
\captionsetup{font=normalsize}
\setlength{\tabcolsep}{1.5pt}
\renewcommand{\arraystretch}{1}
\caption{Distributional evaluation on the constructed multi-solution example. $d_{\text{sym}}$: symmetric Chamfer distance in IoU space ($\downarrow$); $d_{\text{p}\to\text{g}}$: prediction-to-target distance, a precision measure ($\downarrow$); $d_{\text{g}\to\text{p}}$: target-to-prediction distance, a recall measure ($\downarrow$); coverage ($\uparrow$); diversity (descriptive). Bold marks the lowest symmetric distance and highest coverage among stochastic methods.}
\label{tab:supp:distributional}
\begin{tabular*}{\linewidth}{@{\extracolsep{\fill}}lccccc c@{}}
\toprule
Method & $K$ & $d_{\text{p}\to\text{g}}$ & $d_{\text{g}\to\text{p}}$ & $d_{\text{sym}} \downarrow$ & Cov.\,$\uparrow$ & Div. \\
\midrule
All-Obstacle         & 1 & 1.000 & 1.000 & 1.000 & 0.0 & --- \\
All-Floor        & 1 & 0.014 & 0.087 & 0.051 & 0.8 & --- \\
NN Propagation        & 1 & 0.013 & 0.088 & 0.051 & 0.8 & --- \\
U-Net           & 1 & 0.014 & 0.087 & 0.051 & 0.8 & --- \\
PConv-UNet      & 1 & 0.014 & 0.087 & 0.051 & 0.8 & --- \\
LaMa            & 1 & 0.147 & 0.279 & 0.213 & 0.0 & --- \\
Uniform Random   & 1 & 0.511 & 0.528 & 0.519 & 0.0 & --- \\
\midrule
LaMa-Ensemble       & 4 & 0.119 & 0.099 & 0.109 & 0.4 & 0.188 \\
Diffusion       & 4 & 0.021 & 0.087 & \textbf{0.054} & \textbf{0.8} & 0.013 \\
Flow Matching     & 4 & 0.020 & 0.087 & \textbf{0.054} & \textbf{0.8} & 0.012 \\
FM+XAttn           & 4 & 0.051 & 0.087 & 0.069 & \textbf{0.8} & 0.073 \\
\bottomrule
\end{tabular*}
\end{table}

\begin{figure}
\centering
\setlength{\tabcolsep}{4pt}%
\begin{tabular}{@{}c@{\hspace{2pt}}c@{\hspace{4pt}}c@{\hspace{4pt}}c@{\hspace{4pt}}c@{\hspace{8pt}}c@{}}
\footnotesize $\FloorObs$ &
\footnotesize $G^{(1)}$ &
\footnotesize $G^{(2)}$ &
\footnotesize $G^{(3)}$ &
\footnotesize $G^{(4)}$ &
\footnotesize Disagreement \\[3pt]
\begin{tikzpicture}[baseline=(obs.south),outer sep=0pt]
\node[inner sep=0pt,outer sep=0pt] (obs) {\includegraphics[width=1.8cm]{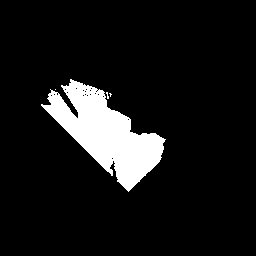}};
\node at ([yshift=0.45cm]obs.south) {\textcolor{yellow}{\small$\blacktriangledown$}};
\end{tikzpicture} &
\includegraphics[width=1.8cm]{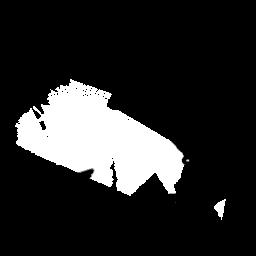} &
\includegraphics[width=1.8cm]{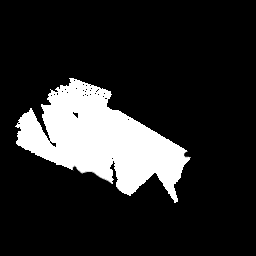} &
\includegraphics[width=1.8cm]{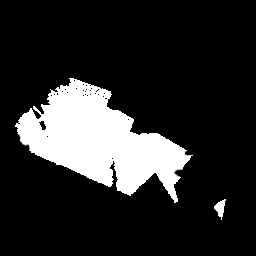} &
\includegraphics[width=1.8cm]{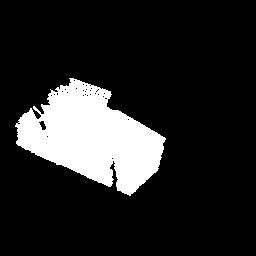} &
\includegraphics[width=1.8cm]{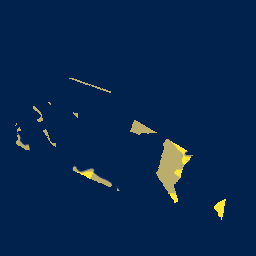}\\
\footnotesize Constructed \\
observation & \multicolumn{4}{c}{$\underbrace{\hspace{7.6cm}}_{\textstyle\text{candidate targets}}$} &
\scriptsize Low\,\includegraphics[width=0.85cm,height=0.12cm]{figures/cividis_cbar_horizontal.png}\,High \\
\end{tabular}
\caption{\textbf{Constructed multi-solution example.}
A shared partial BEV observation $\FloorObs$ is formed as the intersection of five observations from different scenes across the dataset; four of the five real ground-truth floormaps $G^{(j)}$ are shown. White denotes floor and black denotes non-floor. The yellow marker ($\blacktriangledown$) marks the camera.
The final panel shows relative per-cell disagreement among the candidate maps; its color scale is qualitative.}
\label{fig:inverse_problem}
\end{figure}

\noindent\textbf{Results.}
\Cref{tab:supp:distributional} summarizes performance on this multi-solution instance.
All-Floor, NN Propagation, U-Net, and PConv-UNet obtain $d_{\mathrm{sym}}=0.051$ with coverage $0.8$, matching four of the five candidate targets; U-Net and PConv-UNet predict floor on every evaluated cell here, so their rows equal All-Floor's. Because all five targets are mostly floor, a single all-floor map already lies within $0.1$ of four of them, so coverage on this instance cannot separate sample sets from single maps. The missed target is the most distinct one (pairwise IoU as low as $0.719$). LaMa deviates substantially ($d_{\mathrm{sym}}=0.213$, coverage $0.0$), matching no target under the $0.1$ threshold.

Among the stochastic methods, FM+XAttn has the highest diversity ($0.073$) of those reaching coverage $0.8$. Diffusion and Flow Matching achieve slightly lower $d_{\mathrm{sym}}$ ($0.054$ versus $0.069$) because their samples cluster near the dominant mode ($d_{\text{p}\to\text{g}}=0.020$--$0.021$), but their diversity is near zero ($0.012$--$0.013$): their samples are almost identical.

LaMa-Ensemble has high raw diversity ($0.188$) but lower coverage ($0.4$): its independently trained members produce disparate maps, some of which fall outside the match threshold, reflecting seed diversity rather than conditional sampling.

\section{Dataset Construction and Curation}
\label{supp:dataset_details}

\subsection{Source Datasets}
\label{supp:source_harmonization}

This section extends the construction summary in \cref{sec:dataset}. Six indoor datasets are harmonized into a shared metric BEV representation (\cref{tab:supp:sources,fig:dataset_and_split}).
For sources with semantic floor annotations (ScanNet, ScanNet++, Matterport3D, 3RScan), floor geometry is extracted directly; for the remainder (ARKitScenes, ZInD), the floor plane is estimated from the 15th percentile of vertex heights ($\pm 0.05$\,m tolerance).

\phantomsection\label{supp:zind_calibration}%
For ZInD, DA$^2$~\citep{li2025da2} supplies panoramic depth, which is rescaled to match the human-annotated room layout vertices in ZInD's metric floor-plan frame before point-cloud floor extraction. The annotations determine the metric extent of each target floormap; the depth model supplies within-room structure beyond the layout vertices, including geometry that occludes floor.

\begin{table}[htbp]
\centering
\normalsize
\captionsetup{font=normalsize}
\setlength{\tabcolsep}{1.5pt}
\renewcommand{\arraystretch}{1}
\caption{Source datasets in the \dataset benchmark. Upstream counts are before filtering. Per-dataset result breakdowns appear in \cref{tab:supp:per_dataset_breakdown,tab:supp:per_dataset_stochastic}.}
\label{tab:supp:sources}
\begin{tabular*}{\linewidth}{@{\extracolsep{\fill}}lrrlll@{}}
\toprule
\textbf{Dataset} & \textbf{Scenes} & \textbf{Obs.} & \textbf{Format} & \textbf{Floor source} & \textbf{Axis transform} \\
\midrule
3RScan        & 1{,}291  & 30{,}984   & OBJ mesh   & Semantic labels     & Flip $z\!\leftarrow\!{-z}$ \\
ScanNet       & 1{,}508  & 36{,}192   & PLY mesh   & Semantic labels     & Identity \\
ARKitScenes   & 4{,}803  & 115{,}271  & PLY mesh   & Height percentile   & Swap Y/Z \\
Matterport3D  & 2{,}101  & 50{,}424   & PLY mesh   & Semantic labels     & Identity \\
ZInD          & 7{,}026  & 168{,}624  & PLY points & Height percentile   & Metric calibration \\
\midrule
ScanNet++ (OOD) & 927    & 22{,}248   & PLY mesh   & Semantic labels     & Identity \\
\midrule
\textbf{Total} & \textbf{17{,}656} & \textbf{423{,}743} & & & \\
\bottomrule
\end{tabular*}
\end{table}

\begin{figure}[!htbp]
\centering
\resizebox{0.5\linewidth}{!}{
\begin{tikzpicture}
  \definecolor{pieA}{HTML}{2D7D9A}  
  \definecolor{pieB}{HTML}{E07A3A}  
  \definecolor{pieC}{HTML}{7CB342}  
  \definecolor{pieD}{HTML}{AB47BC}  
  \definecolor{pieE}{HTML}{F4C542}  
  \definecolor{pieF}{HTML}{EF5350}  
  \pie[
    text=pin,
    pin distance=0.6cm,
    every pin/.style={font=\tiny},
    before number={\color{white}\tiny\bfseries},
    radius=1.55,
    color={pieA, pieB, pieC, pieD, pieE, pieF},
    explode={0,0,0,0,0,0.08},
    after number={\,\%},
    sum=auto
  ]{
    49.2/ZInD,
    14.9/ARKit,
    14.0/MP3D,
    9.2/ScanNet,
    6.7/3RScan,
    6.0/ScanNet++
  }
\end{tikzpicture}

}
\caption{Breakdown of the 270{,}575 filtered observations across the six indoor source datasets. Source counts and assets are listed in \cref{tab:dataset_sources}; construction appears in \cref{fig:data_observation}.}
\label{fig:dataset_and_split}
\end{figure}

\begin{figure}[!t]
\centering
\begin{tikzpicture}[
  font=\sffamily\small,
  stage/.style={
    draw=black!50, thick, rounded corners=3pt,
    minimum width=60mm, minimum height=9.5mm,
    align=center, inner sep=4pt, fill=#1},
  stage/.default={white},
  stlbl/.style={
    font=\sffamily\scriptsize\bfseries, text=black!55,
    minimum width=8mm, anchor=east},
  badge/.style={
    font=\sffamily\scriptsize\bfseries, rounded corners=2pt,
    inner sep=2.5pt, fill=black!6, text=black!75},
  drop/.style={
    font=\sffamily\scriptsize, text=black!50, align=left, inner sep=2pt},
  arr/.style={-{Stealth[length=2.5mm, width=1.8mm]}, thick, black!55},
  droparr/.style={-{Stealth[length=2mm, width=1.4mm]}, thin, black!30, densely dashed},
  splitbox/.style={
    draw=black!30, thin, rounded corners=2pt,
    minimum height=7mm, inner sep=3pt,
    font=\sffamily\scriptsize, fill=black!3, text=black!65}
]

\node[stage=blue!6]  (s0) {Readable metric data};
\node[stlbl] at ([xshift=-2mm]s0.west) {\textsf{S\,0}};

\node[stage=blue!8, below=14mm of s0]  (s1) {Curation \& crop validation};
\node[stlbl] at ([xshift=-2mm]s1.west) {\textsf{S\,1}};

\node[stage=green!8, below=14mm of s1] (s2) {Strict canonical release};
\node[stlbl] at ([xshift=-2mm]s2.west) {\textsf{S\,2}};

\node[badge, right=3mm of s0.east] {423{,}743};
\node[badge, right=3mm of s1.east] {391{,}024};
\node[badge, right=3mm of s2.east] {\textbf{270{,}575}};

\draw[arr] (s0) -- (s1);
\draw[arr] (s1) -- (s2);

\node[drop, left=16mm of s0.south west, anchor=east] (d1)
  {{\sffamily\scriptsize\bfseries\color{red!55!black}$-$32{,}719}\\\textit{support \& crop filtering}};
\draw[droparr] ([yshift=-3mm]s0.west) -- (d1.east);

\node[drop, left=16mm of s1.south west, anchor=east] (d2)
  {{\sffamily\scriptsize\bfseries\color{red!55!black}$-$120{,}449}\\\textit{$\CondSignalRatio{<}0.1$}};
\draw[droparr] ([yshift=-3mm]s1.west) -- (d2.east);

\node[below=8mm of s2, font=\sffamily\scriptsize, text=black!55] (split) {%
  \begin{tabular}{@{}c@{\;\;}c@{\;\;}c@{}}
  \textbf{Train} & \textbf{Val} & \textbf{Test}\\
  215{,}342 & 26{,}890 & 28{,}343
  \end{tabular}};
\draw[thin, black!25] ([yshift=-2mm]s2.south west) -- ([yshift=-2mm]s2.south east);

\end{tikzpicture}
\caption{\textbf{Dataset curation pipeline.}
Each stage applies deterministic quality filters, retaining observations from all six source datasets.
Dashed branches indicate removed observations.
The strict $\CondSignalRatio{\ge}0.1$ gate accounts for 79\% of all removals.
Crop validation excludes one observation.}
\label{fig:supp:pipeline}
\end{figure}

\subsection{BEV Window and Coordinates}
\label{supp:delta_choice}

Geometry is expressed in an egocentric coordinate frame aligned with the floor plane. The model grid is $256{\times}256$ at $25.6$ pixels per meter ($\CellSize=0.0390625$\,m per pixel), covering a $10{\times}10$\,m window.

The camera is located at pixel $(128,192)$, with forward corresponding to decreasing row index. The window therefore extends $5$\,m to either side, $7.5$\,m forward, and $2.5$\,m behind the camera. The observed floor, complete floor target, unobserved-region mask, and validity mask share this spatial alignment. Maps are downsampled from the $512{\times}512$ synthesis canvas by average pooling and then binarized.

\subsection{Observation Synthesis}
\label{supp:observation_synthesis}

Observation synthesis comprises five processing steps:
\begin{enumerate}
    \item Floor extraction (semantic labels or 15th-percentile height; see \cref{supp:source_harmonization}).
    \item Removal of geometry above $z_{\mathrm{floor}} + 1.25$\,m (robot ceiling).
    \item Camera sampling on floor-valid positions (up to 24 observations per scene).
    \item Rasterization at $512{\times}512$ for four binary channels.
    \item Visibility reasoning using field-of-view and ray-occlusion tests.
\end{enumerate}

\noindent\textbf{Camera placement and parameters.}
Each scene contributes up to 24 observations. Positions are sampled via a spatial-coverage algorithm on a $0.4$\,m floor grid with center-biased weighting; headings are drawn uniformly from 36 discrete angles at $10^{\circ}$ increments.

The virtual sensor is a $90^{\circ}$-HFOV pinhole camera at $h{=}1.25$\,m above $z_{\mathrm{floor}}$ with a frame of $640\times480$. The horizontal field of view matches the Habitat navigation setup~\citep{Savva2019Habitat}; the height and frame size are specific to our pipeline.

In the synthesis-stage $512{\times}512$ BEV canvas, the camera sits at pixel $(256,384)$, horizontally centered and three quarters down the canvas, with the forward direction along decreasing row index.

Candidates are accepted only when valid cells cover at least $10\%$ of the canvas, the target contains at least $100$ floor pixels, and at least $50$ floor pixels are observed. Up to $500$ candidates are attempted per scene. These synthesis-stage coordinates specify visibility calculations; \cref{supp:delta_choice} gives the final model grid, metric window, and camera anchor.

\noindent\textbf{Multi-channel rasterization.}
Each accepted observation produces four aligned binary $512{\times}512$ synthesis maps:
\begin{itemize}
    \item \textbf{Floormap} ($\FloorGT$): complete floor support rasterized in BEV via polygon fill (mesh datasets) or point splatting with morphological closing (ZInD point clouds).
    \item \textbf{Validity mask} ($\MaskValid$): cells with defined scene support, including floor and non-floor labels, on which supervision and evaluation are valid. Visibility is represented separately by the observation and unobserved masks.
    \item \textbf{Observed floor} ($\FloorObs$): visibility-aware floor pixels, computed via z-buffer ray casting (256 samples per ray) that checks for occlusions from walls and furniture.
    \item \textbf{Unobserved} ($\MaskUnknown$): cells where the camera has no direct line of sight, because they are occluded or lie outside the FOV. Scoring intersects this mask with $\MaskValid$.
\end{itemize}

\subsection{Filtering, Crop Validation, and Label Balance}
\label{supp:filtering_stages}

\Cref{fig:supp:pipeline} summarizes the three-stage curation pipeline, which retains observations from all six source datasets.
Of the 423{,}743 synthesized observations, 391{,}024 survive curation and crop validation (train/val/test:\ 312{,}817/39{,}099/39{,}108), and 270{,}575 pass the strict canonical filter ($\CondSignalRatio \ge 0.10$; train/val/test:\ 215{,}342/\allowbreak 26{,}890/\allowbreak 28{,}343).

\noindent\textbf{Observation validity.}
Accepted observations have compatible masks, observed floor contained in target floor ($\FloorObs \subseteq \FloorGT$), nonempty evaluation support ($|\MaskEval| > 0$), and non-degenerate crops.

\begin{figure}[htbp]
\centering
\includegraphics[width=0.9\textwidth]{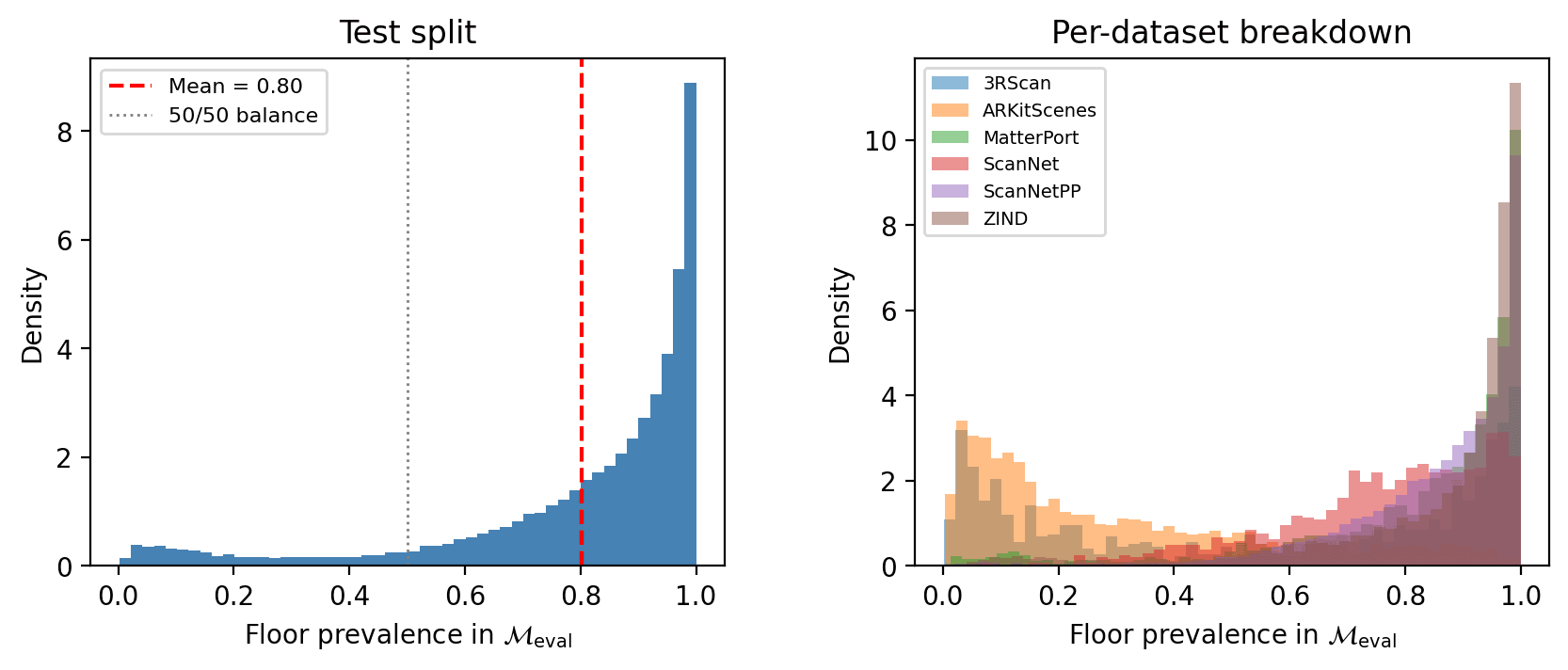}
\caption{Observation-pooled floor prevalence on $\MaskEval$ across the test observations (\cref{supp:filtering_stages}). \textbf{Left:} pooled prevalence density with the mean marked by a dashed red line. \textbf{Right:} per-source prevalence distributions.}
\label{fig:label_balance}
\end{figure}

\noindent\textbf{Label balance.}
Floor prevalence on $\MaskEval$, pooled across all $N=28{,}343$ test observations, is $0.802\pm0.238$ (mean $\pm$ standard deviation; \cref{fig:label_balance}). This floor dominance motivates separate reporting on subsets with lower floor prevalence (\cref{supp:suite_hard}). \Cref{tab:challenging_main} gives the corresponding fidelity and sample-set comparisons.

\subsection{Conditioning Signal and Learnability}
\label{supp:learnability}

The \emph{conditioning signal ratio}
$\CondSignalRatio = |\FloorObs|/|\FloorGT|$
measures the fraction of target floor cells observed, for $|\FloorGT|>0$.

We use three descriptive conditioning-signal tiers on the full $N{=}423{,}743$ upstream observations:

\begin{itemize}
  \item \textbf{Easy} ($\CondSignalRatio > 0.20$): strong conditioning signal. 158{,}435 observations (37.4\%).
  \item \textbf{Learnable} ($0.02 \le \CondSignalRatio \le 0.20$): moderate signal with substantial unobserved structure. 201{,}455 observations (47.5\%).
  \item \textbf{Negligible} ($\CondSignalRatio < 0.02$): negligible conditioning signal. 63{,}853 observations (15.1\%).
\end{itemize}

\begin{figure}[htbp]
    \centering
    \begin{minipage}[t]{0.48\linewidth}
        \vspace{0pt}
        \centering
        \includegraphics[width=\linewidth]{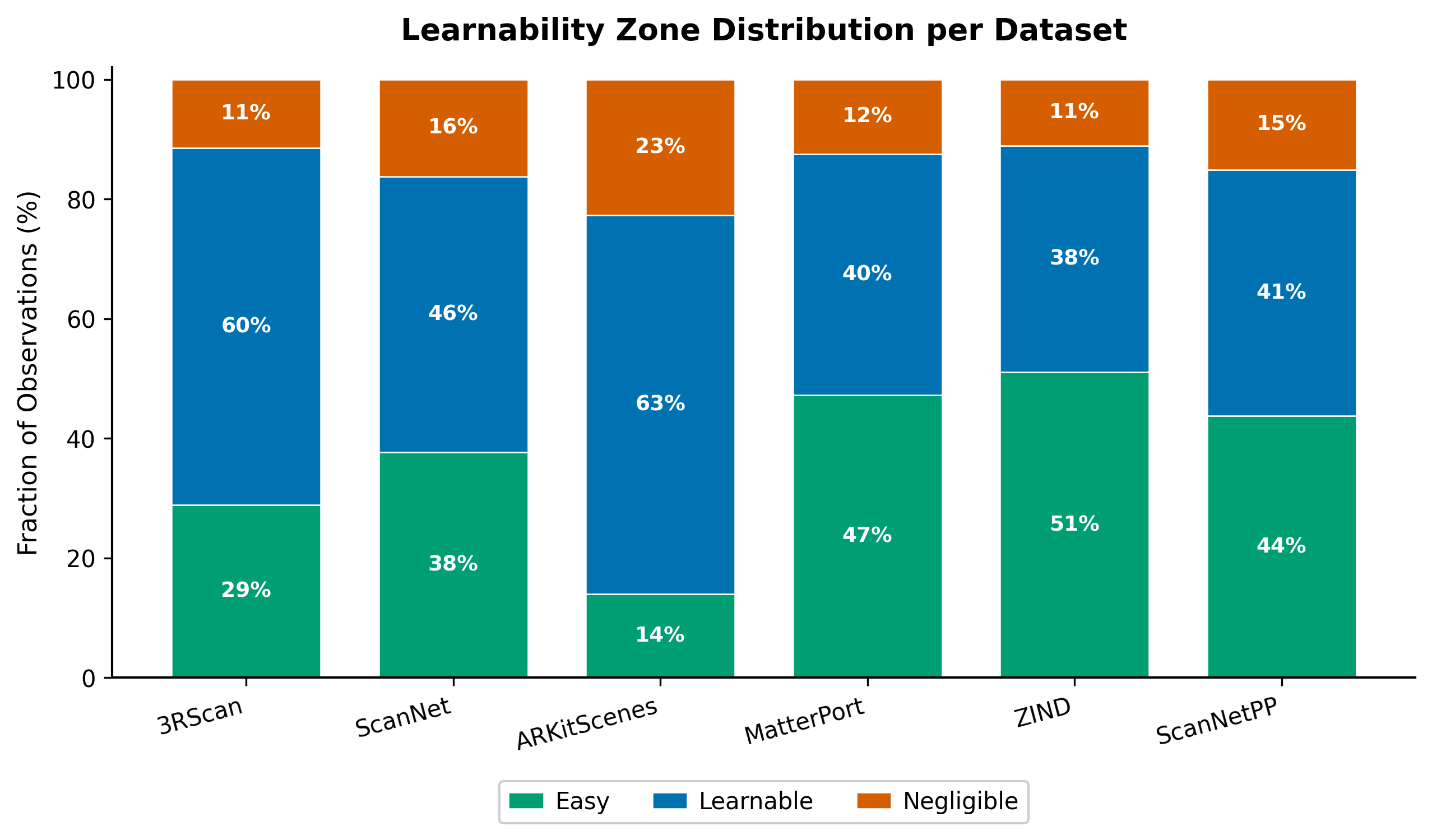}
        \caption{Conditioning-signal tiers across the upstream corpus.
        Three tiers are defined by $\CondSignalRatio$:
        \textbf{Easy} ($>0.20$),
        \textbf{Learnable} ($0.02\le\CondSignalRatio\le0.20$), and
        \textbf{Negligible} ($<0.02$).}
        \label{fig:learnability_zones}
    \end{minipage}\hfill%
    \begin{minipage}[t]{0.48\linewidth}
        \vspace{0pt}
        \centering
        \includegraphics[width=\linewidth]{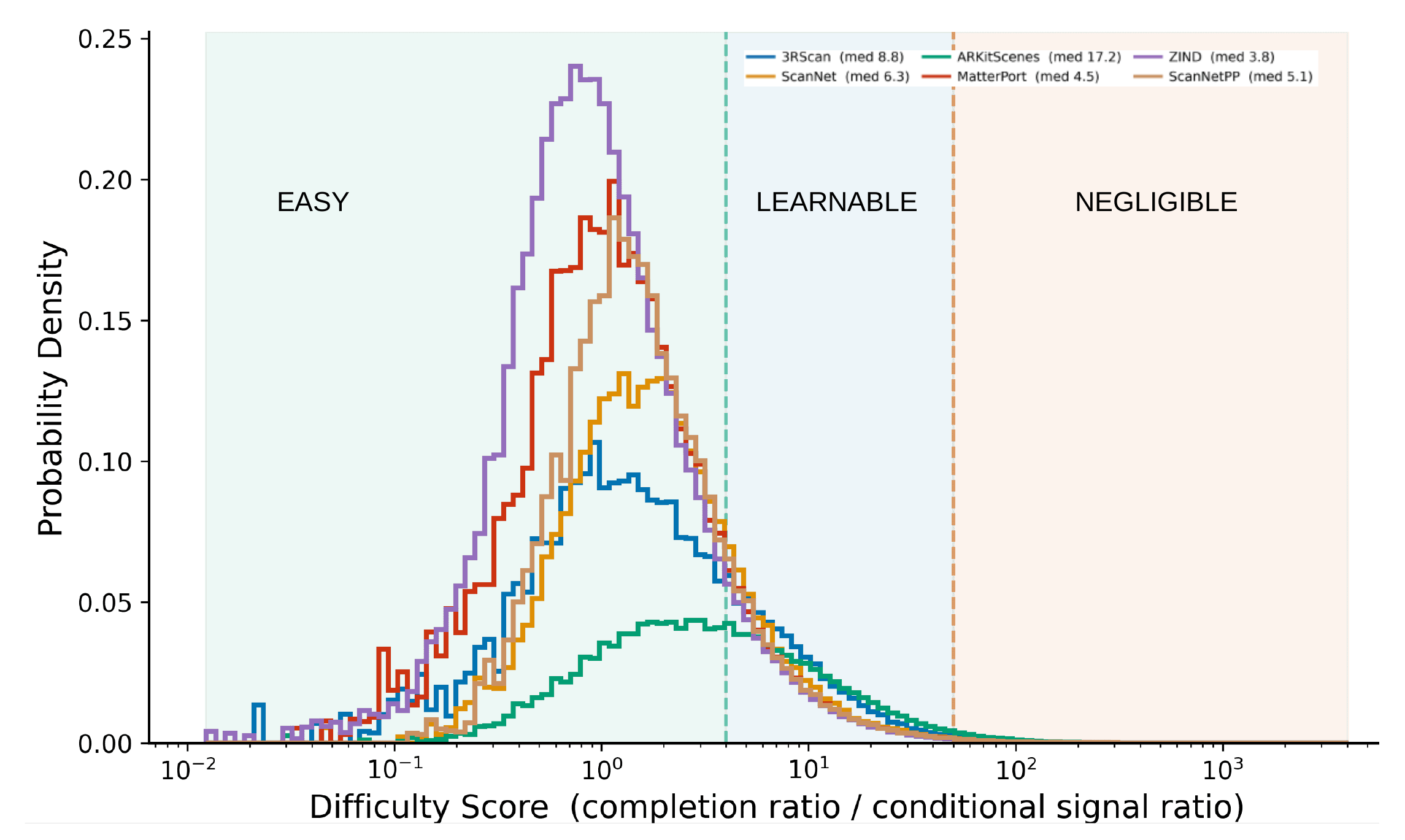}
        \caption{Difficulty-score distribution on a log scale,
        where $D = (1{-}\CondSignalRatio)/\CondSignalRatio$.
        Tier boundaries are $D<4$, $4\le D\le49$, and $D>49$.
        The logarithmic plot labels the upper Learnable cutoff as approximately $50$.
        The canonical filter retains $D\le9$.}
        \label{fig:difficulty_histogram}
    \end{minipage}
\end{figure}

The combined \textbf{Easy+Learnable} set comprises 84.9\% of observations.
\Cref{fig:learnability_zones} shows tier proportions by source.

\noindent\textbf{Difficulty score.}
A scalar difficulty $D = (1 - \CondSignalRatio)/\CondSignalRatio$, defined for $\CondSignalRatio>0$, gives the tier boundaries $D<4$ (Easy), $4\le D\le49$ (Learnable), and $D>49$ (Negligible).
\Cref{fig:difficulty_histogram} shows the distribution. The canonical filter $\CondSignalRatio \ge 0.1$ retains only $D\le9$.

\phantomsection
\noindent\textbf{Threshold selection.}
\label{supp:threshold_sensitivity}
The minimum conditioning signal $\CondSignalRatio\ge\tau$ is one of several curation thresholds. We use $\tau=0.10$, retaining 270{,}575 observations: $69.2\%$ of the 391{,}024 post-curation observations and $63.9\%$ of the 423{,}743 upstream observations.

The filter removes $120{,}449$ post-curation observations ($30.8\%$), all with less than $10\%$ of their target floor observed. Requiring $\CondSignalRatio>0.20$ would keep only Easy-tier observations, of which there are $158{,}435$ upstream ($37.4\%$, versus $63.9\%$ retained at $\tau=0.10$).

Manual inspection found near-degenerate low-signal views, including floor visible through narrow furniture gaps. The threshold balances conditioning strength and retained viewpoint diversity, including observations with limited visible floor. The ratio uses the complete target $\FloorGT$ for dataset selection and evaluation.

\subsection{Split Construction and Stratification}
\label{supp:split_consistency}
Viewpoints are grouped by source scene to prevent viewpoint leakage. The five ID sources use scene-level train/validation/test partitions; ScanNet++ supplies the held-out-source evaluation. Canonical counts are 215{,}342 training, 26{,}890 validation, and 28{,}343 test observations, with 12{,}129 ID and 16{,}214 OOD test observations.

\begin{figure}[!t]
\centering
\begin{minipage}[t]{0.48\linewidth}
\centering
\includegraphics[width=\linewidth]{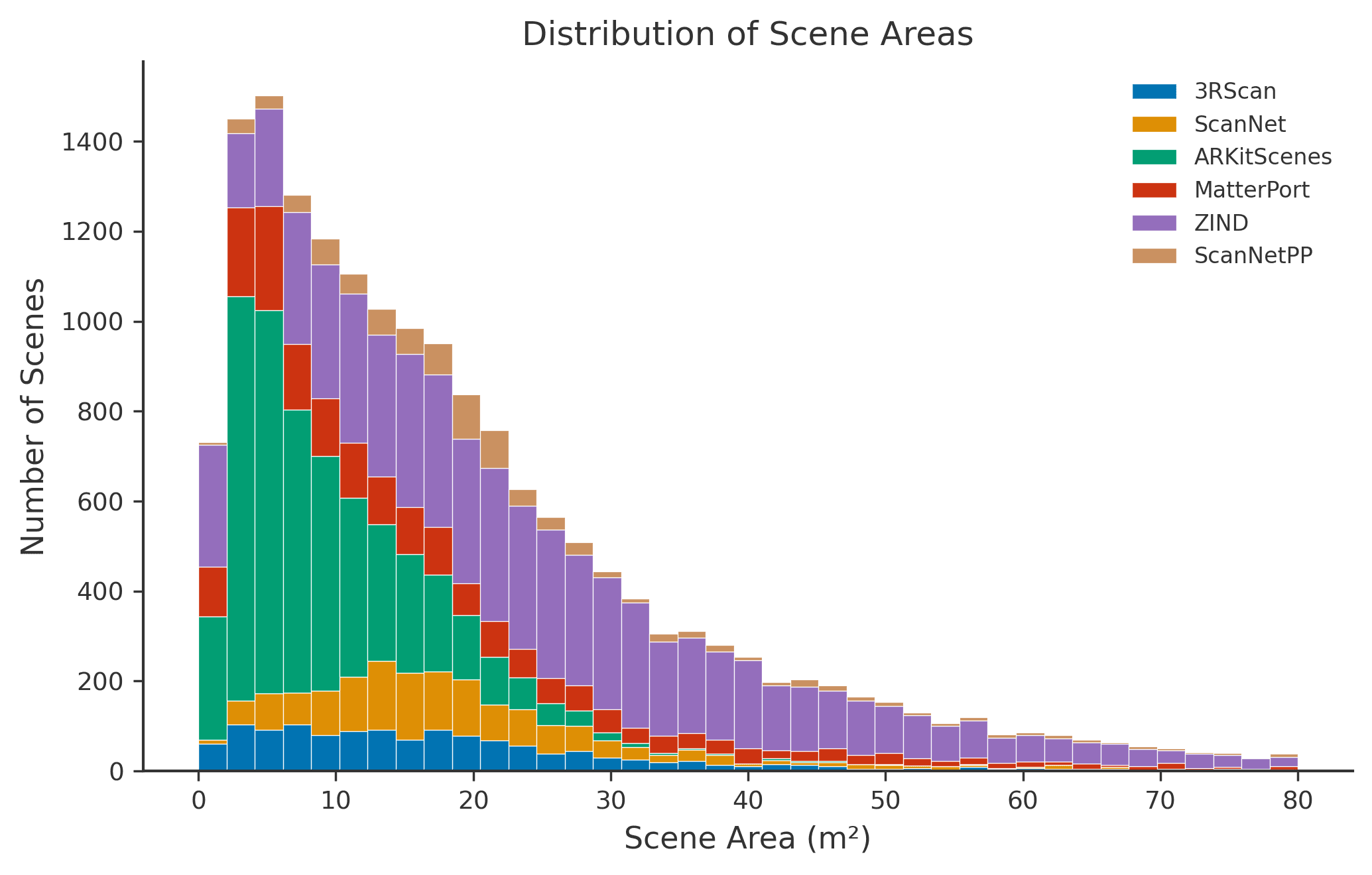}
\caption{Floor areas across 17{,}656 scene records (median $16.4$\,m$^2$; \cref{supp:split_consistency}).}
\label{fig:area_histogram}
\end{minipage}\hfill
\begin{minipage}[t]{0.48\linewidth}
\centering
\includegraphics[width=\linewidth]{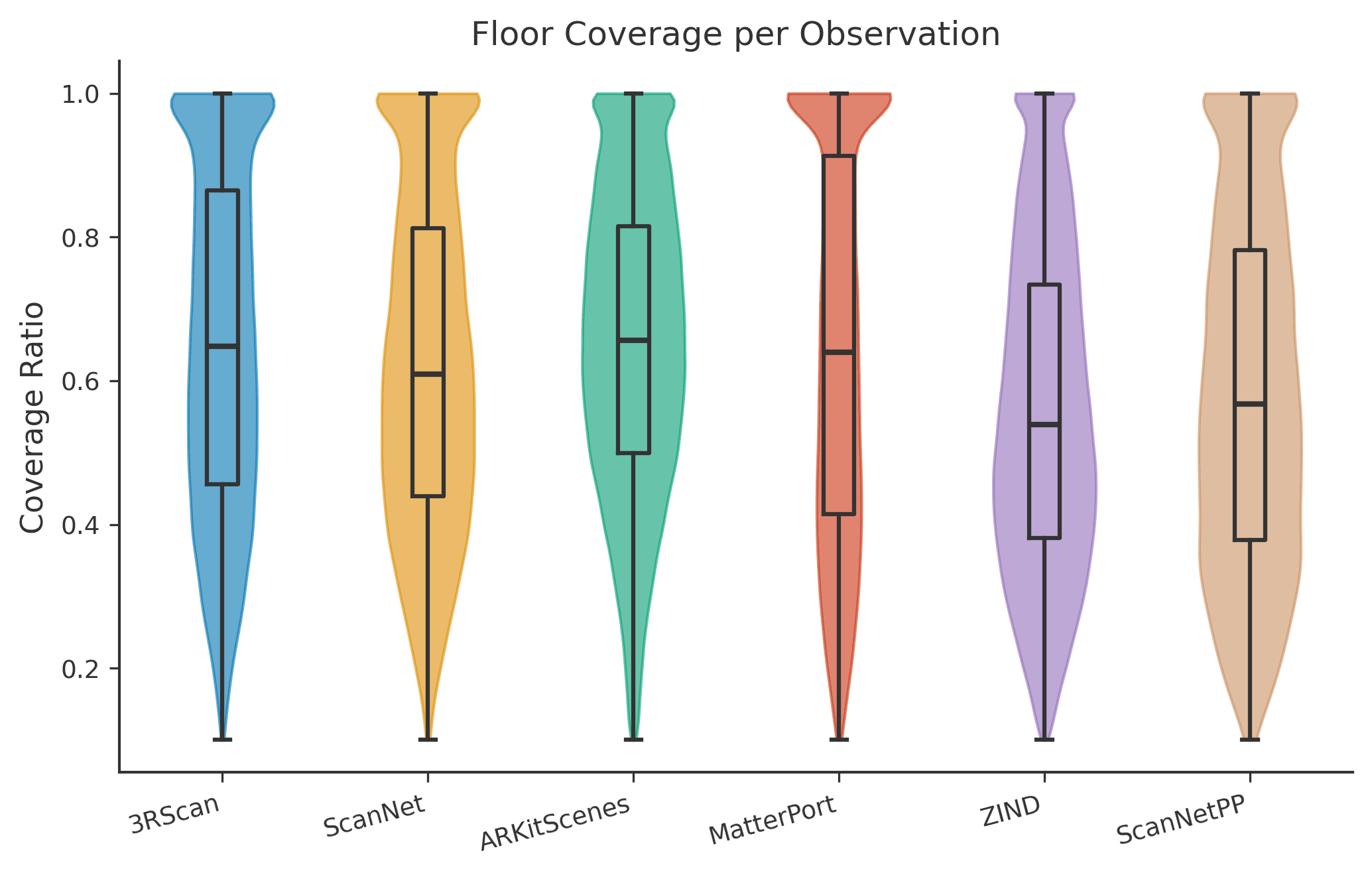}
\caption{Per-viewpoint floor coverage (share of the BEV window that is target floor) across the six source datasets (\cref{supp:split_consistency}).}
\label{fig:coverage_violin}
\end{minipage}
\end{figure}

Before the strict canonical filter, each split's Easy/Learnable/Negligible proportions are within one percentage point of $37.4/47.5/15.1\%$, and the median within-scene standard deviation of $\CondSignalRatio$ is $0.08$, reflecting viewpoint variation within scenes. The canonical $0.10$ filter excludes the Negligible tier. Across 17{,}656 scene records, the median floor area is $16.4$\,m$^2$; per-viewpoint floor coverage, the share of the BEV window that is target floor, ranges from $56\%$ for Matterport3D to $65\%$ for ScanNet (\cref{fig:area_histogram,fig:coverage_violin}).

\section{End-to-End RGB-to-BEV Pipeline}
\label{supp:camera_pipeline}

\noindent\textbf{Test-time processing.}
\Cref{fig:task_pipeline} summarizes the front end and subsequent conditional completion. For each evaluated frame, Depth Pro~\citep{bochkovskiy2025depth} estimates metric depth from one egocentric RGB image. We then back-project via the standard pinhole model. Intrinsics come from each source's camera calibration; Depth Pro focal estimates ($f_x{=}f_y$, image-center principal point) are the fallback for uncalibrated frames. The raw point cloud is voxel-downsampled (0.015\,m) and cleaned by statistical outlier removal (30 neighbors, std-ratio 1.5). Floor pixels are predicted with SegFormer~\citep{Xie2021SegFormer}; the floor plane is estimated with RANSAC~\citep{fischler1981ransac} (1000 trials, 5\,cm threshold) followed by SVD normal estimation on the inlier set.

\noindent\textbf{BEV conditioning.}
Rigid alignment places the estimated floor at $y=0$, with positive $y$ pointing downward. Height filtering retains points with $y\geq-1.25$\,m, removing geometry more than $1.25$\,m above the floor. The retained geometry is rasterized into the common camera-aligned BEV frame in \cref{supp:delta_choice}, yielding the conditioning interface in \cref{sec:problem}. The complete floor target and valid evaluation support come from the source scene and are used only for scoring.

\noindent\textbf{Frame validity filter.}
Frames are rejected when floor-normal estimation fails (floor pixels $<100$, floor coverage $<5\%$, inlier ratio $<60\%$, or geometric degeneracy). The end-to-end qualitative examples (\cref{fig:qualitative_b}) comprise ScanNet++ test-split frames passing this filter. The quantitative cohort contains $1{,}000$ frames across five physical-camera sources; \cref{supp:e2e_multi_source} specifies source composition and the additional ground-truth conditioning-signal threshold used for evaluation.
\par

\section{Baseline Details}
\label{supp:model_formulations}

All learned methods receive the same conditioning pair $(\FloorObs,\MaskUnknown)$ concatenated channel-wise. Cellwise reconstruction, noise-prediction, and velocity-prediction losses use $\MaskEval=\MaskUnknown\odot\MaskValid$; auxiliary losses are specified below.
No data augmentation is applied.
We adapt the U-Net, PConv-UNet, and LaMa backbones to binary BEV inputs and outputs; their losses are described below.
Analog Bits~\citep{threshold} provides a precedent for converting continuous generative outputs into discrete bits; the benchmark uses a common binarization threshold of $0.5$.

\newcolumntype{P}[1]{>{\RaggedRight\arraybackslash}p{#1}}

\begin{table}[!htb]
\centering
\normalsize
\captionsetup{font=normalsize}
\setlength{\tabcolsep}{1.5pt}
\renewcommand{\arraystretch}{1}
\caption{\textit{Baseline catalog.} Unless noted, learned methods share
  inputs $(\FloorObs,\MaskUnknown)$, supervision mask $\MaskEval$, and
  hard evidence clamping before evaluation.}
\label{tab:baseline_catalog}
\begin{tabularx}{\textwidth}{@{}P{0.22\textwidth} >{\RaggedRight\arraybackslash}X P{0.19\textwidth}@{}}
\toprule
\textit{Method} & \textit{Adaptations}
  & \textit{Inference} \\
\midrule

\emph{Training-free baselines}
  & All-Obstacle (fill\,0), All-Floor (fill\,1), NN~Propagation
    (nearest-observed copy), Uniform Random ($p{=}0.5$).
    Parameter-free rules on $\MaskUnknown$; merged with observed
    evidence via hard clamping (\cref{eq:completion_assembly}).
  & Single pass \\

\addlinespace[4pt]
U-Net~\citep{Ronneberger2015UNet,wu2018groupnorm}
  & Masked BCE on $\MaskEval$; binary BEV input/output.
  & Thr.\ 0.5 + clamp \\
PConv-UNet~\citep{Liu2018PartialConv}
  & Masked BCE on $\MaskEval$. Partial-conv.\ mask propagation;
    adapted to binary BEV completion.
  & Thr.\ 0.5 + clamp \\
LaMa~\citep{Chi2020FastFourier,Suvorov2022LAMA}
  & Masked $\ell_1$, adversarial, and feature-matching losses;
    Fourier-conv.\ generator adapted to binary BEV.
  & Thr.\ 0.5 + clamp \\

\addlinespace[4pt]
LaMa-Ensemble\newline\citep{Lakshminarayanan2017DeepEnsembles}
  & Same as LaMa (4 seeds); one prediction per independently trained member.
  & $\NumSamples{=}4$ \\

\addlinespace[4pt]
Diffusion~\citep{ho2020denoising,DDIM}
  & Masked noise-prediction MSE on $\MaskEval$; per-step
    evidence clamping~\citep{christopher2024projecteddiffusion}.
  & DDIM, 50 st., CFG $s{=}2$ \\
Flow Matching~\citep{Lipman2023FlowMatching}
  & Masked velocity-prediction MSE on $\MaskEval$. Same conditioning/clamping as Diffusion;
    ODE transport parameterization.
  & Heun, 50 st., CFG $s{=}2$ \\
FM+XAttn~\citep{Lipman2023FlowMatching,vaswani2017attention}
  & Masked velocity-prediction MSE on $\MaskEval$. Adds condition encoder on
    $[\FloorObs,\MaskUnknown]$ with cross-attn.\
    ($64{\times}64$, $32{\times}32$); see
    \citep{Rombach2022LDM}.
  & Heun, 25 st., CFG $s{=}2$ \\

\bottomrule
\end{tabularx}
\end{table}

\noindent\textbf{U-Net and PConv-UNet.}
Both models are trained with masked binary cross-entropy on $\MaskEval$.
At inference, outputs are binarized at threshold~$0.5$ and hard-clamped to preserve observed evidence.

PConv-UNet additionally propagates a binary partial-convolution mask through its layers using the update rule of \citet{Liu2018PartialConv}.

\noindent\textbf{LaMa and LaMa-Ensemble.}
We adapt LaMa~\citep{Suvorov2022LAMA} from 3-channel RGB inpainting to 1-channel binary completion: the generator accepts 2 input channels $(\FloorObs,\MaskUnknown)$ and produces a single-channel output passed through a sigmoid activation.
LaMa-Ensemble trains four copies with different random seeds, producing one output per independently trained member ($\NumSamples{=}4$ total). Ensembles also appear in map-completion exploration~\citep{ho2025mapex}.

\noindent\textbf{Diffusion.}
We adopt a DDPM~\citep{ho2020denoising} in pixel space with a cosine variance schedule~\citep{nichol2021improved} ($T{=}1000$).
A U-Net backbone $\epsilon_\theta$ predicts the added noise given the noisy sample~$x_t$, timestep~$t$, and conditioning $c=[\FloorObs,\MaskUnknown]$ (concatenated channel-wise). The backbone is trained with masked MSE on~$\MaskEval$. At inference, classifier-free guidance takes the form
\[
\tilde{\epsilon}_\theta = (1{+}s)\,\epsilon_\theta(x_t,t,c) - s\,\epsilon_\theta(x_t,t,\varnothing),
\]
where $\varnothing$ denotes the null input, obtained by zeroing the conditioning channels as in conditioning dropout during training.
At each denoising step, we project onto the observed-label constraints by restoring the observed cells; projected diffusion provides a general precedent~\citep{christopher2024projecteddiffusion}. The final assembly also enforces agreement with the supplied observed evidence.

\noindent\textbf{Flow Matching.}
Flow Matching~\citep{Lipman2023FlowMatching,Liu2023RectifiedFlow,flowcode} learns a velocity field $v_\theta(x_t,t,c)$ along the linear path $x_t=(1{-}t)\,x_0+t\,\FloorGT$ for $t\in[0,1]$ and $x_0\sim\mathcal{N}(0,I)$, trained with masked MSE on $\MaskEval$ against the target velocity $u_t=\FloorGT-x_0$.
Conditioning is identical to that of Diffusion (channel-wise concatenation of $[\FloorObs,\MaskUnknown]$).

\noindent\textbf{FM+XAttn (Flow Matching with Cross-Attention).}
FM+XAttn augments the Flow Matching backbone with an auxiliary condition encoder that processes $[\FloorObs,\MaskUnknown]$ into a sequence of spatial tokens.
Cross-attention layers~\citep{vaswani2017attention} inject these tokens at the coarse resolutions ($64{\times}64$ and $32{\times}32$) of the U-Net, following cross-attention conditioning in generative models~\citep{Rombach2022LDM}.
This supplies spatial condition features to the velocity predictor in addition to channel concatenation.
The cross-attention ablation in \cref{tab:supp:xattn_ablation} compares completion scores with and without these blocks.
The training loss is identical to that of Flow Matching (masked MSE on $\MaskEval$); the architectural additions are the condition encoder and its cross-attention pathway.

\section{Implementation Details}
\label{app:training}

\phantomsection
\noindent\textbf{Hyperparameters.}
\label{supp:hyperparameters}
All single-model runs use seed~$42$; LaMa-Ensemble uses seeds $41$--$44$.
For LaMa and LaMa-Ensemble, the VGG and ResNet perceptual losses are disabled (weight${=}0$), because those features were learned from three-channel natural images rather than single-channel binary maps. Their objectives combine masked $\ell_1$ reconstruction, hinge adversarial loss, and multi-scale feature matching, with weights $\lambda_{\text{rec}}{=}10$, $\lambda_{\text{adv}}{=}10$, and $\lambda_{\text{fm}}{=}250$ (\cref{tab:supp:runtime}).
Training runs on $4{\times}$A100 40\,GB GPUs with distributed data-parallel training,
FP32 precision, and activation checkpointing~\citep{Chen2016Checkpointing};
peak memory is 30--36\,GB per GPU, and training all seven learned method families took approximately $280$ A100 GPU-hours.
At inference, we integrate the learned ODEs used in Flow Matching models from $t{=}0$ to $t{=}1$ using the second-order Heun solver~\citep{heun}.
We use DDIM~\citep{DDIM} with $50$ steps for the diffusion model.
Classifier-free guidance (CFG)~\citep{Ho2021ClassifierFreeGuidance} at scale $s{=}2.0$, conditioning dropout probability $0.1$, and per-step evidence clamping apply to Diffusion, Flow Matching, and FM+XAttn. LaMa-Ensemble produces one prediction per trained member without CFG.

\phantomsection
\noindent\textbf{Runtime and sampling cost.}
\label{sec:supp:runtime_nfe}
Deterministic models use one forward pass per observation; LaMa-Ensemble uses four member predictions. Diffusion and Flow Matching use 50 solver steps per sample; FM+XAttn uses 25. Across $K=4$ outputs, the total solver-update counts are 200, 200, and 100. Each Heun update evaluates the velocity at two stages, and CFG combines conditional and unconditional network predictions at each stage.
Per-observation latencies are 63--81\,ms for deterministic models and 1{,}745--6{,}661\,ms for the four-sample generators (\cref{tab:supp:runtime}). FM+XAttn uses fewer solver updates and has the largest latency in this comparison.
\mbox{Model checkpoints} are selected by peak validation IoU, computed from counts pooled over all validation observations. Fixed random seeds specify the sampled noise.

\begin{table}[!htbp]
\centering
\normalsize
\captionsetup{font=normalsize}
\setlength{\tabcolsep}{1.5pt}
\renewcommand{\arraystretch}{1}
\caption{\textbf{Training configuration and inference runtime.}
\textbf{Top:}\ Optimization settings; model-specific entries are marked.
\textbf{Bottom:}\ Per-observation latency and throughput on a single A100
($256{\times}256$ inputs); generative rows use $K=4$.
Latencies include a fixed per-observation overhead, so LaMa-Ensemble's four member passes take less than four times LaMa's latency.
\emph{Updates} counts forward passes for feedforward methods and solver
steps summed over outputs for generators; network function evaluations
also depend on the solver and guidance.}
\label{tab:supp:runtime}

\begin{tabularx}{\linewidth}{@{}P{0.33\linewidth} >{\RaggedRight\arraybackslash}X@{}}
\toprule
\multicolumn{2}{@{}l@{}}{\textbf{Training hyperparameters}} \\
\midrule
Optimizer
    & AdamW~\citep{Loshchilov2019AdamW} \\
Learning rate
    & $1 \times 10^{-4}$ \\
Weight decay
    & $1 \times 10^{-2}$ \\
LR schedule
    & Cosine annealing~\citep{Loshchilov2017SGDR} (no restarts) \\
Effective batch size
    & $64$ ($16$/GPU $\times$ $4$ GPUs) \\
Training steps
    & $300{,}000$ \\
Precision
    & FP32 \\
Gradient accumulation
    & 1 step \\
BCE positive weight
    & $1.0$ (U-Net and PConv-UNet only) \\
CFG dropout $p_{\text{drop}}$
    & $0.1$ (Diffusion and flow models only) \\
\midrule
\multicolumn{2}{@{}l@{}}{\emph{LaMa / LaMa-Ensemble additional}} \\
\midrule
$\lambda_{\text{rec}}$ / $\lambda_{\text{adv}}$ / $\lambda_{\text{fm}}$
    & $10.0$ / $10.0$ / $250.0$ \\
Seeds (LaMa-Ensemble)
    & 41, 42, 43, 44 \\
\bottomrule
\end{tabularx}

\par\smallskip

\begin{tabular*}{\linewidth}{@{\extracolsep{\fill}}llcrr@{}}
\toprule
\multicolumn{5}{@{}l@{}}{\textbf{Inference runtime}} \\
\midrule
Method        & Sampler & Updates & Latency (ms) & FPS \\
\midrule
U-Net         & Forward & 1       & 81           & 12.3 \\
PConv-UNet    & Forward & 1       & 68           & 14.7 \\
LaMa          & Forward & 1       & \textbf{63}  & \textbf{15.9} \\
LaMa-Ensemble & Forward & 4       & 136          & 7.4 \\
Diffusion     & DDIM    & 200     & 1745         & 0.6 \\
Flow Matching & Heun    & 200     & 3377         & 0.3 \\
FM+XAttn      & Heun    & 100     & 6661         & 0.2 \\
\bottomrule
\end{tabular*}
\end{table}

\phantomsection
\noindent\textbf{Guidance and conditioning ablations.}
\label{supp:cfg_sensitivity}

The default guidance scale $s=2.0$ was selected on the training split. \Cref{tab:supp:cfg_ablation} evaluates post-hoc OOD sensitivity at five scales from $0.5$ to $5.0$, holding all other settings fixed. All three generators reach their highest tested OOD best-of-four IoU at $s=2.0$.

The architecture ablation compares FM+XAttn with and without cross-attention at fixed $s=2.0$ and $K=4$ (\cref{tab:supp:xattn_ablation}). The ablated variant retrains FM+XAttn without its cross-attention blocks, with all other settings fixed. Removing the blocks changes OOD MES from $0.095$ to $0.096$ and best-of-four IoU from $0.873$ to $0.872$.

\begin{table}[!t]
\centering
\normalsize
\captionsetup{font=normalsize}
\setlength{\tabcolsep}{1.5pt}
\renewcommand{\arraystretch}{1}
\caption{Post-hoc guidance-scale sensitivity on the OOD split, evaluated using best-of-four IoU ($\uparrow$). Only the classifier-free guidance scale changes. Bold marks the best tested scale for each method.}
\label{tab:supp:cfg_ablation}
\begin{tabular*}{\linewidth}{@{\extracolsep{\fill}}lccc@{}}
\toprule
$s$ & Diffusion $\uparrow$ & Flow Matching $\uparrow$ & FM+XAttn $\uparrow$ \\
\midrule
0.5            & 0.850 & 0.833 & 0.819 \\
1.0            & 0.852 & 0.816 & 0.867 \\
\textbf{2.0}   & \textbf{0.890} & \textbf{0.880} & \textbf{0.873} \\
3.0            & 0.870 & 0.848 & 0.835 \\
5.0            & 0.843 & 0.804 & 0.826 \\
\bottomrule
\end{tabular*}
\end{table}

\begin{table}[!t]
\centering
\normalsize
\captionsetup{font=normalsize}
\setlength{\tabcolsep}{1.5pt}
\renewcommand{\arraystretch}{1}
\caption{Conditioning architecture ablation for FM+XAttn at $s{=}2.0$, $K{=}4$, on the OOD split. MES evaluates the sample set ($\downarrow$); IoU$_b$ evaluates the best of four candidates ($\uparrow$).}
\label{tab:supp:xattn_ablation}
\begin{tabularx}{\linewidth}{@{}>{\RaggedRight\arraybackslash}Xcc@{}}
\toprule
Variant & MES $\downarrow$ & IoU$_b$ OOD $\uparrow$ \\
\midrule
FM+XAttn (full; cross-attn at $64{\times}64$, $32{\times}32$) & \textbf{0.095} & \textbf{0.873} \\
FM+XAttn (ablated; no cross-attention)                        & 0.096 & 0.872          \\
\bottomrule
\end{tabularx}
\end{table}

\section{Evaluation Metric Details}
\label{supp:metrics}

The completion metrics defined here use the interface and evaluation region specified in \cref{sec:problem,sec:evaluation}.

\phantomsection
\noindent\textbf{Fidelity metrics.}
\label{supp:metrics_fidelity}
Fidelity metrics are computed on the evaluation mask
$\MaskEval=\MaskUnknown\odot\MaskValid$.
Before scoring, we hard-clamp observed evidence into every prediction.

IoU and F1 metrics for the floor class are computed from masked counts of true positives, false positives, and false negatives (TP, FP, FN) restricted to $\MaskEval$. IoU divides TP by $\mathrm{TP}+\mathrm{FP}+\mathrm{FN}$; F1 divides $2\mathrm{TP}$ by $2\mathrm{TP}+\mathrm{FP}+\mathrm{FN}$. Both operate on floor sets within the scored region.

\noindent\textbf{Mean-of-four IoU.}
For observation $i$, let $\hat F_i^{(k)}$ denote the $k$th binary completion, with output values $\geq0.5$ labeled floor and observed evidence clamped. We score each completion separately and average the four IoUs:
\begin{equation}
\mathrm{IoU}_{m,i}=\frac14\sum_{k=1}^{4}
\mathrm{IoU}_{R_{\mathrm{eval},i}}\!\left(\hat F_i^{(k)},F_i^\star\right).
\label{eq:mean_iou}
\end{equation}

\phantomsection\label{app:mes}
\noindent\textbf{Population masked energy score.}
The population form of the masked energy score (MES) underlying the finite-sample estimator in the main paper (\cref{eq:es_hat}) uses masked Jaccard distance $d_{\MaskEval}(A,B)=1-\mathrm{IoU}_{\MaskEval}(A,B)$ as the dissimilarity. On masks restricted to $\MaskEval$, it is a bounded metric when the distance between two empty sets is defined as zero~\citep{Lipkus1999JaccardMetric}:
\begin{equation}
    \mathrm{MES}_{\MaskEval}(P,\FloorGT):=\mathbb{E}[d_{\MaskEval}(Y,\FloorGT)]-\tfrac{1}{2}\mathbb{E}[d_{\MaskEval}(Y,Y')],
\end{equation}
with $Y,Y'\overset{iid}{\sim}P$~\citep{gneiting2007proper,rizzo2016energy}.
The first term is the expected distance to the target; the second subtracts half the expected distance between two independent samples.
For deterministic predictors, $Y{=}Y'$ a.s.\ and MES reduces to $1-\mathrm{IoU}_{\MaskEval}$.

\noindent\textbf{Propriety on a fixed evaluation mask.}
Fix a finite, nonempty scored pixel set $E$ and restrict floor sets to $E$.
With $\mathrm{IoU}(\varnothing,\varnothing)=1$, the Jaccard similarity matrix $\mathbf J_E$ over distinct subsets of $E$ is strictly positive definite~\citep{Bouchard2013Jaccard}.
For distributions $P,Q$ over these subsets, with probability vectors $\mathbf p,\mathbf q$,
\begin{equation}
\mathbb E_{Z\sim Q}\!\left[\mathrm{MES}_E(P,Z)-\mathrm{MES}_E(Q,Z)\right]
=\tfrac12(\mathbf p-\mathbf q)^\top\mathbf J_E(\mathbf p-\mathbf q).
\end{equation}
This quantity is positive when $P\ne Q$, establishing strict propriety for distributions over floor sets on the fixed evaluation mask $E$.

\phantomsection
\noindent\textbf{Finite-sample estimators.}
\label{supp:metric_interpretation}
The second sum in \cref{eq:es_hat} contains all ordered pairs of distinct samples.
For i.i.d.\ draws from one conditional distribution, the $K>1$ expression is an unbiased estimator of the population MES above. Deterministic predictors ($K=1$) use the population limit given above.

\noindent\textbf{Binary sample variance.}
The Var columns and boundary analysis summarize disagreement among four binary completions thresholded at $0.5$. For their labels $b_k(x)\in\{0,1\}$ at cell $x$, define
\begin{equation}
v(x)=\frac14\sum_{k=1}^{4}\bigl(b_k(x)-\bar b(x)\bigr)^2,
\qquad \bar b(x)=\frac14\sum_{k=1}^{4}b_k(x).
\label{eq:binary_variance}
\end{equation}
The Var columns average $v(x)$ over $\MaskEval$; boundary and interior summaries average it within their respective regions (\cref{supp:boundary_radius}).

\noindent\textbf{Spatial error.}
UMR, IoU, and F1 depend on class prevalence and average errors across space; high floor prevalence can obscure local boundary defects.
The offline planning study in \cref{sec:route_safety} evaluates the resulting maps as route-planning inputs.

\par

\section{Extended Qualitative Examples}
\label{supp:qualitative_section}
These examples illustrate completions on the canonical track and extend the RGB comparisons and failure cases in \cref{fig:qualitative_b,fig:failure_case}; numerical results are in \cref{supp:expanded_metrics}.

\begin{figure}
\centering
\setlength{\tabcolsep}{1.5pt}%
\resizebox{\textwidth}{!}{%
\begin{tabular}{@{}cccccccccc@{}}
\footnotesize $\FloorObs$ &
\footnotesize $\MaskUnknown$ &
\footnotesize GT &
\footnotesize All-Obstacle &
\footnotesize All-Floor &
\footnotesize\shortstack{NN\\Propagation} &
\footnotesize\shortstack{Uniform\\Random} &
\footnotesize U-Net &
\footnotesize PConv-UNet &
\footnotesize LaMa \\[3pt]
\fcolorbox{mapborder}{white}{\includegraphics[width=1.6cm]{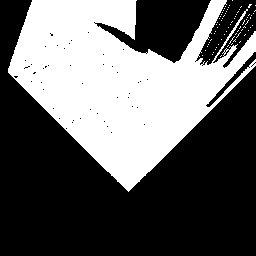}} &
\fcolorbox{mapborder}{white}{\includegraphics[width=1.6cm]{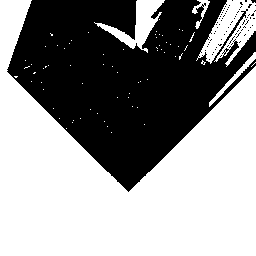}} &
\fcolorbox{mapborder}{white}{\includegraphics[width=1.6cm]{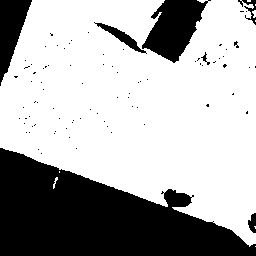}} &
\fcolorbox{mapborder}{white}{\includegraphics[width=1.6cm]{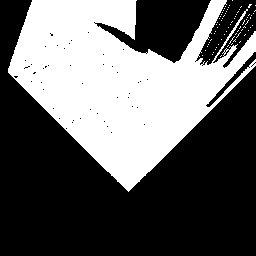}} &
\fcolorbox{mapborder}{white}{\includegraphics[width=1.6cm]{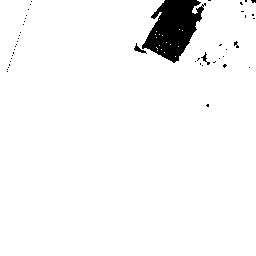}} &
\fcolorbox{mapborder}{white}{\includegraphics[width=1.6cm]{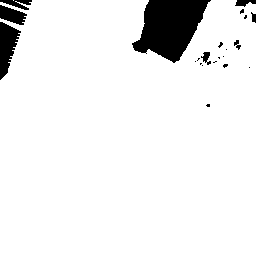}} &
\fcolorbox{mapborder}{white}{\includegraphics[width=1.6cm]{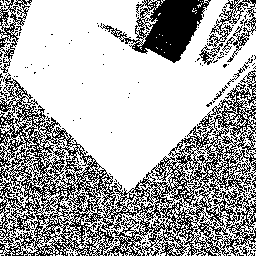}} &
\fcolorbox{mapborder}{white}{\includegraphics[width=1.6cm]{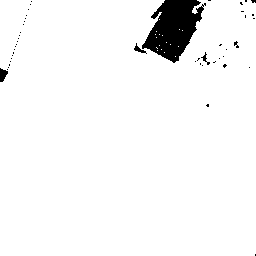}} &
\fcolorbox{mapborder}{white}{\includegraphics[width=1.6cm]{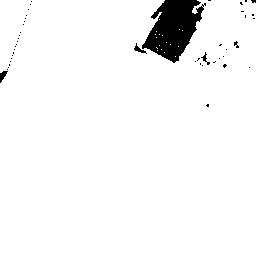}} &
\fcolorbox{mapborder}{white}{\includegraphics[width=1.6cm]{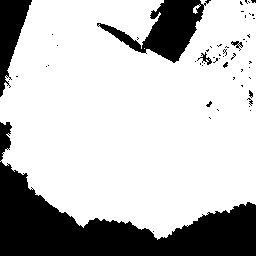}} \\[2pt]
\fcolorbox{mapborder}{white}{\includegraphics[width=1.6cm]{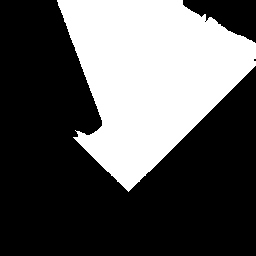}} &
\fcolorbox{mapborder}{white}{\includegraphics[width=1.6cm]{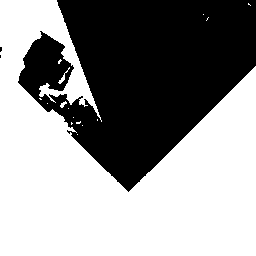}} &
\fcolorbox{mapborder}{white}{\includegraphics[width=1.6cm]{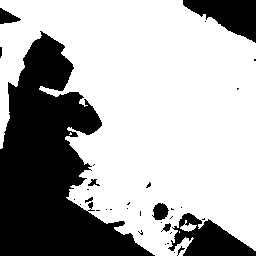}} &
\fcolorbox{mapborder}{white}{\includegraphics[width=1.6cm]{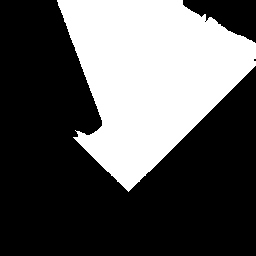}} &
\fcolorbox{mapborder}{white}{\includegraphics[width=1.6cm]{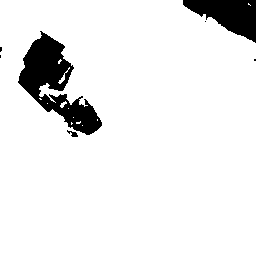}} &
\fcolorbox{mapborder}{white}{\includegraphics[width=1.6cm]{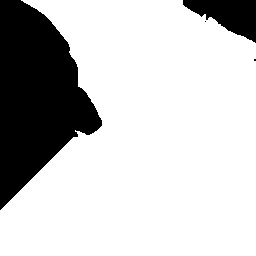}} &
\fcolorbox{mapborder}{white}{\includegraphics[width=1.6cm]{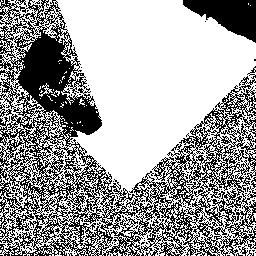}} &
\fcolorbox{mapborder}{white}{\includegraphics[width=1.6cm]{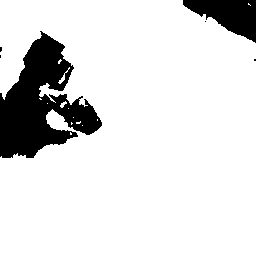}} &
\fcolorbox{mapborder}{white}{\includegraphics[width=1.6cm]{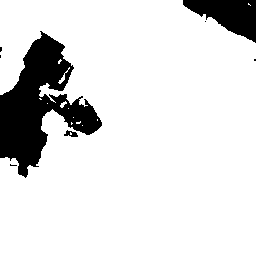}} &
\fcolorbox{mapborder}{white}{\includegraphics[width=1.6cm]{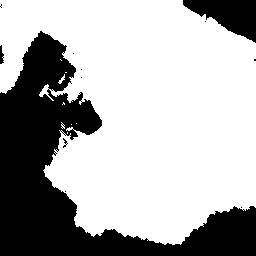}} \\
\end{tabular}}%
\par
\centering
\setlength{\tabcolsep}{1.5pt}%
\resizebox{\textwidth}{!}{%
\begin{tabular}{@{}ccccc@{\hspace{6pt}}ccccc@{}}
\multicolumn{5}{c}{\footnotesize LaMa-Ensemble} &
\multicolumn{5}{c}{\footnotesize Diffusion} \\
\scriptsize $\hat{F}^{(1)}$ & \scriptsize $\hat{F}^{(2)}$ & \scriptsize $\hat{F}^{(3)}$ & \scriptsize $\hat{F}^{(4)}$ & \scriptsize Variance &
\scriptsize $\hat{F}^{(1)}$ & \scriptsize $\hat{F}^{(2)}$ & \scriptsize $\hat{F}^{(3)}$ & \scriptsize $\hat{F}^{(4)}$ & \scriptsize Variance \\[2pt]
\fcolorbox{mapborder}{white}{\includegraphics[width=1.6cm]{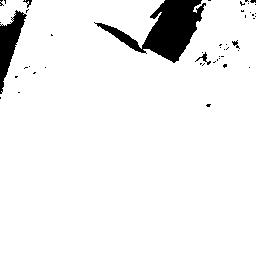}} &
\fcolorbox{mapborder}{white}{\includegraphics[width=1.6cm]{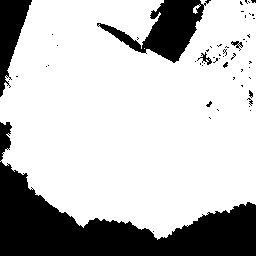}} &
\fcolorbox{mapborder}{white}{\includegraphics[width=1.6cm]{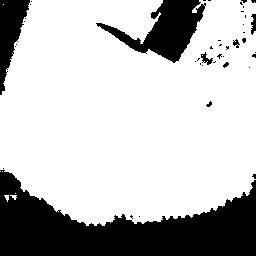}} &
\fcolorbox{mapborder}{white}{\includegraphics[width=1.6cm]{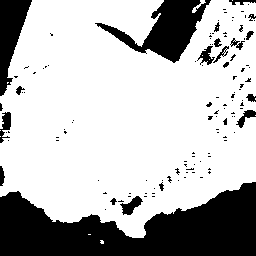}} &
\mapframe{\includegraphics[width=1.6cm]{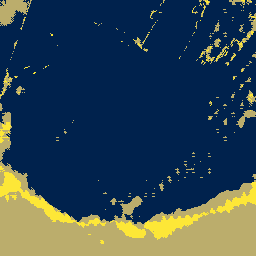}} &
\fcolorbox{mapborder}{white}{\includegraphics[width=1.6cm]{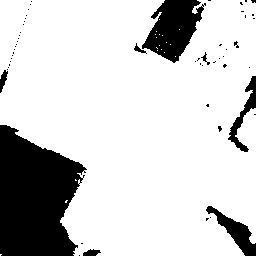}} &
\fcolorbox{mapborder}{white}{\includegraphics[width=1.6cm]{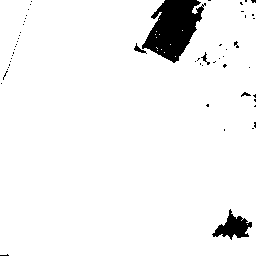}} &
\fcolorbox{mapborder}{white}{\includegraphics[width=1.6cm]{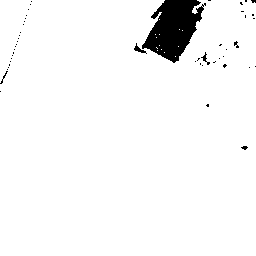}} &
\fcolorbox{mapborder}{white}{\includegraphics[width=1.6cm]{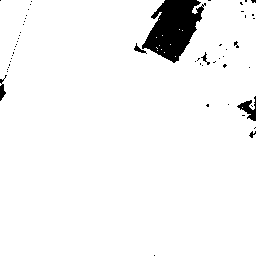}} &
\mapframe{\includegraphics[width=1.6cm]{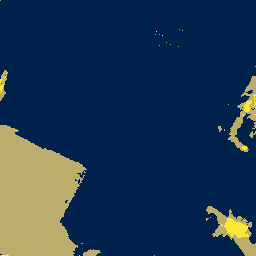}} \\[3pt]
\multicolumn{5}{c}{\footnotesize Flow Matching} &
\multicolumn{5}{c}{\footnotesize FM+XAttn} \\
\scriptsize $\hat{F}^{(1)}$ & \scriptsize $\hat{F}^{(2)}$ & \scriptsize $\hat{F}^{(3)}$ & \scriptsize $\hat{F}^{(4)}$ & \scriptsize Variance &
\scriptsize $\hat{F}^{(1)}$ & \scriptsize $\hat{F}^{(2)}$ & \scriptsize $\hat{F}^{(3)}$ & \scriptsize $\hat{F}^{(4)}$ & \scriptsize Variance \\[2pt]
\fcolorbox{mapborder}{white}{\includegraphics[width=1.6cm]{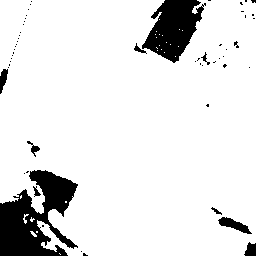}} &
\fcolorbox{mapborder}{white}{\includegraphics[width=1.6cm]{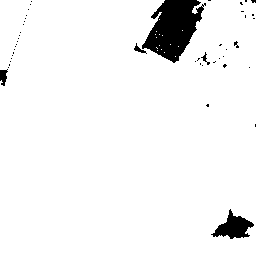}} &
\fcolorbox{mapborder}{white}{\includegraphics[width=1.6cm]{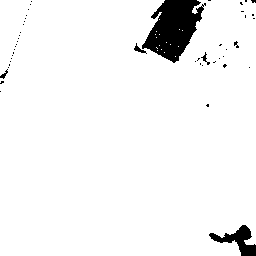}} &
\fcolorbox{mapborder}{white}{\includegraphics[width=1.6cm]{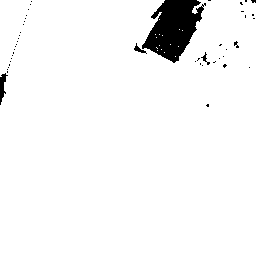}} &
\mapframe{\includegraphics[width=1.6cm]{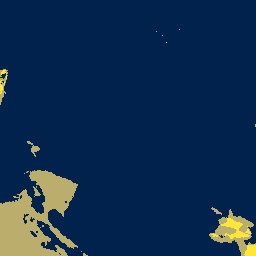}} &
\fcolorbox{mapborder}{white}{\includegraphics[width=1.6cm]{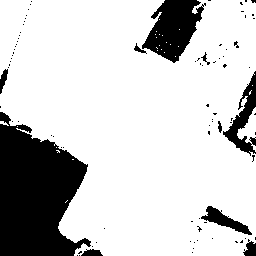}} &
\fcolorbox{mapborder}{white}{\includegraphics[width=1.6cm]{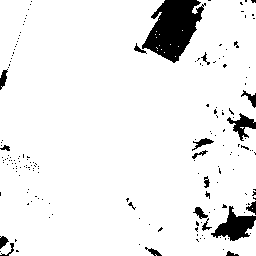}} &
\fcolorbox{mapborder}{white}{\includegraphics[width=1.6cm]{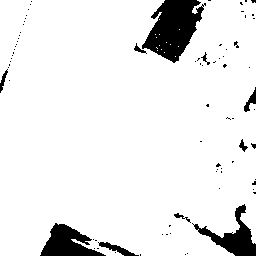}} &
\fcolorbox{mapborder}{white}{\includegraphics[width=1.6cm]{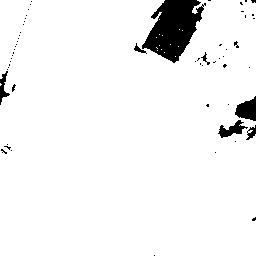}} &
\mapframe{\includegraphics[width=1.6cm]{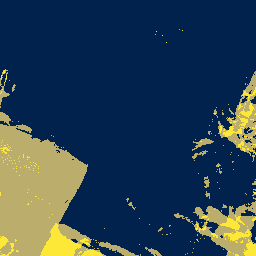}} \\
\end{tabular}}%
\caption{\textbf{Completion from source-geometry BEV observations.} Top: single-output baselines for one ID and one OOD scene (rows 1--2), preceded by conditioning and ground truth. Bottom: four outputs from LaMa-Ensemble and each generator for the ID scene, with per-cell variance (brighter means greater disagreement). White/black denote floor/non-floor; the mask marks unobserved cells. Observed labels remain fixed. Extended scenes: \cref{fig:qualitative_supp_a}.}
\label{fig:canonical_full}
\end{figure}

\Cref{fig:canonical_full} gives the canonical comparison referenced from \cref{sec:results}, and \cref{fig:qualitative_supp_a} extends it. Source geometry supplies the BEV evidence used by every completion method. \Cref{fig:canonical_full} shows single-output baselines and all four outputs from LaMa-Ensemble and each generator. \Cref{fig:qualitative_supp_a} shows one output per method, using best-of-four IoU on $\MaskEval$ for the multi-output methods. Its columns are \textbf{O}: observed floor $\FloorObs$; \textbf{M}: evaluation mask $\MaskEval$; \textbf{G}: ground truth; \textbf{AO}: All-Obstacle; \textbf{AF}: All-Floor; \textbf{NN}: NN Propagation; \textbf{Un}: Uniform Random; \textbf{U}: U-Net; \textbf{PC}: PConv-UNet; \textbf{LE}: LaMa-Ensemble; \textbf{Di}: Diffusion; \textbf{Fl}: Flow Matching; and \textbf{XA}: FM+XAttn.

\newlength{\montwidthA}\setlength{\montwidthA}{0.86\textwidth}%
\newlength{\montcolA}\setlength{\montcolA}{\dimexpr\montwidthA/13\relax}%
\begin{figure}[htbp]
\centering
{\scriptsize\noindent
\makebox[\montcolA][c]{\textbf{O}}%
\makebox[\montcolA][c]{\textbf{M}}%
\makebox[\montcolA][c]{\textbf{G}}%
\makebox[\montcolA][c]{\textbf{AO}}%
\makebox[\montcolA][c]{\textbf{AF}}%
\makebox[\montcolA][c]{\textbf{NN}}%
\makebox[\montcolA][c]{\textbf{Un}}%
\makebox[\montcolA][c]{\textbf{U}}%
\makebox[\montcolA][c]{\textbf{PC}}%
\makebox[\montcolA][c]{\textbf{LE}}%
\makebox[\montcolA][c]{\textbf{Di}}%
\makebox[\montcolA][c]{\textbf{Fl}}%
\makebox[\montcolA][c]{\textbf{XA}}}\\[1pt]
\mapgrid{13}{10}{\includegraphics[width=\montwidthA]{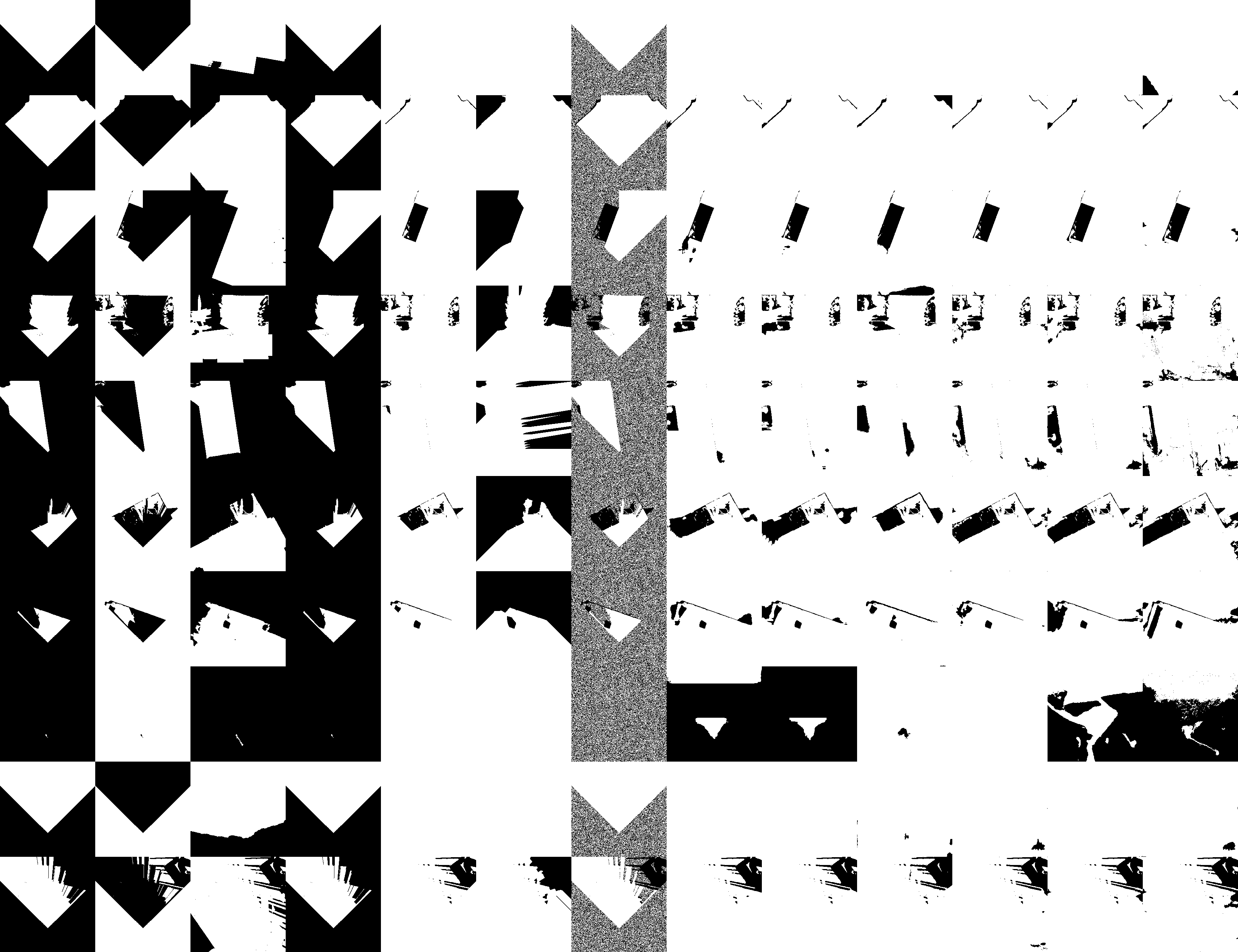}}
\caption{\textbf{Extended canonical comparison.} Source-geometry conditioning with single-output baselines and ground-truth-selected best-of-four outputs from the ensemble and generators. Column definitions and selection are specified in \cref{supp:qualitative_section}.}
\label{fig:qualitative_supp_a}
\end{figure}

\begin{figure}[!htbp]
\centering
\setlength{\tabcolsep}{2pt}
\renewcommand{\arraystretch}{1.05}
\resizebox{\linewidth}{!}{%
\begin{tabular}{@{}ccccccc@{}}
\toprule
$\FloorObs$ & $\FloorGT$ & $\FloorGT \!\setminus\! \FloorObs$ & Diffusion & Flow Matching & FM+XAttn & LaMa \\
\midrule
\fcolorbox{mapborder}{white}{\includegraphics[width=2.0cm]{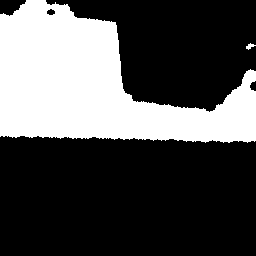}} &
\fcolorbox{mapborder}{white}{\includegraphics[width=2.0cm]{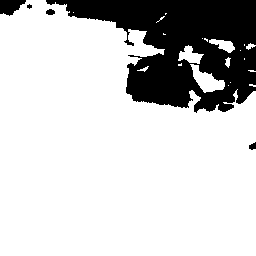}} &
\mapframe{\includegraphics[width=2.0cm]{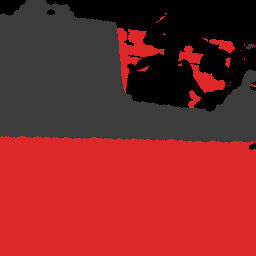}} &
\fcolorbox{mapborder}{white}{\includegraphics[width=2.0cm]{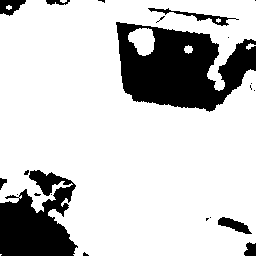}} &
\fcolorbox{mapborder}{white}{\includegraphics[width=2.0cm]{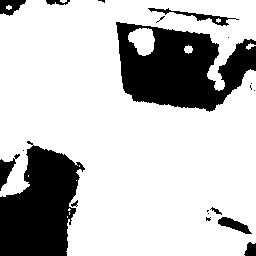}} &
\fcolorbox{mapborder}{white}{\includegraphics[width=2.0cm]{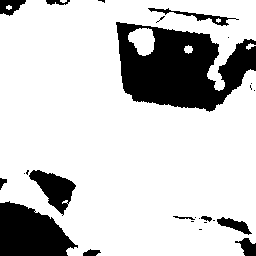}} &
\fcolorbox{mapborder}{white}{\includegraphics[width=2.0cm]{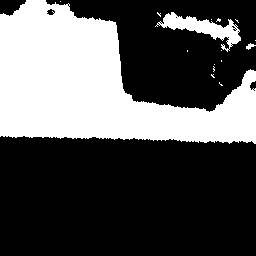}} \\[3pt]
\fcolorbox{mapborder}{white}{\includegraphics[width=2.0cm]{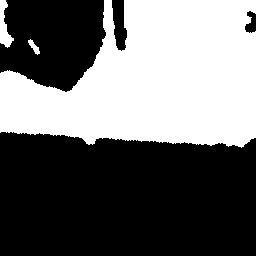}} &
\fcolorbox{mapborder}{white}{\includegraphics[width=2.0cm]{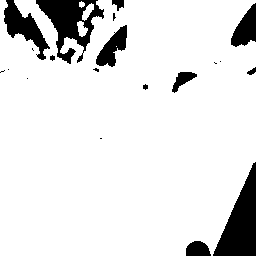}} &
\mapframe{\includegraphics[width=2.0cm]{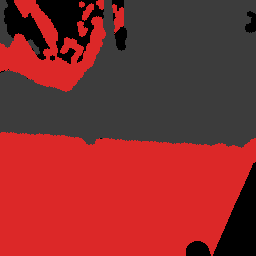}} &
\fcolorbox{mapborder}{white}{\includegraphics[width=2.0cm]{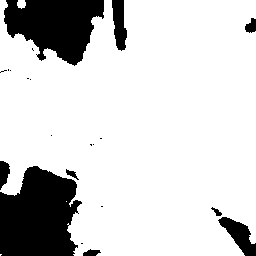}} &
\fcolorbox{mapborder}{white}{\includegraphics[width=2.0cm]{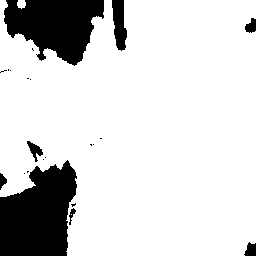}} &
\fcolorbox{mapborder}{white}{\includegraphics[width=2.0cm]{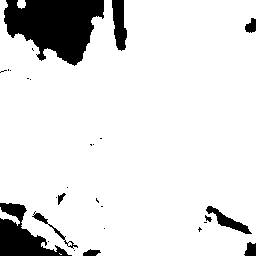}} &
\fcolorbox{mapborder}{white}{\includegraphics[width=2.0cm]{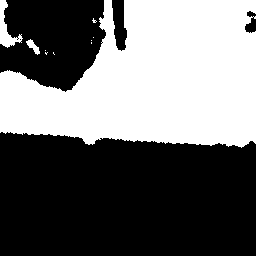}} \\[3pt]
\fcolorbox{mapborder}{white}{\includegraphics[width=2.0cm]{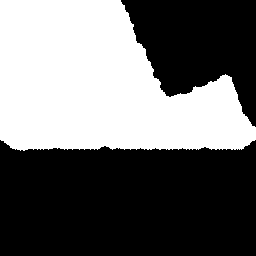}} &
\fcolorbox{mapborder}{white}{\includegraphics[width=2.0cm]{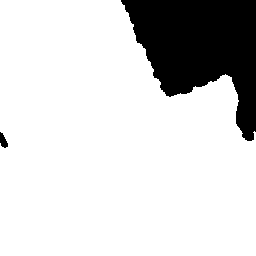}} &
\mapframe{\includegraphics[width=2.0cm]{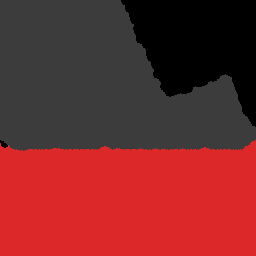}} &
\fcolorbox{mapborder}{white}{\includegraphics[width=2.0cm]{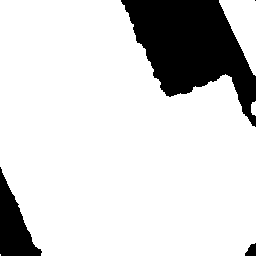}} &
\fcolorbox{mapborder}{white}{\includegraphics[width=2.0cm]{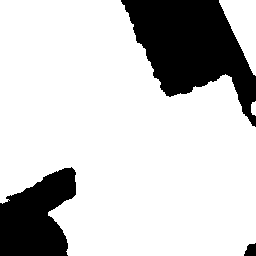}} &
\fcolorbox{mapborder}{white}{\includegraphics[width=2.0cm]{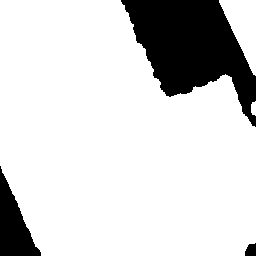}} &
\fcolorbox{mapborder}{white}{\includegraphics[width=2.0cm]{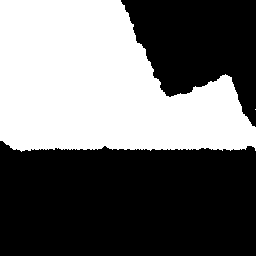}} \\
\bottomrule
\end{tabular}}
\caption{\textbf{Supplementary failure cases.} Three additional test scenes show disconnected floor islands (row~1), coarse-shape drift (row~2), and residual boundary artifacts (row~3).
The boundary-leakage case is shown in \cref{fig:failure_case}.
The red overlay ($\FloorGT \!\setminus\! \FloorObs$) shows target floor absent from the observation (50--61\% across cases).
Each generator shows its first generated sample; LaMa shows the single model trained with seed $42$. LaMa produces small floor regions in every case.}
\label{fig:supp_failure_grid}
\end{figure}

\noindent\textbf{Failure cases.}
In \cref{fig:supp_failure_grid}, LaMa obtains IoU\,$\leq$\,0.009 throughout.
\textbf{Row~1}: IoUs are $0.824$ (FM+XAttn) and $0.821$ (Diffusion).
\textbf{Row~2}: $0.913$ (FM+XAttn), $0.801$ (Flow Matching), and $0.788$ (Diffusion).
\textbf{Row~3}: $0.927$ (Diffusion) and $0.926$ (FM+XAttn).

\par

\phantomsection\label{supp:rgb_gallery}
\newsavebox{\montboximgB}%
\savebox{\montboximgB}{\mapgrid{14}{20}{\includegraphics[width=\textwidth,height=0.80\textheight,keepaspectratio]{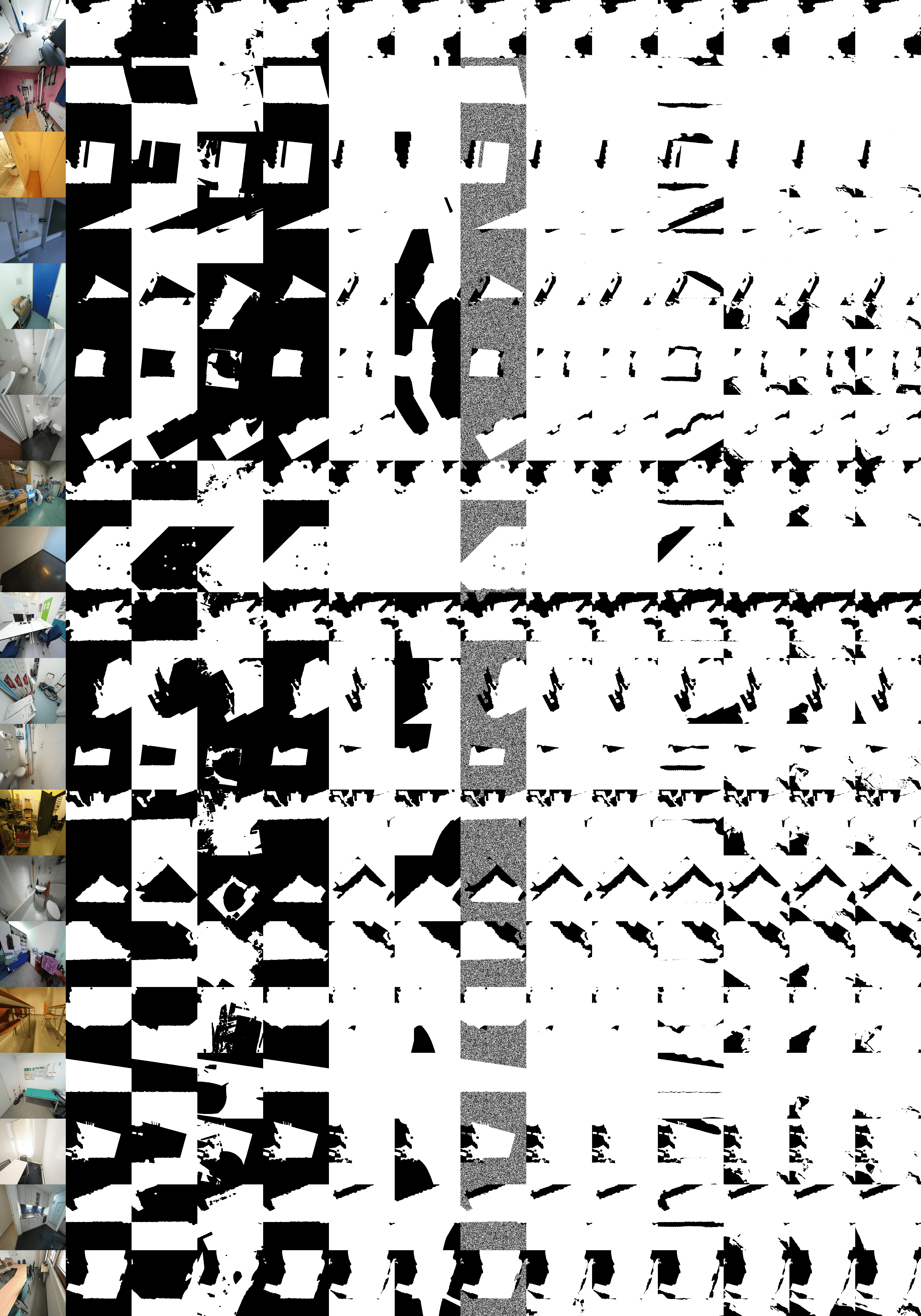}}}%
\newlength{\montcolB}\setlength{\montcolB}{\dimexpr\wd\montboximgB/14\relax}%
\begin{figure}[htbp]
\centering
{\scriptsize\noindent
\makebox[\montcolB][c]{\textbf{R}}%
\makebox[\montcolB][c]{\textbf{O}}%
\makebox[\montcolB][c]{\textbf{M}}%
\makebox[\montcolB][c]{\textbf{G}}%
\makebox[\montcolB][c]{\textbf{AO}}%
\makebox[\montcolB][c]{\textbf{AF}}%
\makebox[\montcolB][c]{\textbf{NN}}%
\makebox[\montcolB][c]{\textbf{Un}}%
\makebox[\montcolB][c]{\textbf{U}}%
\makebox[\montcolB][c]{\textbf{PC}}%
\makebox[\montcolB][c]{\textbf{LE}}%
\makebox[\montcolB][c]{\textbf{Di}}%
\makebox[\montcolB][c]{\textbf{Fl}}%
\makebox[\montcolB][c]{\textbf{XA}}}\\[1pt]
\usebox{\montboximgB}
\caption{\textbf{Extended RGB comparisons (1/2).} \textbf{R}: RGB; \textbf{M}: supplied unobserved-region mask; other columns follow \cref{supp:qualitative_section}. Each generator shows its first generated sample; LaMa-Ensemble shows the individual output of its seed-$41$ member before ensemble averaging. The observed floormap $\FloorObs$ comes from estimated depth, segmentation, and projection. Continued on the next page.}
\label{fig:qualitative_supp_b}
\end{figure}

\addtocounter{figure}{-1}
\newsavebox{\montboximgBii}%
\savebox{\montboximgBii}{\mapgrid{14}{19}{\includegraphics[trim=0 256bp 0 0,clip,width=\textwidth,height=0.83\textheight,keepaspectratio]{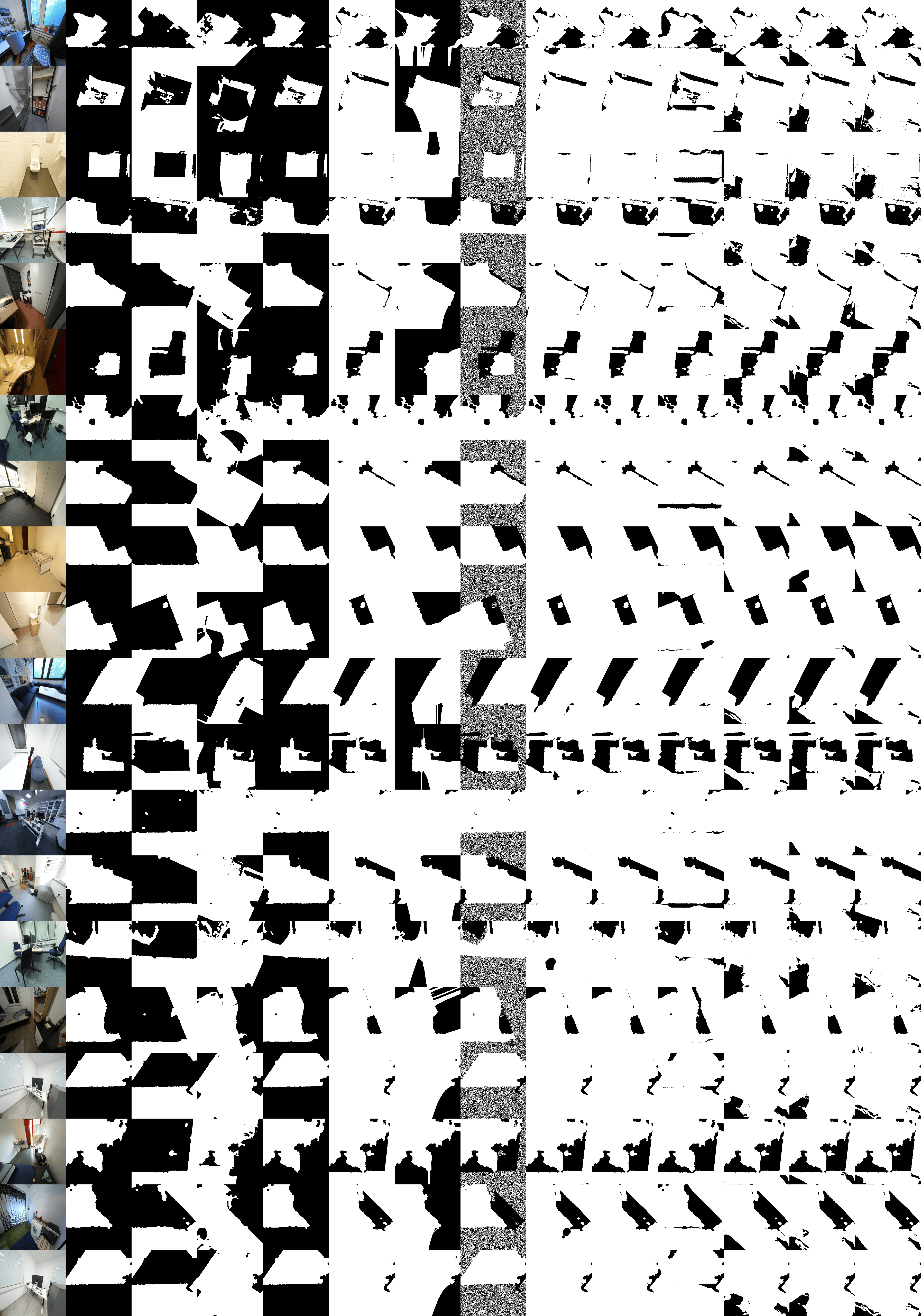}}}%
\newlength{\montcolBii}\setlength{\montcolBii}{\dimexpr\wd\montboximgBii/14\relax}%
\begin{figure*}[p]
\centering
{\scriptsize\noindent
\makebox[\montcolBii][c]{\textbf{R}}%
\makebox[\montcolBii][c]{\textbf{O}}%
\makebox[\montcolBii][c]{\textbf{M}}%
\makebox[\montcolBii][c]{\textbf{G}}%
\makebox[\montcolBii][c]{\textbf{AO}}%
\makebox[\montcolBii][c]{\textbf{AF}}%
\makebox[\montcolBii][c]{\textbf{NN}}%
\makebox[\montcolBii][c]{\textbf{Un}}%
\makebox[\montcolBii][c]{\textbf{U}}%
\makebox[\montcolBii][c]{\textbf{PC}}%
\makebox[\montcolBii][c]{\textbf{LE}}%
\makebox[\montcolBii][c]{\textbf{Di}}%
\makebox[\montcolBii][c]{\textbf{Fl}}%
\makebox[\montcolBii][c]{\textbf{XA}}}\\[1pt]
\usebox{\montboximgBii}
\caption{%
\textbf{Extended RGB comparisons (2/2).} Additional frames with the columns and displayed-output rules defined in \cref{fig:qualitative_supp_b}.}
\end{figure*}

\end{document}